\documentclass{article}

\usepackage[preprint]{neurips_2024}

\usepackage{verbatim}
\usepackage{graphicx}
\usepackage{caption}
\usepackage{algorithm}
\usepackage{algorithmic}
\usepackage{newfloat}
\usepackage{listings}
\usepackage{amsmath}
\usepackage{amssymb}
\usepackage{amsfonts}
\usepackage{amsopn}
\usepackage{epsfig}
\usepackage{microtype}
\usepackage{subfigure}
\usepackage{booktabs}
\usepackage{amsfonts}
\usepackage{nicefrac}
\usepackage{microtype}
\usepackage{xcolor}
\usepackage{stmaryrd}
\usepackage{lipsum}
\usepackage{rotating}
\usepackage{wasysym}
\usepackage{phonetic}
\usepackage{halloweenmath}
\usepackage{bm}
\usepackage{slashed}
\usepackage{amsmath,stackengine,scalerel}
\usepackage{bbm}
\usepackage{censor}
\usepackage[makeroom]{cancel}
\usepackage{mathtools}
\usepackage{amsthm}
\usepackage{tikz}
\usepackage{bigints}
\usepackage{comment}
\usepackage{accents}
\usepackage{trimclip}

\usepackage{url}

\usepackage{framed}
{\begin{leftbar}\begin{quotation}}%
{\end{quotation}\end{leftbar}}

\stackMath
\setstackgap{L}{0pt}
\def\dhat#1{\ThisStyle{\setbox0=\hbox{$\SavedStyle#1$}%
  \stackon{\SavedStyle#1}{\SavedStyle\hspace{.2\ht0}%
  \tilde{\vphantom{#1}}\kern\dimexpr2.2\LMpt+.7pt\relax\hat{\vphantom{#1}}}}%
}

\makeatletter
\DeclareRobustCommand{\shortto}{%
  \mathrel{\mathpalette\short@to\relax}%
}

\newcommand{\short@to}[2]{%
  \mkern2mu
  \clipbox{{.5\width} 0 0 0}{$\m@th#1\vphantom{+}{\shortrightarrow}$}%
  }
\makeatother

\newcommand{\tsp}[0]{{\rm T}}
\newcommand{\He}[0]{{H\!e}}
\newcommand{\Henum}[0]{{\rm He}}
\newcommand{\mindownarrow}{{{\downarrow}}}

\newcommand{\centau}{{\overline{\tau}}}
\newcommand{\abssig}{{{{\overline{s}}}}}
\newcommand{\abstau}{{{{\hat{\tau}}}}}
\newcommand{\absT}{{{{\hat{s}}}}} 

\newcommand{\roctau}{{{{{\rho}}}}}
\newcommand{\abstaumax}{{{{T}}}}
\newcommand{\rocT}{{{{{R}}}}} 
\newcommand{\absTmax}{{{{{S}}}}} 
\newcommand{\xpostm}{{\omega}}

\newcommand{\allofW}{{\Theta}}

\newcommand{\featarg}[2]{\left\lfloor\!{#1}\atop{#2}\!\right\rceil}
\newcommand{\leftbf}[0]{\left<}
\newcommand{\rightbf}[0]{\right]}

\newcommand{\rkhs}[1]{\mathcal{H}_{#1}}

\newcommand{\rkbs}[1]{\mathcal{B}_{#1}}

\newcommand{\bg}[1]{{\bm{#1}}}

\newcommand{\inver}[1]{{\tilde{#1}}}
\newcommand{\Inver}[1]{{\widetilde{#1}}}
\newcommand{\inveri}{{\widetilde{\imath}}}
\newcommand{\inverj}{{\widetilde{\jmath}}}
\newcommand{\prever}[1]{{\tilde{#1}}}
\newcommand{\Prever}[1]{{\widetilde{#1}}}
\newcommand{\postver}[1]{{#1}}
\newcommand{\Postver}[1]{{#1}}

\newcommand{\origin}{{\mathcal{O}}}
\newcommand{\changein}{{\Delta}}
\newcommand{\shadx}{{\mu}}
\newcommand{\shadb}{{\beta}}

\newcommand{\shada}{{\gamma}}

\makeatletter
\def\widebreve{\mathpalette\wide@breve}
\def\wide@breve#1#2{\sbox\z@{$#1#2$}%
     \mathop{\vbox{\m@th\ialign{##\crcr
\kern0.08em\brevefill#1{0.8\wd\z@}\crcr\noalign{\nointerlineskip}%
                    $\hss#1#2\hss$\crcr}}}\limits}
\def\brevefill#1#2{$\m@th\sbox\tw@{$#1($}%
  \hss\resizebox{#2}{\wd\tw@}{\rotatebox[origin=c]{90}{\upshape(}}\hss$}
\makeatletter

\DeclareFontFamily{U}{mathx}{}
\DeclareFontShape{U}{mathx}{m}{n}{<-> mathx10}{}
\DeclareSymbolFont{mathx}{U}{mathx}{m}{n}
\DeclareMathAccent{\widehat}{0}{mathx}{"70}
\DeclareMathAccent{\widecheck}{0}{mathx}{"71}

\DeclareMathOperator\sgn{sgn}

\theoremstyle{definition}
\newtheorem{def_rkbs}{Definition}
\newtheorem{def_rkhs}[def_rkbs]{Definition}

\newtheorem{def_datatree}[def_rkbs]{Definition}

\theoremstyle{remark}

\theoremstyle{plain}
\newtheorem{lem_densekey}{Lemma}

\newtheorem{th_netexpand}[lem_densekey]{Theorem}

\newtheorem{th_netexpand_changein}[lem_densekey]{Theorem}

\newtheorem{th_netexpand_kernel}[lem_densekey]{Theorem}
\newtheorem{th_radbounded}[lem_densekey]{Theorem}
\newtheorem{th_radbounded_lip}[lem_densekey]{Theorem}
\newtheorem{th_radbounded_local}[lem_densekey]{Theorem}

\newtheorem{th_itsbanach}[lem_densekey]{Theorem}
\newtheorem{th_itshilbert}[lem_densekey]{Theorem}

\everypar{\looseness=-1}

\title{Novel Kernel Models and Exact Representor Theory for Neural Networks Beyond the Over-Parameterized Regime}

\author{%
  Alistair Shilton \\
  Applied Artificial Intelligence Institute ${\rm (A}^2{\rm I}^2{\rm )}$ \\
  Deakin University, Geelong, Australia \\
  \texttt{alistair.shilton@deakin.edu.au} \\
  \And
  Sunil Gupta \\
  Applied Artificial Intelligence Institute ${\rm (A}^2{\rm I}^2{\rm )}$ \\
  Deakin University, Geelong, Australia \\
  \texttt{sunil.gupta@deakin.edu.au} \\
  \And
  Santu Rana \\
  Applied Artificial Intelligence Institute ${\rm (A}^2{\rm I}^2{\rm )}$ \\
  Deakin University, Geelong, Australia \\
  \texttt{santu.rana@deakin.edu.au} \\
  \And
  Svetha Venkatesh \\
  Applied Artificial Intelligence Institute ${\rm (A}^2{\rm I}^2{\rm )}$ \\
  Deakin University, Geelong, Australia \\
  \texttt{svetha.venkatesh@deakin.edu.au} \\
}

\begin{document}

\maketitle

\begin{abstract}
This paper presents two models of neural-networks and their training 
applicable to neural networks of arbitrary width, depth and topology, assuming 
only finite-energy neural activations; and a novel representor theory for 
neural networks in terms of a matrix-valued kernel.  The first model is exact 
(un-approximated) and global, casting the neural network as an elements in a 
reproducing kernel Banach space (RKBS); we use this model to provide tight 
bounds on Rademacher complexity.  The second model is exact and local, casting 
the change in neural network function resulting from a bounded change in 
weights and biases (ie. a training step) in reproducing kernel Hilbert space 
(RKHS) in terms of a local-intrinsic neural kernel (LiNK).  This local model 
provides insight into model adaptation through tight bounds on Rademacher 
complexity of network adaptation.  We also prove that the neural tangent 
kernel (NTK) is a first-order approximation of the LiNK kernel.  Finally, and 
noting that the LiNK does not provide a representor theory for technical 
reasons, we present an exact novel representor theory for layer-wise neural 
network training with unregularized gradient descent in terms of a 
local-extrinsic neural kernel (LeNK).  This representor theory gives insight 
into the role of higher-order statistics in neural network training and the 
effect of kernel evolution in neural-network kernel models.  Throughout the 
paper (a) feedforward ReLU networks and (b) residual networks (ResNet) are 
used as illustrative examples.
\end{abstract}

\section{Introduction} \label{sec:intro}

The application of reproducing kernel Hilbert space (RKHS \citep{Aro1}) and 
reproducing kernel Banach space (RKBS \citep{Lin10,Der1,Zha11,Zha14,Son1,Sri3, 
Xu4}) theory to the study of neural networks has a long history \citep{Nea1, 
Wei5,Par2,Lee8,Mat5,Rah2,Bac3,Bac4,Dan1,Dan2,Cho8,Bar9,Spe2}.  Neural tangent 
kernels (NTKs) are an exemplar of this approach, modeling training using a 
first-order (tangent) model.  This approach has led to a wide body of work on 
convergence and generalization \citep{Du1,All3,Du2,Zou1,Zou2,Aro4,Aro6,Cao4}, 
mostly focused on the wide-network (over-parameterized) limit.

Nevertheless, as noted in eg. \cite{Aro4,Lee9,Bai2}, there are limits to the 
descriptive powers of NTK models.  In particular, a gap has been observed 
between NTK-based predictions and actual performance \citep{Aro4,Lee9}, and 
the validity of NTK models naturally breaks down outside of the 
over-parameterized limit.  Recently this has led to attempts to generalize NTK 
models outside of the over-parameterized limit.  For example \citep{Bel2} use 
an exact pathwise kernel, \citep{Shi30} presented an exact model for dense 
feedforward neural networks in RKBS, \citep{Bar10,San9,Par2,Uns1,Uns2} have 
explored links to RKBS, and \citep{Bai2} explored higher-order approximations.  
However the assumptions made (smoothness, over-parameterisation etc) and 
approximations used limit application and raise the question of whether it is 
possible to instead formulate {\em exact, non-approximate} models for neural 
networks that may be used for similar ends.

In this paper we address this question by formulating two {\em exact} or 
{\em non-approximate} models for arbitrary neural networks, and an representor 
theory (also exact) for layerwise, feedforward networks:
\vspace{-\topsep}
\begin{enumerate}
\setlength\itemsep{0em}
 \item Exact global model: for an arbitrary neural network topology with arbitrary 
       weights and biases, subject to very mild assumptions, we construct an 
       RKBS model of the form ${\bf f} ({\bf x}; \allofW) = \leftbf 
       \Postver{\bg{\Psi}} ({\allofW}), \Postver{\bg{\phi}} ({\bf x}) 
       \rightbf_{\postver{\bf g}}$.  We use this model to place bounds on 
       Rademacher complexity.
 \item Exact local model: for the same network, this time assuming some initial 
       weights and biases, we construct an RKHS type model of the change in the 
       operation of the neural network given a bounded change in weights and 
       biases.  We prove that the NTK is a first-order approximation of the 
       kernel describing this RKHS and apply it to bounding the Rademacher 
       complexity of network adaptation.
 \item Exact representor theory: finally, we derive an exact representor 
       theory for the change in neural network operation due to a weight-step 
       in unregularized gradient descent training.  This RKHS model includes 
       several novel features, including (a) kernel warping, which is the 
       effect of model change for finite steps, (b) cross (off-diagonal) terms 
       describing the interaction between features in the non-over-parameterised 
       case, and (c) higher-order terms and interactions between training 
       vectors in the representor theory.  We show that the kernel for the 
       exact local model may be understood as an approximation of the kernel 
       used in the representor theory, and the NTK a more coarse approximation 
       still.
\end{enumerate}
\vspace{-\topsep}

We also discuss the role of network initialization for our model and its 
predictions, particularly in the wide-network limit, and give some results 
specific for our feedforward ReLU and ResNet examples.

\subsection{Notation}

We use 
$\mathbb{N} = \{ 0,1,2,\ldots \}$, 
$\mathbb{N}_n = \{ 0,1,2,\ldots,n-1 \}$, 
$\mathbb{Z}_+ = \{ 1,2,3,\ldots \}$.   
${\rm Span} (\mathcal{X})$ is the linear span of $\mathcal{X}$.  
Column vectors are ${\bf a}, {\bf b}, \ldots$ (elements $a_i, \ldots$).  
${\bf a}_{{\sim}i}$ is ${\bf a}$ with element $i$ removed. 
Matrices are ${\bf A}, {\bf B}, \ldots$ (elements $A_{i,i'}$, rows ${\bf A}_{i:}$, columns ${\bf A}_{:i'}$).  
${\rm diag}_i ({\bf A}_i)$ is a block-diagonal matrix with blocks ${\bf A}_i$.
Elementwise absolute and sign are $|{\bf a}|$ and $\sgn ({\bf a})$.  
$[a]_+ = \max\{a,0\}$.  
Hadamard and Kronecker products are ${\bf A} \odot {\bf B}$, ${\bf A} \otimes {\bf B}$.  
Column- (Khatri-Rao) and row-wise (face-splitting) products are ${\bf A} \otimes^\updownarrow {\bf B}$, ${\bf A} \otimes^\leftrightarrow {\bf B}$.  
Hadamard and Kronecker powers are ${\bf A}^{\odot r} = [A_{i,j}^r]_{i,j}$, ${\bf A}^{\otimes k} = {\bf A} \otimes {\bf A} \otimes \overset{k\;{\rm times}}{\ldots}{\otimes}{\bf A}$.  
The $n^{\rm th}$ derivative of $f : \mathbb{R} \to \mathbb{R}$ is $f^{(n)}$.
Inner and bilinear products are $\left< \cdot,\cdot \right>$, $\leftbf \cdot,\cdot \rightbf$. 
$\left< {\bf a}, {\bf a}' \right>_{\bf g} = \sum_i g_i a_i a'_i$ for ${\bf g} > {\bf 0}$, 
$\leftbf {\bf a}, {\bf a}' \rightbf_{\bf g} = \sum_i g_i a_i a'_i$ $\forall {\bf g}$, 
$\leftbf {\bf A}, {\bf a}' \rightbf_{\bf g} = [\leftbf {\bf A}_{:j}, {\bf a}' \rightbf_{\bf g}]_j$, 
$\leftbf {\bf A}, {\bf a}' \rightbf_{\bf G} = [\leftbf {\bf A}_{:j}, {\bf a}' \rightbf_{{\bf G}_{:j}}]_j$. 
By default $\left< {\bf a}, {\bf a}' \right> = \left< {\bf a}, {\bf a}' \right>_{\bf 1}$. 
$\| {\bf a} \|_2^2 = \sum_i |a_i|^2$ and 
$\| {\bf A} \|_2^2 = \sup_{{\bf x}:\|{\bf x}\|_2=1} \| {\bf A} {\bf x} \|_2^2$ 
are the Euclidean and spectral norms.  
The (probabilist's) Hermite polynomials \citep{Abr2,Mor2,Olv1,Cou1} are denoted $\He_k$ and form an orthogonal basis for $L^2 (\mathbb{R},e^{-x^2}) \!= \!\{\tau:\mathbb{R}\to\mathbb{R} \;|\int_{\infty}^{\infty} |\tau (\zeta)|^2 e^{-\zeta^2} d\zeta < \infty \}$.  
Hermite numbers are $\Henum_k = \He_k(0)$.  
\newline
{\bf Indexing:} $\inverj, j \in \mathbb{N}_D$ indexes node, 
$\inveri_j \in \mathbb{N}_{\Inver{H}^{[j]}}$, $i_j \in \mathbb{N}_{H^{[j]}}$ 
($\Inver{H}^{[j]}$, $H^{[j]}$ are the fan-in/out of node $j$).

\section{Setting and Assumptions} \label{sec:setup}

Consider neural networks defined by directed acyclic graphs (computational 
skeletons \citep{Dan1}) with $D$ nodes indexed by $j \in\mathbb{N}_D$, with 
output node $j=D-1$; a virtual input node $j=-1$; and directed edges $(\inverj 
\to j) \in (\mathbb{N}_D\cup\{-1\})^2$.  Node $j$ has antecedents $\inverj \in \Inver{\mathbb{P}}^{[j]} \subset \mathbb{N}_D \cup \{-1\}$, 
in-degree $\inver{p}^{[j]} = |\Inver{\mathbb{P}}^{[j]}|$, fan-out $H^{[j]}$, 
fan-in $\Inver{H}^{[j]} = \sum_{\inverj \in \Inver{\mathbb{P}}^{[j]}} 
H^{[\inverj]}$ ($H^{[-1]} = n$, $H^{[D-1]} = m$).  Overall:
\begin{equation}
{\!\!\!\!\!\!\!\!\!\!{
 \begin{array}{l}
 \left.
 \begin{array}{l}
  {\bf f} \left( {\bf x}; \allofW \right)
  =
  \postver{\bf x}^{[D-1]} \\
 \end{array}
 \right. 
 \;\;\;\;
 \left|
 \begin{array}{rl}
  \prever{\bf x}^{[j]}
  &\!\!\!\!=
  \big[ \prever{\bf x}^{[\inverj,j]} \big]_{\inverj \in \Inver{\mathbb{P}}^{[j]}},
  \prever{\bf x}^{[\inverj,j]}
  =
  \tau^{[\inverj,j]} (\postver{\bf x}^{[\inverj]}) \\
  \postver{\bf x}^{[j]} 
  &\!\!\!\!= 
  {\bf W}^{[j]\tsp}
  \prever{\bf x}^{[j]}
  +
  \shada {\bf b}^{[j]} \\
  &\!\!\!\!\!\!\!\!\!\;\left(= 
  \sum_{\inverj \in \Inver{\mathbb{P}}^{[j]}}
  {\bf W}^{[\inverj,j]\tsp}
  \prever{\bf x}^{[\inverj,j]}
  +
  \shada {\bf b}^{[j]} \right) \\
 \end{array}
 \right\}
  \;
 \begin{array}{l}
  \forall j \in \mathbb{N}_D, \inverj \in \Inver{\mathbb{P}}^{[j]} \\
  \postver{\bf x}^{[-1]} = {\bf x} \\
 \end{array}
 \end{array}
 \label{eq:yall_main}
}\!\!\!\!\!\!\!}
\end{equation}
where ${\bf W}^{[j]}\in \mathbb{R}^{\Inver{H}^{[j]}\times H^{[j]}}$, ${\bf 
W}^{[\inverj,j]}\in \mathbb{R}^{H^{[\inverj]}\times H^{[j]}}$ and ${\bf 
b}^{[j]} \in \mathbb{R}^{H^{[j]}}$ are weights and biases and 
$\tau^{[\inverj,j]} : \mathbb{R} \to \mathbb{R}$ the neural activation 
functions.  We write 
$\allofW = \{ {\bf W}^{[j]},{\bf b}^{[j]} : j \in \mathbb{N}_D \}$.  We assume 
$\exists \shadx^{[j]}, \shadb^{[j]} \in \mathbb{R}_+$ such that, with high 
probability $\geq 1-\epsilon$, simultaneously $\forall j\in\mathbb{N}_D$:
\vspace{-\topsep}
\begin{enumerate}
\setlength\itemsep{0em}
 \item Bounded inputs: 
       $\| {\bf x} \|_2 \leq 1$.
 \item Bounded weights and biases: 
       $\| {\bf W}^{[j]} \|_2 \leq \shadx^{[j]}$, $\| {\bf b}^{[j]} \|_2 \leq \shadb^{[j]}$.
 \item Bounded activations: 
       $\tau^{[\inverj,j]} \in L^2 (\mathbb{R}, e^{-\zeta^2})$.
\end{enumerate}
\vspace{-\topsep}
We denote the set of inputs satisfying these assumptions $\mathbb{X}$ and the 
set of weights and biases satisfying same $\mathbb{W}$, so ${\bf f}:\mathbb{X} 
\times \mathbb{W} \to \mathbb{R}^m$.  
Given a training set $\mathcal{D} = \{ ({\bf x}_{\{i\}},{\bf y}_{\{i\}}) \in 
\mathbb{R}^{n \times m}: i \in \mathbb{N}_N \}$ we assume the goal of training 
is to minimize the risk (for loss $L$, regularizer $r$, trade-off $\lambda \in 
\mathbb{R}_+$):
\begin{equation}
{\!\!\!\!\!\!\!\!{
 \begin{array}{l}
  \allofW^\star
  =
  \mathop{\rm argmin}\limits_{\allofW \in \mathbb{W}}
  \sum_i L \left( {\bf f} \left( {\bf x}_{\{i\}}; \allofW \right) - {\bf y}_{\{i\}} \right) + \lambda r \left( \allofW \right) \\
 \end{array}
}\!\!\!\!\!\!\!\!}
 \label{eq:traingoalW_main}
\end{equation}
The following examples are used throughout \citep{Glo1,He2}:
\vspace{-\topsep}
\begin{enumerate}
\setlength\itemsep{0em}
 \item Feedforward ReLU: fully connected, unbiased, feedforward, ReLU activations: 
       $\!\Inver{\mathbb{P}}^{[j]}\!=\!\{j-1\}$ $\forall j$; 
       $\shada = b_{i_j}^{[j]} = 0$ $\forall j$; 
       $\tau^{[j-1,j]} (\zeta) = [\zeta]_+$ $\forall j \ne 0,D-1$; 
       $\tau^{[-1,0]} (\zeta) = \tau^{[D-2,D-1]} (\zeta) = \zeta$.
 \item Residual Network (ResNet): unbiased, alternating ReLU/skip, 
       $D \in 2\mathbb{Z}_+$; 
       $\Inver{\mathbb{P}}^{[j]} = \{j-1\}$ $\forall j$ even; 
       $\Inver{\mathbb{P}}^{[j]} = \{j-1,j-2\}$, 
       ${\bf W}^{[j]} = [ {\bf W}_C^{[j]}; {\bf I} ]$ $\forall j$ odd; 
       $\shada = b_{i_j}^{[j]} = 0$ $\forall j$; 
       $\tau^{[j-1,j]} (\zeta) = [\zeta]_+$ $\forall j \ne 0,D-1$; 
       $\tau^{[j-2,j]} (\zeta) = \zeta$ $\forall j$ odd; 
       $\tau^{[-1,0]} (\zeta) = \tau^{[D-2,D-1]} (\zeta) = \zeta$.
\end{enumerate}
\vspace{-\topsep}

With regard to the weight-bound assumptions, if $W_{\inveri_j,i_j}^{[j]} \in \mathcal{N} 
(0,\sigma^{[j]2})$ then $\|{\bf W}_{:i_j}^{[j]}\|_2^2 
\sim \sigma^{[j]2} \chi_{\Inver{H}^{[j]}}^2$. 
So, w.h.p. $1-\epsilon$, 
$\|{\bf W}_{:i_j}^{[j]}\|_2^2 \leq \sigma^{[j]2} ( \Inver{H}^{[j]} + 2{\Inver{H}^{[j]\frac{1}{2}}}\ln ({DH^{[j]}}/{2\epsilon}) + 2\ln ({DH^{[j]}}/{\epsilon}))$ 
simultaneously $\forall j,i_j$ \citep[Lemma 1, pg 1325]{Lau2}.  
Given sufficiently strong regularization, 
a wide-network, or early stopping, network weights may be expected to 
remain close to their initialization, so the initialization can be used as 
a basis for norm-bounds.  For example:
\vspace{-\topsep}
\begin{enumerate}
\setlength\itemsep{0em}
 \item LeCun ($\sigma^{[j]2} = \frac{1}{H^{[j]}}$): 
       $\shadx^{[j]2} = \frac{\Inver{H}^{[j]}}{H^{[j]}} + \frac{2 \sqrt{\Inver{H}^{[j]}}}{H^{[j]}} \left(\ln\left(\frac{DH^{[j]}}{2\epsilon}\right) + \ln\left(\frac{DH^{[j]}}{\epsilon}\right)\right)$.
 \item He ($\sigma^{[j]2} = \frac{1}{\Inver{H}^{[j]}}$):
       $\shadx^{[j]2} = 1 + \frac{2}{\sqrt{\Inver{H}^{[j]}}} \left(\ln\left(\frac{DH^{[j]}}{2\epsilon}\right) + \ln\left(\frac{DH^{[j]}}{\epsilon}\right)\right)$.
 \item Glorot ($\sigma^{[j]2} = \frac{1}{H^{[j]}+\Inver{H}^{[j]}}$): 
       $\shadx^{[j]2} = \frac{\Inver{H}^{[j]}}{H^{[j]}+\Inver{H}^{[j]}} + \frac{2 \sqrt{\Inver{H}^{[j]}}}{{H^{[j]}}+{\Inver{H}^{[j]}}} \left(\ln\left(\frac{DH^{[j]}}{2\epsilon}\right) + \ln\left(\frac{DH^{[j]}}{\epsilon}\right)\right)$.
\end{enumerate}
The norm-bound for LeCun initialization varies with fan-in/out 
and may be arbitrarily large.  The norm-bound for He initialization is 
better behaved with respect to fan-in but may be arbitrarily large depending on fan-out.  
In contrast, the norm bound for Glorot initialization may be finitely bounded 
by $\shadx^{[j]2} \leq 1 + {({\frac{5}{3}}({{\epsilon}/{D}})^{{1}/{2}} + 
{({{D}/{\epsilon}})^{{1}/{2}}}\ln 2)} \sqrt{2} e^{-{2\epsilon}/{3D}}$, 
independent of both fan-in and fan-out, which is most helpful in our context 
as it is well-behaved in the wide-network limit.

\subsection{Related Work - NNGP, NTK and Beyond} \label{sec:relwork}

The use of kernel methods to model neural networks dates to at least 
\citep{Nea1}.  Fixing all weights and biases except for node $j$ and defining 
feature map ${\bg{\varphi}}^{[j]}({\bf x}) = [\prever{\bf x}^{[j]};\shada]$, 
the network can be written as ${\bf f} ({\bf x}; \allofW) = {\bf q}^{[j]} 
([{\bf W}^{[j]\tsp}; {\bf b}^{[j]}] {\bg{\varphi}}^{[j]} ({\bf x}), {\bf x})$ 
for fixed ${\bf q}^{[j]}$, and (\ref{eq:traingoalW_main}) becomes kernel 
regression with:
\begin{equation}
 \begin{array}{l}
 \!\!\mbox{NNGP Kernel:}
 \begin{array}{l}
  \Sigma^{[j]} \left( {\bf x}, {\bf x}' \right)
  =
  \mathbb{E}_k \left[ \varphi_k^{[j]} \left( {\bf x} \right) \varphi_k^{[j]} \left( {\bf x}' \right) \right]
  =
  \mathbb{E}_{\inver{j} \in \Inver{\mathbb{P}}^{[j]}} \left[ \Sigma^{[\inverj,j]} \left( {\bf x}, {\bf x}' \right) \right] \\

  \Sigma^{[\inverj,j]} \left( {\bf x}, {\bf x}' \right)
  =
  \gamma^2
  +
  \mathbb{E}_{i_\inverj} \left[ \prever{x}_{i_{\inverj}}^{[\inverj,j]} \prever{x}_{i_\inverj}^{[\inverj,j]} \right] \\
 \end{array}
 \end{array}
\label{eq:basic_nngp}
\end{equation}
In the infinitely wide limit, for suitable initialization, (\ref{eq:basic_nngp}) 
is deterministic (dependent on the distribution $\allofW \sim \nu$), 
and it can be demonstrated that $\postver{\bf x}^{[j]} (\cdot) \sim {\rm GP} (0, 
\Sigma^{[j]})$ \citep{Nea1,Lee8,Mat5,Gar3,Nov2}.  This is the NNGP model, and 
may be used to eg. derive insights into the types of function the network is 
best suited to model.  The NNGP kernel for our ReLU example is the arc-cosine 
kernel \citep{Cho8}.

To model network training neural tangent kernels (NTKs \citep{Jac2,Aro4}) are 
a first-order model approximating the behavior of a neural network as the 
weights and biases vary about their initialization $\allofW$, i.e. ${\bf 
f} ({\bf x}; \allofW + \changein \allofW) \approx {\bf f} ({\bf x}; 
\allofW) + \changein \allofW^\tsp \nabla_\allofW {\bf f} ({\bf 
x}; \allofW)$.  Training is cast as kernel regression, with the neural 
tangent kernel (NTK):
\begin{equation}
{\!\!\!\!\!\!\!\!{
 \begin{array}{l}
 \!\!\mbox{NTK:}
 \begin{array}{l}
  \left. \begin{array}{l}
  K_{\scriptscriptstyle{\rm NTK}} \left( {\bf x}, {\bf x}' \right)
  =
  \mathbb{E}_k \left[ \nabla_{\allofW_k} {\bf f} ({\bf x}; \allofW)^\tsp \nabla_{\allofW_k} {\bf f} ({\bf x}; \allofW) \right]
  =
  K_{\scriptscriptstyle{\rm NTK}}^{[D-1]} \left( {\bf x}, {\bf x}' \right)
  \end{array} \right. \\
  \left. \begin{array}{rl}
  K_{\scriptscriptstyle{\rm NTK}}^{[j]} \left( {\bf x}, {\bf x}' \right)
  &\!\!\!\!=
  \Sigma^{[\inverj]} \left( {\bf x}, {\bf x}' \right)
  +
  \mathbb{E}_{\inverj \in \Inver{\mathbb{P}}^{[j]}} \left[ \theta^{[\inverj,j]} \left( {\bf x}, {\bf x}' \right) K_{\scriptscriptstyle{\rm NTK}}^{[\inverj]} \left( {\bf x}, {\bf x}' \right) \right] \\
  \theta^{[\inverj,j]} \left( {\bf x}, {\bf x}' \right)
  &\!\!\!\!=
  \mathbb{E}_{i_\inverj} \left[ \left< \tau^{[\inverj,j](1)} \left( \postver{x}_{i_\inverj}^{[\inverj]} \right) \tau^{[\inverj,j](1)} \left( \postver{x}_{i_\inverj}^{\prime[\inverj]} \right) \right> \right] \; \forall \inverj \in \Inver{\mathbb{P}}^{[j]} \\
 \end{array} \right\} \; \forall j \\
 \end{array}
 \end{array}
}\!\!\!\!\!\!\!\!}
\end{equation}
and $K_{\scriptscriptstyle{\rm NTK}}^{[-1]} ({\bf x}, {\bf x}') = 0$.  The NTK model is accurate 
for small variations in weights/biases.  In the infinitely wide limit the NTK 
is deterministic, and weights/biases remain close to their initial values, 
leading to the gradient flow model where weights flow rather than change in 
discrete steps.  This approach gives insight in areas including convergence 
and generalization \citep{Du1,All3,Du2,Zou1,Zou2,Aro4,Aro6,Cao4}.

Beyond this first-order model, while NTKs have made significant progress, a 
gap has been observed between NTK-based predictions and actual performance 
\citep{Aro4,Lee9}.  One approach to bridging this gap is to construct 
higher-order or exact models.  Works in this direction include \citep{Bai2}, 
which presented a higher-order approximation; \citep{Bel2}, which used a 
pathwise kernel; and \citep{Shi30}, which used an RKBS model.\footnote{In a 
similar vein, \citep{Bar10,San9,Par2,Uns1,Uns2} explore the link to RKBS 
theory, though excepting \citep{Uns2} they only consider shallow networks.}

\section{Hermite Representation of Neural Activations} \label{sec:hermiteI}

The local and global models given here are based on a Hermite transform of the 
activations.  This transform of $\tau^{[\inverj,j]}$ is constructed around a 
{\em center} $\xi$.  We begin by defining the centered activations:
\[
 \begin{array}{c}
 \mbox{Centered activation:}
 \begin{array}{c}
  \centau^{[\inverj,j]} \left( \zeta ; \xi \right)
  = 
  \tau^{[\inverj,j]} \left( \xi + \zeta \right) - \tau^{[\inverj,j]} \left( \xi \right)
 \end{array}
 \end{array}
\]
By assumption $\tau^{[\inverj,j]} \in L^2 (\mathbb{R}, e^{-\zeta^2})$, so 
$\centau^{[j',j]} (\cdot,\xi) \in L^2 (\mathbb{R}, e^{-\zeta^2})$ and the 
Hermite transform \citep{Abr2,Mor2,Olv1} exists.  Precisely:
\begin{equation}
{\!\!\!\!\!\!\!\!\!\!{
 \begin{array}{l}
 \begin{array}{rl}
  \centau^{[\inverj,j]} \left( \zeta ; \xi \right)
  &\!\!\!\!=
  \sum_{k \geq 0} a^{[\inverj,j]}_{\left( \xi \right)k} \He_k \left( \zeta \right) 
  =
  \sum_{k \geq 1} a_{\left( \xi \right)k}^{[\inverj,j]} 
  \sum_{l=1}^k \binom{k}{l}
  \Henum_{k-l} \zeta^l \\
  \end{array} \\
  \begin{array}{l}
  \mbox{where: } a_{\left( \xi \right)k}^{[\inverj,j]} 
  =
  \frac{1}{\sqrt{2\pi} k!} \int_{-\infty}^\infty \centau^{[\inverj,j]} \left( \zeta; \xi \right) \He_k \left( \zeta \right) e^{-{\zeta^2}/{2}} d\zeta 
  \end{array} \\
  \end{array}
}\!\!\!\!\!\!}
\label{eq:tauherm_a}
\end{equation}
where $\Henum_k = \He_k (0)$ are the Hermite numbers (in the 
second form of $\centau^{[\inverj,j]}$ we use the additivity properties 
of the Hermite polynomials (supplementary \ref{sec:hermite})).  
We define the {\em 
magnitude functions}:
\begin{equation}
 \begin{array}{c}
 \mbox{Magnitude function:}
 \begin{array}{rl}
  \abssig^{[\inverj,j]} \left( \zeta \right)
  &\!\!\!\!=
  \sum_{k \geq 0} \left| a_{(0) k}^{[\inverj,j]} \right| \left( 1+\zeta \right)^k
  -
  \sum_{k \geq 0} \left| a_{(0) k}^{[\inverj,j]} \right| \\
 \end{array}
 \end{array}
\label{eq:defmagfn}
\end{equation}
for the global model. 
For the local model we have the {\em rectified activation functions} and 
their envelopes:
\begin{equation}
{\!\!\!\!\!\!\!\!\!\!\!\!\!\!\!\!{
 \begin{array}{rl}
  \mbox{Rectified activation function:} &
  \abstau_\eta^{[\inverj,j]} \left( \zeta ; \xi, \xi' \right)
  =
  \sum_{k \geq 1}
  \frac{a_{(\xi)k}^{[\inverj,j]} a_{(\xi')k}^{[\inverj,j]}}{\eta^{2k}}
  \sum_{l=1}^k
  {\binom{k}{l}}^2 \Henum_{k-l}^2 \zeta^{l} \\

  \mbox{Envelope:} &
  \abstau_\eta^{[\inverj,j]} \left( \zeta; \xpostm \right) 
  = \mathop{\rm sup}\limits_{|\xi|, |\xi'| \leq \xpostm} \abstau_\eta^{[\inverj,j]} \left( \zeta; \xi, \xi' \right) 
  = \mathop{\rm sup}\limits_{|\xi| \leq \xpostm} \abstau_\eta^{[\inverj,j]} \left( \zeta; \xi, \xi \right) \\ 
 \end{array}
\label{eq:rectactfn}
}\!\!\!\!\!\!\!\!}
\end{equation}
where $\xi, \xi' \in [-\xpostm,\xpostm]$ are the centers of the rectified 
activation functions and $\eta \in (0,1)$ is fixed.  
The magnitude 
functions converge everywhere, while the rectified activations have a 
finite ROC $|\abstau_\eta^{[\inverj,j]} (\zeta;\xi,\xi')| \leq \abstaumax_{(\xi, 
\xi')\eta}^{[\inverj,j]2}$ $\forall |\zeta| \leq \roctau_{(\xi,\xi')\eta}^{[\inverj, 
j]2}$, and likewise $|\abstau_\eta^{[\inverj,j]} (\zeta;\xpostm)| \leq 
\abstaumax_{(\xpostm)\eta}^{[\inverj,j]2}$ $\forall |\zeta| \leq 
\roctau_{(\xpostm)\eta}^{[\inverj,j]2}$ - see supplementary \ref{sec:raf_fun}.  
The magnitude function and rectified activation envelope 
are origin crossing, monotonically 
increasing and superadditive 
on $\mathbb{R}_+$.  We also define (supplementary \ref{sec:raf_fun}):
\[
 \begin{array}{rl}
  \absT_\eta \left( \zeta \right)
  &\!\!\!\!=
  {\sum}_{k \geq 1}
  \eta^{2k}
  {\sum}_{1 \leq l \leq k}
  \zeta^l
  =
  \frac{\zeta}{1-\zeta}
  \left(
  \frac{\eta^2}{1-\eta^2}
  -
  \frac{\zeta\eta^2}{1-\zeta\eta^2}
  \right)
 \end{array}
\]
which converges as given $\forall |\zeta| < \rocT_\eta^{2} < 1$, whereon 
$\absTmax_\eta^{2} = \frac{\rocT_\eta^2}{1-\rocT_\eta^2} 
(\frac{\eta^2}{1-\eta^2}-\frac{\eta^2\rocT_\eta^2}{1-\eta^2\rocT_\eta^2})$.

For linear activations $\tau (\zeta) = \zeta$ then $a_{(\xi)k} = \delta_{k,0}$ 
and $\abssig (\zeta) = \zeta$, $\abstau_\eta (\zeta;\xi,\xi') = \zeta/\eta$.  For 
ReLU activations $\tau (\zeta) = [\zeta]_+$, as shown in supplementary 
\ref{sec:hermxformact} (where the case $\forall\xi\in\mathbb{R}$ is also given):
\begin{equation}
 \begin{array}{rlrl}
  a_{\left( 0 \right)k} &\!\!\!\!= \left\{ \begin{array}{ll}
  \frac{\left( -1 \right)^{p+1}}{\sqrt{2\pi} (2p-1) 2^p p!} & \mbox{if } k = 2p, p \in \mathbb{Z}_+ \\
  \frac{1}{2} \delta_{k,1} & \mbox{otherwise} \\
  \end{array} \right.
 \end{array}
\end{equation}
\begin{equation}
{\!\!\!\!\!\!\!\!\!\!\!{
 \begin{array}{rl}
  \abssig \left( \zeta \right) &\!\!\!\!= 
  \frac{1}{2} \zeta \left( {\rm erfi} \left( \frac{1+\zeta}{\sqrt{2}} \right) + 1 \right) +
  \frac{1}{\sqrt{2\pi}} \Big( e^{\frac{1}{2} } -
  e^{\frac{1}{2} \left( 1+\zeta \right)^2} \Big) +
  \frac{1}{2}   \left( {\rm erfi} \left( \frac{1+\zeta}{\sqrt{2}} \right) - {\rm erfi} \left( \frac{1}{\sqrt{2}} \right) \right) \\
 \end{array}
}\!\!\!\!\!\!\!\!}
 \label{eq:abssig_relu}
\end{equation}
It is difficult to obtain a closed-form expression for the rectified activation 
or its envelope for the ReLU, but they are relatively straightforward to 
calculate, as are their convergence bounds.  Figure \ref{fig:rectactrelu} in 
supplementary \ref{sec:hermxformact} shows a sample of various rectified activations 
for the ReLU with different centers.

\section{Global Dual Model in Reproducing Kernel Banach Space} \label{sec:globdual}

In this section we derive a dual model for the network described, where by 
global we mean not constructed about some weight initialization.  Our 
derivation is similar to \citep{Shi30}, but based on a Hermite polynomial expansion 
rather than a Taylor series, making it applicable to a wider range of activation 
functions with fewer caveats.  Our key result for this section is:
\begin{th_netexpand}
 Let ${\bf f} : \mathbb{X} \times \mathbb{W} \to \mathbb{R}^m$ be a neural 
 network (\ref{eq:yall_main}) satisfying our assumptions.  Then:
 \begin{equation}
  \begin{array}{rl}
   {\bf f} \left( {\bf x}; \allofW \right) 
   &\!\!\!\!= \leftbf \Postver{\bg{\Psi}} \left( {\allofW} \right), \Postver{\bg{\phi}} \left( {\bf x} \right) \rightbf_{\postver{\bf g}} \\
  \end{array}
 \label{eq:ourdualmodel}
 \end{equation}
 with feature maps 
 $\Postver{\bg{\Psi}} : \mathbb{W} \to \Postver{\mathcal{W}} = \overline{\rm span} (\Postver{\bg{\Psi}} (\mathbb{W}))$ 
 and
 $\Postver{\bg{\phi}} : \mathbb{X} \to \Postver{\mathcal{X}} = \overline{\rm span} (\Postver{\bg{\phi}} (\mathbb{X}))$ 
 and metric $\postver{\bf g}$ defined in Figure \ref{fig:featmapdef}, where 
 $\| \Postver{\bg{\Psi}} (\allofW) \|_2 \leq \postver{\psi}$ 
 and 
 $\postver{\phi}_{\mindownarrow} \leq \| \Postver{\bg{\phi}} ({\bf x}) \|_2 \leq \postver{\phi}$ 
 $\forall \allofW \in \mathbb{W}, {\bf x} \in \mathbb{X}$.  Moreover:
 \begin{equation}
  {\!\!\!\!\!\!\!\!\!\!\!\!{
  \begin{array}{l}
   \left\| {\bf f} \left( {\bf x}; \allofW \right) \right\|_2
   \leq
   \left\| \Postver{\bg{\Psi}} \left( \allofW \right) \right\|_{\Henum[\tau]}
   \left\| \Postver{\bg{\phi}} \left( {\bf x} \right) \right\|_{2}, \;\;

   \left\| \Postver{\bg{\Psi}} \right\|_{\Henum[\tau]}^2 
   = \mathop{\rm sup}\limits_{{\bf x} \in \mathbb{X}} 
   \frac{\left\| \leftbf \Postver{\bg{\Psi}}, \Postver{\bg{\phi}} \left( {\bf x} \right) \rightbf_{\postver{\bf g}} \right\|_2^2}{\left\| \Postver{\bg{\phi}} \left( {\bf x} \right) \right\|_2}
  \end{array}
 }\!\!\!\!\!\!\!\!\!\!\!\!}
 \label{eq:ourcs_cont}
 \end{equation}
 where $\| \Postver{\bg{\Psi}} (\allofW) \|_{\Henum [\tau]} 
 \leq \undertilde{\postver{\psi}}$ $\forall \allofW \in 
 \mathbb{W}$, as per definitions in Figure \ref{fig:featmapdef}.
\label{th:netexpand}
\end{th_netexpand}
See supplementary \ref{append:dualform} for a proof of this theorem.  
Intuitively, this result may be derived recursively, starting from the input 
node and progressing to the output, using the Hermite expansion of the 
activation for the edges.  The operator-norm based bound (\ref{eq:ourcs_cont}) 
is included here due to the fact that the indefinite metric prevents us from 
naively bound $\|{\bf f}({\bf x};\allofW)\|_2$ in terms of $\postver{\phi} 
\postver{\psi}$ using the Cauchy-Schwarz inequality (as may be required e.g. 
when bounding Rademacher complexity).

\subsection{Implication: Neural Networks in Reproducing Kernel Banach Space} \label{sec:weightspace}

A reproducing kernel Banach space is defined as:
\begin{def_rkbs}[Reproducing kernel Banach space (RKBS)]
 A reproducing kernel Banach space on a set $\mathbb{X}$ is a Banach space 
 $\mathcal{F}$ of functions ${\bf f} : \mathbb{X} \to \mathbb{Y}$ for which the 
 point evaluation functionals ${\bg{\delta}}_{\bf x} ({\bf f}) = {\bf f}({\bf x})$ on 
 $\mathcal{F}$ are continuous (i.e. $\forall {\bf x} \in \mathbb{X}$ $\exists 
 C_{\bf x} \in \mathbb{R}_+$ such that $\|{\bg{\delta}}_{\bf x} ({\bf f})\|_2 \leq C_{\bf x} 
 \|{\bf f}\|_\mathcal{F}$ $\forall {\bf f} \in \mathcal{F}$).
\end{def_rkbs}
This definition is somewhat generic, so \citep{Lin10}\footnote{See e.g. 
\citep{Der1,Lin10,Zha11,Zha14,Son1,Sri3,Xu4} for other perspectives.} study 
the special case:
\begin{equation}
{\!\!\!\!\!\!\!\!{
 \begin{array}{c}
  \rkbs{} = \left\{ \left. {\bf f} \left( \cdot; \allofW \right)
  =
  \leftbf \Postver{\bg{\Psi}} \left( \allofW \right), \Postver{\bg{\Phi}} \left( \cdot \right) \rightbf_{\mathcal{W} \times \mathcal{X}} \right| \allofW \in \mathbb{W} \right\} \\
 \end{array}
}\!\!\!\!\!\!\!\!}
\label{eq:rkbs_recipe}
\end{equation}
where ${\bg{\Phi}} : \mathbb{X} \to \mathcal{X}$ is a data feature map, 
${\bg{\Psi}} : \mathbb{W} \to \mathcal{W}$ is a weight feature map, 
$\mathcal{X}$ and $\mathcal{W}$ are 
Banach spaces, and $\leftbf\cdot,\cdot\rightbf_{\mathcal{W}\times\mathcal{X}} 
: \mathcal{W} \times \mathcal{X} \to \mathbb{R}^m$ is continuous. 
The following result follows from theorem 
\ref{th:netexpand}:\footnote{Note that the RKBS defined in theorem \ref{th:itsbanach} is 
non-reflexive, which appears to rule out a trivial representor theory based 
on this dual in the global case \citep{Lin10}.}
\begin{th_itsbanach}
 The set 
 $\mathcal{F} = \{ {\bf f} (\cdot; \allofW) : \mathbb{R}^n \to \mathbb{R}^m | \allofW \in \mathbb{W} \}$ 
 of networks (\ref{eq:yall_main}) satisfying our assumptions forms 
 a RKBS of form (\ref{eq:rkbs_recipe}), where 
 $\| {\bf f} (\cdot;\allofW) \|_\mathcal{F} = \| \Postver{\bg{\Psi}} (\allofW) \|_{\Henum[\tau]} \leq {{\undertilde{\postver{\psi}}}}$ 
 and 
 $C_{\bf x} = \| \Postver{\bg{\phi}} ({\bf x}) \|_2 \leq \postver{\phi}$. 
 \label{th:itsbanach}
\end{th_itsbanach}

\subsection{Application: Bounding Rademacher Complexity for Neural Network Training}

The global dual formulation may be used to bound Rademacher complexity, which 
in turn bounds the uniform convergence properties of the network class 
\citep{Bar1,Ste3}. 
Assuming ${\bf x} \sim \nu$ the Rademacher complexity is defined as 
$\mathcal{R}_N (\mathcal{F}) = \mathbb{E}_{\nu} \mathbb{E}_{\epsilon} [{\rm sup}_{f \in \mathcal{F}} \frac{1}{N} \sum_{i \in \mathbb{N}_N} \epsilon_i f ({\bf x}_i)]$ 
for Rademacher random variables $\epsilon_i \in \{\pm 1\}$.  We have:
\begin{th_radbounded}
 The set $\mathcal{F} = \{ f (\cdot; \allofW) : \mathbb{R}^n\to\mathbb{R} 
 | \allofW\in\mathbb{W} \}$ of networks (\ref{eq:yall_main}) 
 satisfying our assumptions has Rademacher complexity bounded by 
 $\mathcal{R}_N (\mathcal{F}) \leq \frac{1}{\sqrt{N}} {\postver{\phi}} \undertilde{\postver{\psi}}$ 
 (definitions as per Figure \ref{fig:featmapdef}).
 \label{th:radbounded}
\end{th_radbounded}
The proof follows the usual template (see e.g. \citep{Bar1}) using our feature 
map, with (\ref{eq:ourcs_cont}) used instead of the Cauchy-Schwarz inequality 
(supplementary \ref{supp:radbound}).  If 
$\shadx^{[j]}$, $\shadb^{[j]}$ are width independent 
(eg. after Glorot initialization) then so too is the bound.  For Lipschitz 
activations we have:
\begin{th_radbounded_lip}
 Let $\mathcal{F} = \{ f (\cdot; \allofW) : \mathbb{R}^n\to\mathbb{R} 
 | \allofW\in\mathbb{W} \}$ be the set of unbiased networks 
 (\ref{eq:yall_main}) with $L$-Lipschitz activations satisfying our 
 assumptions.  Then 
 $\mathcal{R}_N (\mathcal{F}) \leq \max_{\mathcal{S} \in \mathbb{S}} 
 \prod_{j \in \mathcal{S}} L^2 \inver{p}^{[j]} \shadx^{[j]}$, where $\mathbb{S}$ 
 is the set of all input-output paths in the network graph.
 \label{th:radbounded_lip}
\end{th_radbounded_lip}
We observe that this bound applies to both our ReLU and ResNet examples, as 
both linear and ReLU activations are $1$-Lipschitz.  Indeed, for a ReLU 
network, assuming Glorot initialization and sufficiently strong (spectral or 
equivalent) regularization we have $\shadx^{[j]} \approx 1$, so 
$\mathcal{R}_N (\mathcal{F}) \lessapprox {1}/{\sqrt{N}}$.

\begin{figure*}
\begin{center}
\resizebox{\textwidth}{!}{%
$
\begin{array}{l|l}
\hspace{3.5cm}\text{\bf Incoming Edge Feature Map} & \hspace{3.5cm}\text{\bf Node Feature Map}_{{\;\!}_{{\;\!}_{{\;\!}_{\;\!}}}} \\
{\!\!\!{
 \rotatebox[]{90}{\text{\bf Feature Maps}}
 \begin{array}{rl}
  \Prever{\bg{\phi}}^{[\inverj,j]} \left( {\bf x} \right)
  &\!\!\!\!\!=\! \left[
             a_{(0)k}^{[\inverj,j]\left<\frac{1}{2}\right>}\!
             \left[
             {\binom{k}{l}}^{\frac{1}{2}}\!
             \left(
             \sqrt{\frac{\abssig^{[\inverj,j]-1} \left( \prever{\phi}^{[\inverj,j]2} \right)}{\shada^2+1}}
             \Postver{\bg{\phi}}^{[\inverj]} \left( {\bf x} \right)
             \right)^{\!\otimes l}
             \right]_{\!1 \leq l \leq k}
             \right]_{\!k \geq 1} \\

  \Prever{\bg{\Psi}}_{:i_{\inverj}}^{[\inverj,j]} \left( {\allofW} \right)
  &\!\!\!\!\!=\! \left[
             \left| a_{(0)k}^{[\inverj,j]} \right|^{\frac{1}{2}}\!
             \left[
             {\binom{k}{l}}^{\frac{1}{2}}\!
             \left(
             \sqrt{\frac{\shada^2+1}{\abssig^{[\inverj,j]-1} \left( \prever{\phi}^{[\inverj,j]2} \right)}}
             \Postver{\bg{\Psi}}_{:i_{\inverj}}^{[\inverj]} \left( {\allofW} \right)
             \right)^{\!\otimes l}
             \right]_{\!1 \leq l \leq k}
             \right]_{\!k \geq 1} \\

  \prever{\bf g}^{[\inverj,j]}
  &\!\!\!\!\!=\! \left[
             \left[
             \Henum_{k-l}
             \postver{\bf g}^{[\inverj]\otimes l}
             \right]_{1 \leq l \leq k}
             \right]_{k \geq 1}
             \hfill \forall \inverj \in \Inver{\mathbb{P}}^{[j]} \;\;\;\;\;\;\;\; \\
 \end{array}
}\!\!\!}
&
{\!\!\!{
 \begin{array}{rl}
  \Postver{\bg{\phi}}^{[j]} \left( {\bf x} \right)
  &\!\!\!\!\!=\! \left[ \begin{array}{c}
                    \shada \\
                    \frac{1}{\sqrt{\inver{p}^{[j]}}}
                    \left[ \frac{1}{\prever{\phi}^{[\inverj,j]}} {\Prever{\bg{\phi}}}^{[\inverj,j]} \left( {\bf x} \right) \right]_{\inverj \in \Inver{\mathbb{P}}^{[j]}} \\
             \end{array} \right] \;\;\;\;\;\;\;\;\;\;\;\;\;\;\;\;\;\;\;\;\;\; \\

  \Postver{\bg{\Psi}}^{[j]} \left( {\allofW} \right)
  &\!\!\!\!\!=\! \left[ \begin{array}{c}
                    {\bf b}^{[j]\tsp} + {\bg{\upsilon}}_\tau^{[j]\tsp} \\
                    \sqrt{\inver{p}^{[j]}}
                    \mathop{\rm diag}_{{{\inverj \in \Inver{\mathbb{P}}^{[j]}}}} \left( \prever{\phi}^{[\inverj,j]} \Prever{\bg{\Psi}}^{[\inverj,j]} \left( {\allofW} \right) \right) {\bf W}^{[j]} \\
             \end{array} \right] \\

  \postver{\bf g}^{[j]}
  &\!\!\!\!\!=\! \left[ \begin{array}{c}
                    1 \\
                    \left[ \prever{\bf g}^{[\inverj,j]} \right]_{\inverj \in \Inver{\mathbb{P}}^{[j]}} \\
             \end{array} \right]_{{\;}_{{{\;}_{\;}}}}
  \Big({\bg{\upsilon}}_\tau^{[j]} = \mathop{\sum}\limits_{\inverj \in \Inver{\mathbb{P}}^{[j]}} \frac{\tau^{[\inverj,j]} (0)}{\shada} {\bf W}^{[\inverj,j]\tsp} {\bf 1}_{H^{[\inverj]}} \Big) \;\;
 \end{array}
}\!\!\!}
\\
\hline
{\!\!\!{
 \rotatebox[]{90}{\text{\bf Norm Bounds}}
 \begin{array}{rl}
  \left\| \Prever{\bg{\phi}}^{[\inverj,j]} \left( {\bf x} \right) \right\|_2^2
  &\!\!\!\!\!\in\!
  \left[
  \prever{\phi}_{\mindownarrow}^{[\inverj,j]2} \!=\! {\abssig}^{[\inverj,j]} \left( \frac{{\abssig}^{[\inverj,j]-1} \left( \prever{\phi}^{[\inverj,j]2} \right)}{\shada^2+1} \postver{\phi}_{\mindownarrow}^{[\inverj]2} \right),
  \prever{\phi}^{[\inverj,j]2}
  \right]^{{\;\!\!}^{{\;\!\!}^{\;\!\!}}} \\

  \left\| \Prever{\bg{\Psi}}^{[\inverj,j]} \left( {\allofW} \right) \right\|_2^2
  &\!\!\!\!\!\leq\!
  \prever{\psi}^{[\inverj,j]2} \!=\! {\abssig}^{[\inverj,j]} \left( \frac{\shada^2+1}{{\abssig}^{[\inverj,j]-1} \left( \prever{\phi}^{[\inverj,j]2} \right)} \postver{\psi}^{[\inverj]2} \right) \\

  \left\| \Prever{\bg{\Psi}}^{[\inverj,j]} \left( {\allofW} \right) \right\|_{\Henum[\tau]}^2
  &\!\!\!\!\!\leq\!
  \undertilde{\prever{\psi}}^{[\inverj,j]2} \!=\! 
              {\!\begin{array}{c}
              \!\!\!\!\mathop{\sup}\limits_{{{\postver{\phi}_{\mindownarrow}^{[\inverj]} \leq \postver{\phi}^{[\inverj]} \leq \postver{\phi}^{[\inverj]}}\atop{-\undertilde{\postver{\psi}}^{[\inverj]} \leq \undertilde{\postver{\psi}}^{[\inverj]} \leq \undertilde{\postver{\psi}}^{[\inverj]} }}}
              \end{array}\!}
              \left\{
              \frac{
              \tau^{[\inverj,j]} \left( \postver{\phi}^{[\inverj]} \undertilde{\postver{\psi}}^{[\inverj]} \right)^2
              }{
              {\abssig}^{[\inverj,j]} \left( \frac{{\abssig}^{[\inverj,j]-1} \left( \prever{\phi}^{[\inverj,j]2} \right)}{\shada^2+1} \postver{\phi}^{[\inverj]2} \right)
              }
              \right\} \\
 \end{array}
}\!\!\!}
&
{\!\!\!{
 \begin{array}{rl}
  \left\| \Postver{\bg{\phi}}^{[j]} \left( {\bf x} \right) \right\|_2^2
  &\!\!\!\!\!\in\!
  \left[ 
  \postver{\phi}_{\mindownarrow}^{[j]2} \!=\! \shada^2+\frac{1}{\inver{p}^{[j]}} \mathop{\sum}\limits_{\inverj\in\Inver{\mathbb{P}}^{[j]}} \frac{\prever{\phi}_{\mindownarrow}^{[\inverj,j]2}}{\prever{\phi}^{[\inverj,j]2}},
  \postver{\phi}^{[j]2} \!=\! \shada^2+1
  \right] \\

  \left\| \Postver{\bg{\Psi}}^{[j]} \left( {\allofW} \right) \right\|_2^2
  &\!\!\!\!\!\leq\!
  \postver{\psi}^{[j]2}
  \!=\!
  \left( \shadb^{[j]} \!+\! \frac{\shadx^{[j]} \left| \tau^{[\inverj,j]} (0) \right|}{\shada} \right)^2
  \!\!+\!
  \inver{p}^{[j]}
  \shadx^{[j]2}
  \prever{\phi}^{[j]2} \prever{\psi}^{[j]2} \\

  \left\| \Postver{\bg{\Psi}}^{[j]} \left( {\allofW} \right) \right\|_{\Henum[\tau]}^2
  &\!\!\!\!\!\leq\!
  \undertilde{\postver{\psi}}^{[j]2}
  \!=\!
  \left( \shadb^{[j]} \!+\! \frac{\shadx^{[j]} \left| \tau^{[\inverj,j]} (0) \right|}{\shada} \right)^2
  \!\!+\!
  \inver{p}^{[j]}
  \shadx^{[j]2}
  \prever{\phi}^{[j]2} \undertilde{\prever{\psi}}^{[j]2} \\
 \end{array}
}\!\!\!}
\\
\hline
\hline
{\!\!\!{
 \begin{array}{l} \;
 \prever{\phi}^{[j]} = \mathop{\max}\limits_{\inverj \in \Inver{\mathbb{P}}^{[\inverj,j]}} \prever{\phi}^{[\inverj,j]},
 \prever{\psi}^{[j]} = \mathop{\max}\limits_{\inverj \in \Inver{\mathbb{P}}^{[\inverj,j]}} \prever{\psi}^{[\inverj,j]},
 \undertilde{\prever{\psi}}^{[j]} = \mathop{\max}\limits_{\inverj \in \Inver{\mathbb{P}}^{[\inverj,j]}} \undertilde{\prever{\psi}}^{[\inverj,j]} \\
 \end{array}
}\!\!\!}
&
{\!\!\!{
 \begin{array}{l}
  \;
  \Postver{\bg{\phi}}^{[-1]} ({\bf x}) = {\bf x}, \;
  \Postver{\bg{\Psi}}^{[-1]} ({\allofW}) = {\bf I}_n, \;
  \postver{\bf g}^{{[-1]}^{{\;\!\!}^{{\;\!\!}^{\;\!\!}}}} = {\bf 1}_n \\
  \;
  \postver{\phi}_{\mindownarrow}^{[-1]2} = 0, \; 
  \postver{\phi}^{[-1]2} = 
  \postver{\psi}^{[-1]2} = 
  \undertilde{\postver{\psi}}^{{[-1]2}} = 1 \;\;
  {\rm (} \prever{\phi}^{[\inverj,j]} \in \mathbb{R}_+ \;{\rm arbitrary)} \\
 \end{array}
}\!\!\!}
\\
\hline
{
\begin{array}{l}
{\bf f} \left( {\bf x}; \allofW \right) 
 =
 \leftbf \Postver{\bg{\Psi}} \left( {\allofW} \right), \Postver{\bg{\phi}} \left( {\bf x} \right) \rightbf_{\postver{\bf g}}^{{\;}^{{\;}^{{\;}^{\;}}}} \\
\end{array}
}
 &
{
\begin{array}{l}
  \Postver{\bg{\phi}} = \Postver{\bg{\phi}}^{[D-1]},
  \Postver{\bg{\Psi}} = \Postver{\bg{\Psi}}^{[D-1]},
  \postver{\bf g} = \postver{\bf g}^{{[D-1]}^{{\;}^{{\;}^{\;}}}} \\
  \postver{\phi}_{\mindownarrow} = \postver{\phi}_{\mindownarrow}^{[D-1]}, \postver{\phi} = \postver{\phi}^{[D-1]}], 
  \postver{\psi} = \postver{\psi}^{[D-1]},
  \undertilde{\postver{\psi}} = \undertilde{\postver{\psi}}^{[D-1]} \\
\end{array}
}
\end{array}
$
}
\caption[]{Recursive definition of the global dual and bounds.  
         See theorem \ref{th:netexpand}, section \ref{sec:globdual} for details.}
\label{fig:featmapdef}

$\;$\newline

\resizebox{\textwidth}{!}{%
$
\begin{array}{l|l}
\hspace{3.5cm}\text{\bf Incoming Edge Feature Map} & \hspace{3.5cm}\text{\bf Node Feature Map}_{{\;\!}_{{\;\!}_{{\;\!}_{\;\!}}}} \\
{\!\!\!{
 \rotatebox[]{90}{\text{\bf Feature Maps}}
 \begin{array}{rl}
  \Prever{\bg{\phi}}_{\changein}^{[\inverj,j]} \left( {\bf x} \right)
  &\!\!\!\!= \left[
             \frac{1}{\eta^{k}}
             \left[
             \left(
             \sqrt{\frac{\roctau_{(\tilde{\xpostm})\eta}^{[\inverj,j]2}}{\shada^2+1}}
             \Postver{\bg{\phi}}_{\changein}^{[\inverj]} \left( {\bf x} \right)
             \right)^{\otimes l}
             \right]_{1 \leq l \leq k}
             \right]_{k \geq 1} \\

  \Prever{\bg{\Psi}}_{\changein :i_{\inverj}}^{[\inverj,j]} \left( {\allofW} \right)
  &\!\!\!\!= \left[
             \eta^{k}
             \left[
             \left(
             \sqrt{\frac{\shada^2+1}{\roctau_{(\tilde{\xpostm})\eta}^{[\inverj,j]2}}}
             \Postver{\bg{\Psi}}_{\changein :i_{\inverj}}^{[\inverj]} \left( {\allofW} \right)
             \right)^{\otimes l}
             \right]_{1 \leq l \leq k}
             \right]_{{k \geq 1}_{{\;\!}_{{{\;\!}_{\;\!}}}}\!\!} \\

  \Prever{\bf G}_{\changein :i_{\inverj}}^{[\inverj,j]} \left( {\bf x} \right)
  &\!\!\!\!= \left[
             a_{\big(\postver{x}_{i_{\inverj}}^{[\inverj]}\big)k}^{[\inverj,j]}
             \left[
             \binom{k}{l} \Henum_{k-l}
             \Postver{\bf G}_{\changein :i_{\inverj}}^{[\inverj]} \left( {\bf x} \right)^{\otimes l}
             \right]_{1 \leq l \leq k}
             \right]_{{k \geq 1}_{{\;\!}_{{{\;\!}_{\;\!}}}}\!\!} \\
 \end{array}
}\!\!\!}
&
{\!\!\!{
 \begin{array}{rl}
  \Postver{\bg{\phi}}_\changein^{[j]} \left( {\bf x} \right)
  &\!\!\!\!\!=\! \left[ \!\!\begin{array}{c}
                    \shada \\

                    \frac{1}{\sqrt{3\inver{p}^{[j]}}}
                    \left[
                    \frac{1}{\tilde{\xpostm}^{[\inverj,j]}}
                    \prever{\bf x}^{[\inverj,j]}
                    \right]_{\inverj \in \Inver{\mathbb{P}}^{[j]}}
                    \\

                    \frac{1}{\sqrt{3\inver{p}^{[j]}}}
                    \left[
                    \frac{1}{\prever{\psi}_{\changein}^{[\inverj,j]}}
                    \Prever{\bg{\phi}}_{\changein}^{[\inverj,j]} \left( {\bf x} \right)
                    \right]_{\inverj \in \Inver{\mathbb{P}}^{[j]}}
                    \\

                    \frac{1}{\sqrt{3\inver{p}^{[j]}}}
                    \left[
                    \frac{1}{\prever{\psi}_{\changein}^{[\inverj,j]}}
                    \Prever{\bg{\phi}}_{\changein}^{[\inverj,j]} \left( {\bf x} \right)
                    \right]_{\inverj \in \Inver{\mathbb{P}}^{[j]}}
                    \\
             \end{array} \!\!\right] \\ 

  \Postver{\bg{\Psi}}_\changein^{[j]} \left( {\allofW} \right)
  &\!\!\!\!\!=\! \left[ \!\!\begin{array}{c}
                    \changein {\bf b}^{[j]\tsp} \\

                    \sqrt{3\inver{p}^{[j]}}
                    \mathop{\rm diag}_{{{{\inverj \in \Inver{\mathbb{P}}^{[j]}}}}}
                    \left(
                    {\tilde{\xpostm}^{[\inverj,j]}}
                    {\bf I}_{H^{[\inverj]}}
                    \right)
                    \changein {\bf W}^{[j]}
                    \\

                    \sqrt{3\inver{p}^{[j]}}
                    \mathop{\rm diag}_{{{{\inverj \in \Inver{\mathbb{P}}^{[j]}}}}}
                    \left(
                    {\prever{\psi}_{\changein}^{[\inverj,j]}}
                    \Prever{\bg{\Psi}}_{\changein}^{[\inverj,j]} \left( \allofW \right)
                    \right)
                    {\bf W}^{[j]}
                    \\

                    \sqrt{3\inver{p}^{[j]}}
                    \mathop{\rm diag}_{{{{\inverj \in \Inver{\mathbb{P}}^{[j]}}}}}
                    \left(
                    {\prever{\psi}_{\changein}^{[\inverj,j]}}
                    \Prever{\bg{\Psi}}_{\changein}^{[\inverj,j]} \left( \allofW \right)
                    \right)
                    \changein {\bf W}^{[j]}
                    \\
                 \end{array}\!\! \right] \\

  \Postver{\bf G}^{[j]}_\changein \left( {\bf x} \right)
  &\!\!\!\!\!=\! \left[ \!\!\begin{array}{c}
                    {\bf 1}_{H^{[\inverj]}}^\tsp \\

                    \mathop{\rm diag}_{{{{\inverj \in \Inver{\mathbb{P}}^{[j]}}}}}
                    \left(
                    {\bf I}_{H^{[\inverj]}}
                    \right)
                    {\bf 1}_{\inver{H}^{[j]}} {\bf 1}_{H^{[j]}}^\tsp \\

                    \mathop{\rm diag}_{{{{\inverj \in \Inver{\mathbb{P}}^{[j]}}}}}
                    \left(
                    \Prever{\bf G}^{[\inverj,j]}_\changein \left( {\bf x} \right)
                    \right)
                    {\bf 1}_{\inver{H}^{[j]}} {\bf 1}_{H^{[j]}}^\tsp \\

                    \mathop{\rm diag}_{{{{\inverj \in \Inver{\mathbb{P}}^{[j]}}}}}
                    \left(
                    \Prever{\bf G}^{[\inverj,j]}_\changein \left( {\bf x} \right)
                    \right)
                    {\bf 1}_{\inver{H}^{[j]}} {\bf 1}_{H^{[j]}}^\tsp \\
             \end{array} \!\!\right] \\
 \end{array}
}\!\!\!}
\\
\hline
{\!\!\!{
 \rotatebox[]{90}{\text{\bf Bounds}}
 \begin{array}{rl}
  \left\| \Prever{\bg{\phi}}_{\changein}^{[\inverj,j]} \left( {\bf x} \right) \odot \Prever{\bf G}_{\changein :i_{\inverj}}^{[\inverj,j]} \left( {\bf x} \right) \right\|_2^{2^{{\;}^{{\;}^{\;}}}}
  &\!\!\!\!\leq
  \prever{\phi}_{\changein}^{[\inverj,j]2} = \abstaumax_{(\tilde{\xpostm})\eta}^{[\inverj,j]2} \;\;\;\;\forall i_{\inverj} \\

  \left\| \Prever{\bg{\Psi}}_{\changein}^{[\inverj,j]} \left( {\allofW} \right) \right\|_2^2
  &\!\!\!\!\leq
  \prever{\psi}_{\changein}^{[\inverj,j]2}
  =
  \absT_\eta \left( \frac{\shada^2+1}{\roctau_{(\tilde{\xpostm})\eta}^{[\inverj,j]2}} \postver{\psi}_{\changein  }^{[\inverj]2} \right)_{{\;}_{\;}}
  \\
 \end{array}
}\!\!\!}
&
{\!\!\!{
 \begin{array}{rll}
  \left\| \Postver{\bg{\phi}}_{\changein}^{[j]} \left( {\bf x} \right) \odot \Postver{\bf G}_{\changein :i_{j}}^{[j]} \left( {\bf x} \right) \right\|_2^2
  &\!\!\!\!\leq
  \postver{\phi}_{\changein }^{[j]2}
  &\!\!\!\!=
  \shada^2+1 \;\;\;\;\forall i_j \\

  \left\| \Postver{\bg{\Psi}}_{\changein}^{[j]} \left( {\allofW} \right) \right\|_2^2
  &\!\!\!\!\leq
  \postver{\psi}_{\changein}^{[j]2}
  &\!\!\!\!=
   \shadb_\changein^{[j]2}
   + 
   3\inver{p}^{[j]}
   \left(
   \shadx_\changein^{[j]2}
   \tilde{\xpostm}^{[j]2}
   +
   2\shadx^{[j]2}
   \prever{\psi}_{\changein}^{[j]2}
  \right)\!\!\!
 \end{array}
}\!\!\!}
\\
\hline
\hline
{\!\!\!{
 \begin{array}{l} \;
 \tilde{\xpostm}^{[j]} = \mathop{\max}\limits_{\inverj \in \Inver{\mathbb{P}}^{[j]}} \tilde{\xpostm}^{[\inverj,j]}, \; 
 \abstaumax_{(\tilde{\xpostm})\eta}^{[j]} = \mathop{\max}\limits_{\inverj \in \Inver{\mathbb{P}}^{[j]}} \abstaumax_{(\tilde{\xpostm})\eta}^{[\inverj,j]}, \; 
 \prever{\psi}_{\changein}^{[j]} = \mathop{\max}\limits_{\inverj \in \Inver{\mathbb{P}}^{[j]}} \prever{\psi}_{\changein_{{\;}_{{\;}_{\;}}}}^{{[\inverj,j]}^{{\;}^{{\;}^{\;}}}} \\
 \end{array}
}\!\!\!}
&
{\!\!\!{
 \begin{array}{l}
  \;
  \Postver{\bg{\phi}}_\changein^{[-1]} ({\bf x}) = {\bf 0}_0, \;
  \Postver{\bg{\Psi}}_\changein^{[-1]} ({\allofW}) = 
  \Postver{\bf G}_\changein^{[-1]} ({\bf x}) = {\bf 1}_{0 \times n}, \;
  \postver{\phi}_{\changein}^{[-1]2} = 
  \postver{\psi}_{\changein}^{[-1]2} = 0 \\
 \end{array}
}\!\!\!}
\\
\hline
{
\begin{array}{rl}
   {\bf f} \left( {\bf x}; \allofW + \changein \allofW \right) 
   &\!\!\!\!=
   {\bf f} \left( {\bf x}; \allofW \right) 
   +
   \changein {\bf f} \left( {\bf x}; \changein \allofW \right)^{{\;}^{{\;}}} \\

   \changein {\bf f} \left( {\bf x}; \changein \allofW \right) 
   &\!\!\!\!=
   \leftbf \Postver{\bg{\Psi}}_{\changein} \left( \changein {\allofW} \right), \Postver{\bg{\phi}}_\changein \left( {\bf x} \right) \rightbf_{{\bf G}_\changein ({\bf x})}^{{\;}^{{\;}^{\;}}} \\
\end{array}
}
 &
{
\begin{array}{l}
  \Postver{\bg{\phi}}_\changein = \Postver{\bg{\phi}}_{\changein}^{[D-1]}, 
  \Postver{\bg{\Psi}}_\changein = \Postver{\bg{\Psi}}_{\changein}^{[D-1]}, 
  \Postver{\bf G}_\changein = \Postver{\bf G}_{\changein}^{{[D-1]}^{{\;}^{{\;}^{\;}}}} \\
  \postver{\phi}_{\changein} = \postver{\phi}_{\changein}^{[D-1]}, 
  \postver{\psi}_{\changein} = \postver{\psi}_{\changein}^{[D-1]} \\
\end{array}
}
\end{array}
$
}
\caption[]{Recursive definition of local dual and bounds.  
         See theorem \ref{th:netexpand_changein}, section \ref{sec:locmod} for details.}
\label{fig:changein_feat}
\end{center}
\end{figure*}

\section{Local Dual Model in Reproducing Kernel Hilbert Space} \label{sec:converge} \label{sec:locmod}

When considering training or network adaptation it is better to model the 
change in the network rather than the network in-toto.  To this end, in this 
section we present an exact (non-approximate) local RKHS model.  
Let $\allofW$ be the initial weight and biases (before a training step) 
and $\changein \allofW$ the change in weights and biases (the training 
step). Let $\prever{\bf x}^{[j]}, \postver{\bf x}^{[j]}$ denote the 
pre-activation (input) and post-activation (output) of node $j$ with 
initial weights given input ${\bf x}$; and $\changein \prever{\bf x}^{[j]}, 
\changein \postver{\bf x}^{[j]}$ the change in same due to the 
change in weights.  The change in network operation is denoted $\changein 
{\bf f} : \mathbb{X} \times \mathbb{W}_\changein \to \mathbb{R}^m$:
\begin{equation}
 \begin{array}{c}
  {\bf f} \left( {\bf x}; \allofW + \changein \allofW \right)
  =
  {\bf f} \left( {\bf x}; \allofW \right) 
  +
  \changein {\bf f} \left( {\bf x}; \changein \allofW \right), \;\; 
 \end{array}
\label{eq:f_changein}
\end{equation}
Simultaneously with previous assumptions, whp $\geq 1-\epsilon$,  we assume 
$\exists \shadx_\changein^{[j]}, \shadb_\changein^{[j]}, 
\tilde{\xpostm}^{[\inverj,j]} \in \mathbb{R}_+$ so that:
\vspace{-\topsep}
\begin{enumerate}
\setcounter{enumi}{3}
\setlength\itemsep{0em}
 \item Finite base activation: $\| \prever{\bf x}^{[\inverj,j]} \|_2 \leq \tilde{\xpostm}^{[\inverj,j]}$ $\forall \inverj \in \Inver{\mathbb{P}}^{[j]}$ 
       (note that $\tilde{\xpostm}^{[\inverj,j]} \leq {\prever{\phi}}{}^{[\inverj,j]} \undertilde{\prever{\psi}}^{[\inverj,j]}$).
 \item Finite weight and bias step: $\| \changein {\bf W}^{[j]} \|_{2} \leq \shadx_\changein^{[j]}$, $\| \changein {\bf b}^{[j]} \|_{2} \leq \shadb_\changein^{[j]}$. 
\end{enumerate}
\vspace{-\topsep}
We define $\mathbb{W}_\changein$ to be the set of all weight-steps satisfying 
these assumptions.  The parameter $\tilde{\xpostm}^{[\inverj,j]}$ is a bound 
on the magnitude of the output of edge $(\inverj\to j)$ in our initial network 
$\forall {\bf x} \in \mathbb{X}$.  With this prequel, we have the following 
local analogue of theorem \ref{th:netexpand} (see proof in supplementary 
\ref{append:dualformlocal}):\footnote{The decision to use a position dependent 
metric ${\bf G}_\changein$ here is largely stylistic.  We could of course 
absorb ${\bf G}_\changein$ into $\Postver{\bg{\phi}}_\changein$ without 
substantively changing our results.}
\begin{th_netexpand_changein}
 Let $\changein {\bf f} : \mathbb{X} \times \mathbb{W}_\changein \to \mathbb{R}^m$ 
 be the change in neural network operation (\ref{eq:f_changein}).  Then:
 \begin{equation}
  \begin{array}{rl}
   \changein {\bf f} \left( {\bf x}; \changein \allofW \right) 
   &\!\!\!\!= \leftbf \Postver{\bg{\Psi}}_{\changein} \left( \changein {\allofW} \right), \Postver{\bg{\phi}}_\changein \left( {\bf x} \right) \rightbf_{{\bf G}_\changein ({\bf x})} \\
  \end{array}
 \label{eq:ourdualmodel_changein}
 \end{equation}
 with feature maps 
 $\Postver{\bg{\phi}}_\changein : \mathbb{X}           \to \Postver{\mathcal{X}}_\changein = \overline{\rm span} (\Postver{\bg{\phi}}_\changein (\mathbb{X}          ))$, 
 $\Postver{\bg{\Psi}}_\changein : \mathbb{W}_\changein \to \Postver{\mathcal{W}}_\changein = \overline{\rm span} (\Postver{\bg{\Psi}}_\changein (\mathbb{W}_\changein))$ 
 and metric 
 $\Postver{\bf G}_\changein ({\bf x})$ 
 as per Figure \ref{fig:changein_feat}, where 
 $\| \Postver{\bg{\phi}}_\changein ({\bf x}) \odot \Postver{\bf G}_{\changein:i_{D-1}} ({\bf x}) \|_2 \leq \postver{\phi}_{\changein}$ $\forall i_{D-1}$ 
 and 
 $\| \Postver{\bg{\Psi}}_\changein (\changein \allofW) \|_2 \leq \postver{\psi}_{\changein}$ 
 $\forall {\bf x} \in \mathbb{X}$, $\changein \allofW \in \mathbb{W}_\changein$.  
 Moreover 
 $\| \Postver{\bg{\Psi}}_\changein (\changein \allofW) \|_2 \leq \postver{\psi}_{\changein} = \absTmax_\eta^2 < 1$ 
 if $\forall j$:
\label{th:netexpand_changein}
\end{th_netexpand_changein}
\vspace{-0.5cm}
 \begin{equation}
 {\!\!\!\!\!\!\!\!\!\!\!\!{
  \begin{array}{rl}
   \shadx_\changein^{[j]2}
   +
   \frac{1}{3\inver{p}^{[j]} \tilde{\xpostm}^{[j]2}}
   \shadb_\changein^{[j]2}
   \leq
   \frac{u^{[j]2}}{6\inver{p}^{[j]} \tilde{\xpostm}^{[j]2}} \; : 
   \;\;\;
   u^{[j]2}
   =
   \mathop{\min}\limits_{\inverj:j \in \Inver{\mathbb{P}}^{[\inverj]}}
   \frac{1}{\shada^2+1} \roctau_{(\tilde{\xpostm}^{[j,\inverj]})\eta}^{[j,\inverj]2}
   \left\{
   \rocT_\eta^2,
    \absT_\eta^{-1}
    \left(
    \frac{u^{[\inverj]2}}{12\inver{p}^{[\inverj]} \shadx^{[\inverj]2}}
    \right)
   \right\}
  \end{array}
 }\!\!\!\!\!\!\!\!\!\!\!\!}
 \label{eq:stepbnddef}
 \end{equation}

\subsection{Implication: Neural Network Change in Reproducing Kernel Hilbert Space}

A vector-valued (v-v) reproducing kernel Hilbert space is defined as follows 
\citep{Aro1,Ste3,Sha3,Mer1,Mit1,Cap1,Rae1,Car3,Sch50}:
\begin{def_rkhs}[Reproducing kernel Hilbert space (RKHS)]
 A v-v reproducing kernel Hilbert space $\rkhs{}$ on a set $\mathbb{X}$ is a 
 Hilbert space $\mathcal{F}$ of functions ${\bf f} : \mathbb{X} \to 
 \mathbb{R}^m$ for which the point evaluation functionals ${\bg{\delta}}_{\bf 
 x} ({\bf f}) = {\bf f}({\bf x})$ on $\mathcal{F}$ are continuous ($\forall 
 {\bf x} \in \mathbb{X}$ $\exists C_{\bf x} \in \mathbb{R}_+$ s.t. 
 $\|{\bg{\delta}}_{\bf x} ({\bf f})\|_2 \leq C_{\bf x} \|{\bf f}\|_\mathcal{F}$ 
 $\forall {\bf f} \in \mathcal{F}$).
\end{def_rkhs}
For an RKHS, Reisz representor theory implies that $\forall {\bf x} \in 
\mathbb{X}$ $\exists$ unique ${\bf K}_{\bf x} \in \mathcal{F} \times 
\mathbb{R}^m$ such that $\left< {\bf f} ({\bf x}), {\bf v} \right> = \left< 
{\bf f}, {\bf K}_{\bf x} {\bf v} \right>_{\rkhs{}}$ $\forall {\bf v} \in 
\mathbb{R}^m$.  From this, the kernel ${\bf K} : \mathbb{X} \times \mathbb{X} 
\to \mathbb{R}^{m \times m}$ is defined as:
\[
 \begin{array}{l}
  {\bf K} \left( {\bf x},{\bf x}' \right)
  =
  \Big[ \left< {\bf K}_{\bf x} {\bg{\delta}}{}_{(i_{D-1})}^{[D-1]}, {\bf K}_{{\bf x}'} {\bg{\delta}}{}_{(i'_{D-1})}^{[D-1]} \right>_{\rkhs{}} \Big]{}_{{{}\atop{\!i_{D-1},i'_{D-1}}}},
\;\;\; \mbox{where} \; {\bg{\delta}}_{(k)}^{[j]} = [\delta_{k,i_j}]_{i_j}
 \end{array}
\]
Moore-Aronszajn theorem allows us to run the argument in reverse: any 
symmetric, positive definite ${\bf K} : \mathbb{X} \times \mathbb{X} \to 
\mathbb{R}^{m \times m}$ uniquely defines an RKHS, $\rkhs{\bf 
K}$ for which ${\bf K}$ is the kernel.  
From theorem \ref{th:netexpand_changein}:
\begin{th_itshilbert}
 The set $\mathcal{F}_\changein \!= \!\{ \changein {\bf f} (\cdot; \changein 
 \allofW ) : \mathbb{R}^n \to \mathbb{R}^m | \changein \allofW \in 
 \mathbb{W}_\changein \}$ of changes in network behavior satisfying our 
 assumptions, including the bound, lies in an RKHS $\rkhs{\bf K}$ (that 
 is, $\mathcal{F}_\changein \subset \rkhs{\bf K}$) 
 with kernel ${\bf K} = {\bf I}_{m} K_{\scriptscriptstyle{\rm LiNK}}$, where 
 $K_{\scriptscriptstyle{\rm LiNK}} = \Postver{K}^{[D-1]}$, is the {\em 
 Local-intrinsic Neural Kernel (LiNK)}, $\forall j$:
 \begin{equation}
 {\!\!\!\!\!\!\!\!\!\!\!\!\!\!\!\!\!\!\!\!\!\!\!{
  \begin{array}{rl}
  \Postver{K}^{[j]} \!\left( {\bf x}, \!{\bf x}' \right)
  &\!\!\!\!=
  \shada^2
  +
  \mathbb{E}_{\inverj \in \Inver{\mathbb{P}}^{[j]}}
  \left[
  \frac{\Sigma^{[\inverj,j]} \left( {\bf x}, {\bf x}' \right)}{\tilde{\xpostm}^{[\inverj,j]2}}
  +
  \frac{1}{\prever{\psi}_{\changein}^{[\inverj,j]}}
  \mathbb{E}_{i_\inverj}
  \left[
   \abstau_\eta^{[\inverj,j]}
   \left(
   \frac{\roctau_{(\tilde{\xpostm})\eta}^{[\inverj,j]2}}{\shada^2+1}
   \Postver{K}^{[\inverj]} \left( {\bf x}, {\bf x}' \right)
   ; \postver{x}_{i_{\inverj}}^{[\inverj]}, \postver{x}_{i_{\inverj}}^{\prime[\inverj]}
   \right)
  \right]
  \right] \\
  \end{array}
 }\!\!\!\!\!\!\!\!\!\!\!\!\!\!}
 \label{eq:link_def}
 \end{equation}
 and $K_{\scriptscriptstyle{\rm LiNK}}^{[-1]} ({\bf x}, {\bf x}') = 0$ and 
 $\Postver{\Sigma}^{[j]} ({\bf x},{\bf x}')$ is the NNGP kernel.  Moreover:
 \begin{equation}
 {\!\!\!\!\!\!\!\!{
  \begin{array}{l}
   \mathop{\lim}\limits_{\eta \to 1}
  \mathbb{E}_{i_\inverj}
  \left[
   \abstau_\eta^{[\inverj,j]}
   \left(
   \frac{\roctau_{(\tilde{\xpostm})\eta}^{[\inverj,j]2}}{\shada^2+1}
   \Postver{K}^{[\inverj]} \left( {\bf x}, {\bf x}' \right)
   ; \postver{x}_{i_{\inverj}}^{[\inverj]}, \postver{x}_{i_{\inverj}}^{\prime[\inverj]}
   \right)
  \right]
   =
  \mathop{\sum}\limits_{q\geq 1}
   \theta_l^{[\inverj,j]} \left( {\bf x}, {\bf x}' \right)
   \left(
   \frac{\roctau_{(\tilde{\xpostm})\eta}^{[\inverj,j]2}}{\shada^2+1}
   \Postver{K}^{[\inverj]} \left( {\bf x}, {\bf x}' \right)
   \right)^q \\
   \theta_q^{[\inverj,j]} \left( {\bf x}, {\bf x}' \right)
   =
  \mathbb{E}_{i_\inverj}
  \left[
  \frac{1}{q!}
  \tau^{[\inverj,j](q)} \left( \postver{x}_{i_{\inverj}}^{[\inverj]} \right)
  \frac{1}{q!}
  \tau^{[\inverj,j](q)} \left( \postver{x}_{i_{\inverj}}^{\prime[\inverj]} \right)
  \right]
  \end{array}
  \label{eq:define_abstau_extend}
 }}
 \end{equation}
 (here $\theta_q^{[\inverj,j]} ({\bf x}, {\bf x}')$ is the raw covariance of the 
 $q^{\rm th}$ derivative of link $(\inverj,j)$'s activation given 
 ${\bf x}$, ${\bf x}'$.)
 \label{th:itshilbert}
\end{th_itshilbert}
The proof is somewhat long and is provided in supplementary \ref{sup:itshilbert}.  
Note that the NTK is essentially (with some additional scaling factors) a 
first-order (in $q$) approximation of the LiNK.  Assuming random initialization 
the LiNK is well-defined for almost all ${\bf x} \in \mathbb{X}$ if $\tau^{[\inverj,j]} \in 
\mathcal{C}^\infty$ for almost all ${\bf x} \in \mathbb{X}$.  
Note however that $\mathcal{F}_\changein \subset 
\rkhs{\bf K}$ - ie. $\mathcal{F}_\changein$ is not an RKHS in 
general, but rather a subspace inside of one.  Nor can we meaningfully replace 
$\mathcal{F}_\changein$ with its span or completion, as this will contain 
elements that do not correspond to physically realizable networks.  
Nevertheless, as we show in section \ref{sec:banach_rep}, we can use the local model to 
construct a representor theory in terms of Banach kernels that, while related 
to the LiNK, differ in certain important respects.

\subsection{Application: Bounding Rademacher Complexity for Neural Network Adaptation}

Like the global model, an obvious application of the local dual model is the 
bounding of Rademacher complexity.  The following result may be viewed as the 
local analogue of our previous bound:
\begin{th_radbounded_local}
 The set $\mathcal{F}_\changein = \{\changein f(\cdot;\changein \allofW): 
 \mathbb{R}^n \to \mathbb{R} | \changein \allofW \in \mathbb{W}_\changein 
 \}$ of change in neural-network operation satisfying (\ref{eq:stepbnddef}) 
 has Rademacher complexity $\mathcal{R}_N (\mathcal{F}) \leq\frac{1}{\sqrt{N}} 
 {\postver{\phi}}_{\changein} {\postver{\psi}}_{\changein}$ (defined in Figure 
 \ref{fig:changein_feat}).
 \label{th:radbounded_local}
\end{th_radbounded_local}
The proof follows the template of \citep{Bar1} using the local feature map and 
the Cauchy-Schwarz inequality (see supplementary \ref{supp:radbound}).  
${\postver{\psi}}_{\changein}$ scales with $\shadx_\changein$, so 
methods such as LoRA \citep{Hu4} may be expected to get 
tighter bounds by restricting the weight update rank.  
As an aside for future investigation we note that 
spectral analysis of the LiNK could, in principle, be used to bound local 
Rademacher complexity \citep{Cor5,Bar11} up to $\mathcal{O} ({1}/{N})$.

\section{Exact Representor Theory} \label{sec:banach_rep}

In this section we construct a representor theory for the change in 
neural-network operation $\changein {\bf f} (\cdot; \changein \allofW)$ 
for layer-wise neural networks.  The representor theory may be viewed as an 
extension of the standard NTK-based approximate representor theory to an 
exact, higher-order model.  Distinct from standard representor theory, our 
representor theory is expressed in terms of weighted trees of height $D$ with 
labeled leaves.  Inverting the structure of the layer-wise neural network, 
the root node of these trees is node $D-1$, and ``layers'' (working down from 
the root) are $D-2,\ldots,1,0,-1$.  Precisely:
\begin{def_datatree}
 Let $\mathcal{E} = \{ {\bf x}_{\{l\}} : l \in \mathbb{N}_N \}$, ${\bf e}, 
 {\bf f} \in \mathbb{Z}_+^M$, ${\bg{\varsigma}}\in\mathbb{N}_N^{f_0 f_1 
 \ldots}$.  Then $\mathcal{E}{}^{\lfloor\!{\bf e}, {\bf f}, {\bg{\varsigma}}\! 
 \rceil\!}$ denotes a weighted tree of height $M$ with root node $M-1$; layers 
 $M-2,\ldots,0,-1$; edge weights (per layer) $e_{M-1},e_{M-2},\ldots,e_0$; 
 fan-outs (per layer) $f_{M-1},f_{M-2},\ldots,f_0$; and leaf labels ${\bf 
 x}_{\{\varsigma_0\}}$, ${\bf x}_{\{\varsigma_1\}}$, $\ldots$.  Sub-trees are 
 denoted $\mathcal{E}{}^{\lfloor\!{\bf e}, {\bf f}, {\bg{\varsigma}}\!\rceil 
 \!}_k = \mathcal{E}{}^{\lfloor\!{\bf e}_{{\backslash}M-1}, {\bf 
 f}_{{\backslash}M-1}, {\bg{\varsigma}}_{\backslash f_{M-2}\ldots{f_1f_0k}: 
 f_{M-2}\ldots{f_1f_0(k+1)-1}}\!\rceil\!}$ where $k \in \mathbb{N}{}_{f_{M-1}}$.
 \label{def:datatree}
\end{def_datatree}

We use the shorthand ${\{{\bf x}\}}^{\scriptscriptstyle{\bf e},{\bf 1},-}$ 
where the data label is superfluous.  Our key result for this section is:
\begin{th_netexpand_kernel}
 Let $\changein {\bf f} : \mathbb{X} \times \mathbb{W}_\changein \to 
 \mathbb{R}^m$ be the change in neural network operation (\ref{eq:f_changein}) 
 satisfying the constraints given in theorem \ref{th:netexpand_changein}.  For 
 a dense layer-wide feedforward network, assume $\changein \allofW$ is 
 generated by gradient a descent step on unregularized risk
 on training set $\mathcal{D}$.  Then:
 \begin{equation}
 {\!\!\!\!\!\!\!\!\!\!\!\!\!\!\!\!\!\!\!\!\!\!\!\!\!\!\!\!\!\!\!\!\!\!\!\!\!\!\!{
  \begin{array}{l}
   \changein {\bf f} \left( {\bf x}; \changein \allofW \right)
   \!=\!\! 
   \mathop{\sum}\limits_{{\bf k} \in \mathbb{Z}_+^D}
   \!\!\!\frac{\left( -\eta \right)^{k_0k_1\ldots}}{{\bf k}!}\!\!\!\!\!\!\!\!
   \mathop{\sum}\limits_{{\bf p} \in \mathbb{N}_N^{k_0k_1\ldots}}
   \!\!\!\!\!\!\postver{\bf K}_{\scriptscriptstyle{\rm LeNK}}
   \!\left(\!
   \left\{ \!\,{\bf x}\,\! \right\}^{\lfloor\!{\bf k},{\bf 1},-\!\rceil\!}\!\!,
   \mathcal{D}^{\lfloor\!{\bf 1},{\bf k},{\bf p}\!\rceil\!};
   \allofW+\changein\allofW,\allofW
   \right)
   \!\postver{\bg{\alpha}}_{\{p_{0}\!\}} \!\otimes
   \!\postver{\bg{\alpha}}_{\{p_{1}\!\}} \!\otimes
   \!\ldots\!
  \end{array}
 }\!\!\!\!\!\!\!\!\!\!\!\!\!\!\!\!\!\!\!\!\!\!\!\!}
 \label{eq:repstep}
 \end{equation}
 where ${\bf k}! = k_0!k_1!\ldots$ and $\postver{\bg{\alpha}}_{\{l\}} \in 
 \mathbb{R}^n$ $\forall l \in \mathbb{N}_N$.  Here $\Postver{\bf 
 K}_{\scriptscriptstyle{\rm LeNK}}$ is the (matrix-valued) Local-extrinsic 
 Neural Kernel (LeNK) on the space of weighted trees (definition 
 \ref{def:datatree}), defined as $\Postver{\bf 
 K}_{\scriptscriptstyle{\rm LeNK}} = \Postver{\bf K}^{[D-1]}$:\newline
\begin{equation}
{\!\!\!\!\!\!\!\!\!\!\!\!\!\!\!\!{
\begin{array}{l}
\resizebox{0.95\textwidth}{!}{%
$
{\!\!\!\!{
  \begin{array}{r}
   \postver{\bf K}^{[j]}
   \left(
   {\mathcal{E}}^{\prime\lfloor{\bf e}', {\bf f}', {\bg{\varsigma}}' \rceil},
   {\mathcal{E}}^{\lfloor{\bf e}, {\bf f}, {\bg{\varsigma}}\rceil};
   \allofW', \allofW
   \right)
   =
   \delta_{{\bf e}'}^{{\bf 1}}
   \delta_{{\bf f}'}^{{\bf 1}}
   \delta_{{\bf e}}^{{\bf 1}}
   \delta_{{\bf f}}^{{\bf 1}}
   {\bf I} \Sigma^{[j]} \left( {\bf x}'_{\{\varsigma'_0\}}, {\bf x}_{\{\varsigma_0\}} \right)
   +
   \Big(
   \mathop{\bigotimes}\limits_{k \in \mathbb{N}_{f'_j}}
   \Big(
   \mathop{\rm diag}\limits_{i_\inverj}
   \left(
   \tau^{[\inverj,j](e'_j)} \left( {\mathcal{E}}_{k}^{\prime\lfloor{\bf e}', {\bf f}', {\bg{\varsigma}}' \rceil}{}_{i_\inverj} \right)
   \right)
   {\bf W}^{\prime[\inverj,j]}
   \Big)
   \Big)^\tsp
   \\
   \ldots
   \Big(
   \mathop{\bigotimes^\updownarrow}\limits_{k' \in \mathbb{N}_{f'_j}}
   \mathop{\bigotimes^\leftrightarrow}\limits_{k \in \mathbb{N}_{f_j}}
   \postver{\bf K}^{[\inverj]}
   \left(
   {\mathcal{E}}^{\prime\lfloor{\bf e}', {\bf f}', {\bg{\varsigma}}' \rceil}_{k'},
   {\mathcal{E}}^{\lfloor{\bf e}, {\bf f}, {\bg{\varsigma}} \rceil}_{k};
   \allofW', \allofW
   \right)
   \Big)
   \Big(
   \mathop{\bigotimes}\limits_{k \in \mathbb{N}_{f_j}}
   \Big(
   \mathop{\rm diag}\limits_{i_\inverj}
   \left(
   \tau^{[\inverj,j](e_j)} \left( {\mathcal{E}}_{k}^{\lfloor{\bf e}, {\bf f}, {\bg{\varsigma}} \rceil}{}_{i_\inverj} \right)
   \right)
   {\bf W}^{[\inverj,j]}
   \Big)
   \Big)
  \end{array}
}}
$}
\end{array}
}\!\!\!\!\!\!\!\!\!\!\!\!\!\!\!\!\!}
\label{eq:blinkdef}
\end{equation}
 recursively, with ${\bf K}_{\scriptscriptstyle{\rm LeNK}}^{[-1]} = 
 0$;\footnote{Recall that ${\bf A} \otimes^\updownarrow {\bf B} $ and ${\bf A} 
 \otimes^\leftrightarrow {\bf B}$ denote, respectively, the column- and 
 row-wise Kronecker products.} and $\tau^{[\inverj,j](\varepsilon)} 
 ({\mathcal{E}}^{\lfloor{\bf e}, {\bf f}, {\bg{\varsigma}} \rceil}{}_{i_\inverj}) 
 = \prod_r \tau^{[\inverj,j](\varepsilon)} (x_{\{\varsigma_r\}i_\inverj})$.  
 Furthermore:
\label{th:netexpand_kernel}
\end{th_netexpand_kernel}
\vspace{-0.5cm}
 \begin{equation}
 {\!\!\!\!\!\!\!\!\!\!\!\!\!\!\!\!\!\!\!\!\!\!\!\!{
  \begin{array}{l}
   \left.
   \begin{array}{l}
    \changein {\bf W}^{[j-1,j]} \!=\! \sum_l \prever{\bf x}_{\{l\}}^{[j-1,j]} \postver{\bg{\alpha}}_{\{l\}}^{[j]\tsp} \\
    \changein {\bf b}^{[j]}     \!=\! \sum_l \shada \postver{\bg{\alpha}}_{\{l\}}^{[j]} \\
   \end{array}
   \!\!\!\right|\!\!
   \begin{array}{l}
    \postver{\bg{\alpha}}_{\{l\}}^{[j-1]} \!=\! \mathop{\rm diag}\limits_{{i \in \mathbb{N}_m}} \left( \tau^{[j-1,j](1)} \!\left( \postver{x}_{\{l\}i}^{[j-1]} \right) \right) \!\!{\bf W}^{[j-1,j]} \postver{\bg{\alpha}}_{\{l\}}^{[j]} \forall j>0 \\
    \postver{\bg{\alpha}}_{\{l\}}^{[D-1]} \!=\! \postver{\bg{\alpha}}_{\{l\}} = \nabla_{\postver{\bf x}_{\{l\}}^{[D-1]}} L \left( \postver{\bf x}_{\{l\}}^{[D-1]} - {\bf y}_{\{l\}} \right) \\
   \end{array} \\
  \end{array}
 }\!\!\!\!\!\!\!\!\!\!\!\!}
 \label{eq:changewrep}
 \end{equation}
A proof of this theorem is given in supplementary \ref{sup:reptheory}.  We now 
consider the novel characteristics of this representor theory (and the LeNK in 
particular) and give successive approximations connecting with the LiNK and 
NTK regime:\newline
       {\bf Kernel warping:} As may be observed from (\ref{eq:repstep}) and 
       (\ref{eq:blinkdef}), the weight-step itself is a part of the 
       kernel definition. This arises due to interactions between $\changein 
       \prever{\bf x}^{[\inverj,j]}$ and $\changein \postver{\bf W}^{[\inverj, 
       j]}$.  For a sufficiently small weight-step it is reasonable to 
       approximate $\allofW + \changein \allofW \approx 
       \allofW$, in which case (\ref{eq:repstep}) simplifies to:
\begin{equation}
 {\!\!\!\!\!\!\!\!\!\!\!\!\!\!\!\!\!\!\!\!\!\!\!\!\!\!\!\!\!\!\!\!\!\!\!{
 \begin{array}{l}
   \changein {\bf f} \left( {\bf x}; \changein \allofW \right)
   \!\approx\! 
   {\sum}_{\bf k}
   \frac{\left( -\eta \right)^{k_0k_1\ldots}}{{\bf k}!}
   {\sum}_{\bf p}
   \postver{\bf K}_{\scriptscriptstyle{\rm LeNK}}\!
   \left(
   \left\{ {\bf x} \right\}^{\lfloor{\bf k},{\bf 1},-\rceil},
   \mathcal{D}^{\lfloor{\bf 1},{\bf k},{\bf p}\rceil};
   \allofW,\allofW
   \right)
   \postver{\bg{\alpha}}_{\{p_{0}\}} \!\otimes\!
   \postver{\bg{\alpha}}_{\{p_{1}\}} \!\otimes\!
   \ldots
 \end{array}
 }\!\!\!\!\!\!\!\!\!\!\!\!\!\!\!\!\!\!\!\!\!\!\!\!\!\!\!\!\!\!\!\!}
\label{eq:kernwarpapprox}
\end{equation}
       which is more conventional insofar as the kernel is ``fixed'' (not a 
       function of $\changein \allofW$).\newline
{\bf Cross terms:} Unlike the LiNK and the NTK, the LeNK is an unavoidably 
       matrix-valued kernel.  In the wide-network limit, after Gaussian 
       initialization, the off-diagonal terms will tend to $0$, so:
\begin{equation}
 {\!\!\!\!\!\!\!\!\!\!\!\!\!\!\!\!\!\!\!\!\!\!\!\!\!\!\!\!\!\!\!\!\!\!\!{
 \begin{array}{l}
   \changein {\bf f} \left( {\bf x}; \changein \allofW \right)
   \approx 
   {\sum}_{\bf k}
   \frac{\left( -\eta \right)^{k_0k_1\ldots}}{{\bf k}!}
   {\sum}_{\bf p}
   \postver{K}_{\scriptscriptstyle{\rm LeNK}}^{\scriptscriptstyle{\rm NCT}}
   \left(
   \left\{ {\bf x} \right\}^{\lfloor{\bf k},{\bf 1},-\rceil},
   \mathcal{D}^{\lfloor{\bf 1},{\bf k},{\bf p}\rceil}
   \right)
   \postver{\bg{\alpha}}_{\{p_{0}\}} \odot
   \postver{\bg{\alpha}}_{\{p_{1}\}} \odot
   \ldots
 \end{array}
 }\!\!\!\!\!\!\!\!\!\!\!\!\!\!\!\!\!\!\!\!\!\!\!\!\!\!\!\!\!\!\!\!}
\label{eq:kerncrossapprox}
\end{equation}
       where $\postver{K}_{\scriptscriptstyle{\rm LeNK}}^{\scriptscriptstyle{\rm NCT}} 
       = K^{[D-1]}$ is defined by recursively by (with $K^{[-1]} = 0$, $\inverj=j-1$):
\[
{\!\!\!\!\!\!\!\!\!\!\!\!\!\!\!\!\!\!\!\!{
  \begin{array}{l}
   \postver{K}^{[j]}
   \left(
   {\mathcal{E}}^{\prime\lfloor\ldots\rceil},
   {\mathcal{E}}^{\lfloor\ldots\rceil}
   \right)
   =
   \delta_{{\bf e}'}^{{\bf 1}}
   \delta_{{\bf f}'}^{{\bf 1}}
   \delta_{{\bf e}}^{{\bf 1}}
   \delta_{{\bf f}}^{{\bf 1}}
   \Sigma^{[j]}
   \left( 
   {\bf x}'_{\!\{\varsigma'_0\}}\!,
   {\bf x}_{\!\{\varsigma_0\}}
   \right)
   \!+\!
   \mathop{\prod}\limits_{k',k}
   \theta^{[\inverj,j]}_{\!(e'_j\!,e_j\!)}
   \left(
   {\mathcal{E}}^{\prime\lfloor\ldots\rceil}_{k'}\!,
   {\mathcal{E}}^{\lfloor\ldots\rceil}_{k}
   \right)
   \postver{K}^{[\inverj]}
   \left(
   {\mathcal{E}}^{\prime\lfloor\ldots\rceil}_{k'}\!,
   {\mathcal{E}}^{\lfloor\ldots\rceil}_{k}
   \right) \\
  \theta^{[\inverj,j]}_{(e',e)}
  \left(
  {\mathcal{E}}^{\prime\lfloor\ldots\rceil}_{k'},
  {\mathcal{E}}^{\lfloor\ldots\rceil}_{k}
  \right)
  =
  \mathop{\sum}\limits_{i_\inverj}
  {W}_{i_\inverj,i_j}^{\prime[\inverj,j]2}
  \tau^{[\inverj,j](e')} \left( {\mathcal{E}}_{k'}^{\prime\lfloor\ldots\rceil}{}_{i_\inverj} \right)
  \tau^{[\inverj,j](e)} \left( {\mathcal{E}}_{k}^{\lfloor\ldots\rceil}{}_{i_\inverj} \right)
 \end{array}
}\!\!\!\!\!\!\!\!\!\!\!\!\!\!\!\!\!\!\!\!}
\]
{\bf Higher-order Interactions:} Note that (\ref{eq:kerncrossapprox}) 
       includes terms that mix the contributions of multiple training vectors 
       ${\bf x}_{\{l\}}$.  If we remove these terms from (\ref{eq:kerncrossapprox}) 
       we obtain:
\begin{equation}
 \begin{array}{l}
   \changein {\bf f} \left( {\bf x}; \changein \allofW \right)
   \approx 
   {\sum}_{\bf k}
   \frac{\left( -\eta \right)^{k_0k_1\ldots}}{{\bf k}!}
   {\sum}_{p}
   \postver{K}_{\scriptscriptstyle{\rm LeNK}}^{\scriptscriptstyle{\rm NCT}}
   \left(
   \left\{ {\bf x} \right\}^{\lfloor{\bf k},{\bf 1},-\rceil},
   \mathcal{D}^{\lfloor{\bf 1},{\bf k},p{\bf 1}_{k_0k_1\ldots}\rceil}
   \right)
   \postver{\bg{\alpha}}_{\{p\}}^{\odot k_0k_1\ldots}
 \end{array}
\label{eq:kernfirstapprox}
\end{equation}
       This approximation is reasonable in the wide-network limit where the 
       size of the weight-step - and hence the size of ${\bg{\alpha}}$ - is 
       small, so the linear term ${\bf k}={\bf 1}$ dominates.  
       The ${\bf k}={\bf 1}$ term of $\postver{K}_{\scriptscriptstyle{\rm 
       LeNK}}^{\scriptscriptstyle{\rm NCT}}$ is equivalent to the NTK 
       up to scale factors; and moreover if we apply the approach of 
       \citep{Sai1} to (\ref{eq:kernfirstapprox}) we may obtain theorem 
       \ref{th:itshilbert} and the LiNK (again up to scale factors).

We note that higher-order statistical terms have been studied previously (eg. 
\citep{Bai2}), while off-diagonal terms may be viewed as an artifact of 
working in the non-over-parameterized regime; however to the best of our 
knowledge ours is the first paper to introduce the idea of kernel warping and 
describe its role in an exact representor theory.

\section{Conclusions and Future Directions}

In this paper we have presented two models of neural networks and neural 
network training for neural network of arbitrary width, depth and topology.  
First we presented an exact (non-approximated) RKBS model of the overall 
network in the form of a bilinear product between a data- and weight- feature 
map.  We have used this model to construct a bound on Rademacher complexity, 
which is particularly tight for feedforward ReLU networks.  The second model 
we have presented models the {\em change} in the neural network due to a 
bounded change in weights and biases.  This model cast the change in RKHS 
with the local-intrinsic neural kernel (LiNK).  We have shown that this can be 
used to bound Rademacher complexity for network adaptation.  Finally, for 
unregularized gradient descent, we have provided an exact representor theory 
in terms of a kernel (the local-extrinsic neural kernel (LeNK)), providing 
insight into the role of model shift (kernel warping), cross (off-diagonal) 
terms and higher-order interaction between training vectors.  We have 
demonstrated that the LiNK is, in a sense, an approximation of the LeNK when 
cross-products and kernel warping are neglected; and moreover that the NTK is 
a first-order approximation of the LiNK.  We have also discussed the role of 
weight initialization and implications for feedforward ReLU networks and 
residual networks (ResNets).
We foresee a number of possible extensions to this work.  Most obviously it 
would be informative to derive the LeNK and LiNK in closed form for common 
neural activations, which may subsequently be subject to spectral analysis.  
The exact representor theory needs to be generalized to arbitrary topologies, 
and the impact of regularization incorporated.  Finally, we intend to use our 
analysis to study the convergence of neural networks in terms of (corrected 
for higher-order terms) gradient flow.

\newpage
\bibliography{universal}
\bibliographystyle{icml2023}

\newpage
\appendix
\onecolumn

\section{Properties of Hermite polynomials} \label{sec:hermite}

The (probabilist's) Hermite polynomials are given by \citep{Abr2,Mor2,Olv1,Cou1}:
\[

\label{eq:hermpoly}
\end{equation}
and form an orthogonal basis of $L^2 (\mathbb{R},e^{-x^2})$.  Thus for any $f 
\in L^2 (\mathbb{R},e^{-x^2})$ there exist Hermite coefficients $a_0, a_1, 
\ldots \in \mathbb{R}$ (i.e. the Hermite transform of $f$) so that:
\[
 \begin{array}{rl}
  f \left( \zeta \right)
  &\!\!\!\!=
  \mathop{\sum}\limits_{k \in \mathbb{N}} a_{k} \He_k \left( \zeta \right) \;\;\; \forall \zeta \in \mathbb{R}
 \end{array}
\]
where:
\[
 \begin{array}{rl}
  a_{k} 
  &\!\!\!\!=
  \frac{1}{k!\sqrt{2\pi}} \int_{-\infty}^\infty f \left( \zeta \right) e^{-\frac{\zeta^2}{2}} \He_k \left( \zeta \right) d\zeta \\
 \end{array}
\]
This series representation converges everywhere on the real line.  Moreover 
\citep{Hil3,Boy3} this series converges on a strip $\mathbb{S}_{\rho} = \{ z 
\in \mathbb{C} : -\rho < {\rm Im} (z) < \rho \}$ of width $\rho$ about the 
real axis in the complex plane, where:\footnote{Note that \citep{Hil3,Boy3} 
use the normalized physicist's Hermite polynomials.  The additional scale 
factor here arises in the translation to the un-normalized probabilist's 
Hermite polynomials used here.}
\begin{equation}
 \begin{array}{rl}
  \rho = -\mathop{\rm lim \; sup}\limits_{k \to \infty} \frac{1}{\sqrt{2k+1}} \log \left( \left| \frac{a_{k}}{\sqrt{k!\sqrt{\pi}}} \right| \right)
 \end{array}
\label{eq:comphermroc}
\end{equation}

The Hermite numbers derive from the Hermite polynomials:\footnote{Typically 
the Hermite numbers are defined from the physicist's Hermite polynomials, but as 
we use the Probabilist's form we find these more convenient.}
\[
 %
\]
It is well known that (see e.g. \citep{Mor2}):
\[
 \begin{array}{rl}
  \He_k     \left( \zeta+\xi \right) &\!\!\!\!= \mathop{\sum}\limits_{0 \leq l \leq k} \binom{k}{l} \He_{k-l}     \left( \zeta \right) \xi^l \\
 \end{array}
\]
and so:
\[
 \begin{array}{rl}
  \He_k     \left( \zeta \right) &\!\!\!\!= \mathop{\sum}\limits_{0 \leq l \leq k} \binom{k}{l} \Henum_{k-l}     \zeta^l \\
 \end{array}
\]
It follows that, taking care not to change or order of summation (remember 
this is an alternating series):
\[
 \begin{array}{rl}
  f \left( \zeta \right)
  &\!\!\!\!=
  \mathop{\sum}\limits_{k=0}^\infty a_{k} \mathop{\sum}\limits_{l=0}^{k} \binom{k}{l} \Henum_{k-l} \zeta^l \\
 \end{array}
\]

Next we derive a helpful property involving {\em rectified} Hermite 
expansions.  Let $f \in L^2 (\mathbb{R},e^{-x^2})$, $f(0) = 0$, then:
\[
 \begin{array}{rll}
  f \left( x \right)
  &\!\!\!\!=
  \mathop{\sum}\limits_{k=1}^\infty a_k \He \left( x \right) 
  &\!\!\!\!=
  \mathop{\sum}\limits_{k=1}^\infty a_k
  \mathop{\sum}\limits_{l=1}^k \binom{k}{l} \Henum_{k-l} x^l \\
 \end{array}
\]
Denoting the imaginary element ${\tt i}$:
\[
 \begin{array}{rl}
  f \left( {\tt i} x \right)
  &\!\!\!\!=
  \mathop{\sum}\limits_{k=1}^\infty a_k
  \mathop{\sum}\limits_{l=1}^k \binom{k}{l} \Henum_{k-l} \left( {\tt i} x \right)^l \\
  &\!\!\!\!=
  \mathop{\sum}\limits_{k=1}^\infty a_k
  \mathop{\sum}\limits_{l=0}^{k-1} \binom{k}{k-l} \Henum_{l} \left( {\tt i} x \right)^{k-l} \\
  &\!\!\!\!=
  \mathop{\sum}\limits_{k=1}^\infty {\tt i}^k a_k
  \mathop{\sum}\limits_{l=0}^{k-1} \binom{k}{k-l} \left( {\tt i}^l \Henum_{l} \right) x^{k-l} \\
 \end{array}
\]
Recall that $\Henum_l = 0$ for $l=1,3,5,\ldots$, and $\sgn (\Henum_{2p}) = 
(-1)^p$.  Therefore
\[
 \begin{array}{rl}
  f \left( {\tt i} x \right)
  &\!\!\!\!=
  \mathop{\sum}\limits_{k=1}^\infty {\tt i}^k a_k
  \mathop{\sum}\limits_{l=0}^{k-1} \binom{k}{k-l} \left| \Henum_{l} \right| x^{k-l} \\
  &\!\!\!\!=
  \mathop{\sum}\limits_{k=1}^\infty {\tt i}^k a_k
  \mathop{\sum}\limits_{l=1}^k \binom{k}{l} \left| \Henum_{k-l} \right| x^{l} \\
 \end{array}
\]
and so:
\[
 \begin{array}{rl}
  {\rm Im} \left( f \left( {\tt i} x \right) \right)
  &\!\!\!\!=
  \mathop{\sum}\limits_{k=1,3,5,\ldots} \left( -1 \right)^{\lfloor \frac{k}{2} \rfloor} a_k
  \mathop{\sum}\limits_{l=1}^k \binom{k}{l} \left| \Henum_{k-l} \right| x^{l} \\
  {\rm Re} \left( f \left( {\tt i} x \right) \right)
  &\!\!\!\!=
  \mathop{\sum}\limits_{k=2,4,6,\ldots} \left( -1 \right)^{\lfloor \frac{k}{2} \rfloor}  a_k
  \mathop{\sum}\limits_{l=1}^k \binom{k}{l} \left| \Henum_{k-l} \right| x^{l} \\
 \end{array}
\]
\begin{equation}
 \begin{array}{rl}
  {\rm Im} \left( f \left( {\tt i} x \right) \right)
  +
  {\rm Re} \left( f \left( {\tt i} x \right) \right)
  &\!\!\!\!=
  \mathop{\sum}\limits_{k=1}^\infty \left( -1 \right)^{\lfloor \frac{k}{2} \rfloor} a_k
  \mathop{\sum}\limits_{l=1}^k \binom{k}{l} \left| \Henum_{k-l} \right| x^{l} \\
 \end{array}
\label{eq:replusimhermite}
\end{equation}

Finally we make some observations regarding derivatives that will be required later.  
Let $f \in L^2(\mathbb{R},e^{-x^2})$, $f(0) = 0$.  Then:
\[
 \begin{array}{rll}
  f \left( x \right)
  &\!\!\!\!=
  \mathop{\sum}\limits_{k=1}^\infty a_k
  \mathop{\sum}\limits_{l=1}^k \binom{k}{l} \Henum_{k-l} x^l \\
 \end{array}
\]
and subsequently:
\[
 \begin{array}{rll}
  f^{(1)} \left( x \right)
  &\!\!\!\!=
  \mathop{\sum}\limits_{k=1}^\infty a_k
  \mathop{\sum}\limits_{l=1}^k l \binom{k}{l} \Henum_{k-l} x^{l-1} \\
  &\!\!\!\!=
  \mathop{\sum}\limits_{k=1}^\infty a_k
  \mathop{\sum}\limits_{l=1}^k l \frac{k!}{l!(k-l)!} \Henum_{k-l} x^{l-1} \\
  &\!\!\!\!=
  \mathop{\sum}\limits_{k=1}^\infty a_k
  \mathop{\sum}\limits_{l=1}^k k \frac{(k-1)!}{(l-1)!(k-l)!} \Henum_{k-l} x^{l-1} \\
  &\!\!\!\!=
  \mathop{\sum}\limits_{k=1}^\infty a_k
  \mathop{\sum}\limits_{l=1}^k k \binom{k-1}{l-1} \Henum_{k-l} x^{l-1} \\
  &\!\!\!\!=
  \mathop{\sum}\limits_{k=0}^\infty (k+1) a_{k+1}
  \mathop{\sum}\limits_{l=0}^{k-1} \binom{k}{l} \Henum_{k-l} x^{l} \\
 \end{array}
\]
and:
\[
 \begin{array}{rll}
  f^{(2)} \left( x \right)
  &\!\!\!\!=
  \mathop{\sum}\limits_{k=0}^\infty (k+1) a_{k+1}
  \mathop{\sum}\limits_{l=1}^{k-1} (l-1) \binom{k}{l} \Henum_{k-l} x^{l-1} \\
  &\!\!\!\!=
  \mathop{\sum}\limits_{k=0}^\infty (k+1) a_{k+1}
  \mathop{\sum}\limits_{l=1}^{k-1} (l-1) \frac{k!}{l!(k-l)!} \Henum_{k-l} x^{l-1} \\
  &\!\!\!\!=
  \mathop{\sum}\limits_{k=0}^\infty (k+1) a_{k+1}
  \mathop{\sum}\limits_{l=1}^{k-1} k \frac{(k-1)!}{(l-l)!(k-l)!} \Henum_{k-l} x^{l-1} \\
  &\!\!\!\!=
  \mathop{\sum}\limits_{k=1}^\infty (k+1) a_{k+1}
  \mathop{\sum}\limits_{l=1}^{k-1} k \frac{(k-1)!}{(l-l)!(k-l)!} \Henum_{k-l} x^{l-1} \\
  &\!\!\!\!=
  \mathop{\sum}\limits_{k=0}^\infty (k+1) (k+2) a_{k+2}
  \mathop{\sum}\limits_{l=0}^{k-2} \binom{k}{l} \Henum_{k-l} x^{l} \\
 \end{array}
\]
and so on to:
\[
 \begin{array}{rll}
  f^{(n)} \left( x \right)
  &\!\!\!\!=
  \mathop{\sum}\limits_{k=0}^\infty \frac{(k+n)!}{k!} a_{k+n}
  \mathop{\sum}\limits_{l=0}^{k-n} \binom{k}{l} \Henum_{k-l} x^{l} \\
 \end{array}
\]

\subsection{Activation Functions} \label{app:actfuns}

Following the previous method we introduce our notation for the activation 
functions.  Recall $\tau^{[\inverj,j]} \in L^2 (\mathbb{R},e^{-x^2})$ by 
assumption.  Subsequently $\centau^{[\inverj,j]} \in L^2 (\mathbb{R}, 
e^{-x^2})$, where:
\begin{equation}
 \begin{array}{rl}
  \centau^{[\inverj,j]} \left( \zeta ; \xi \right)
  =
  \tau^{[\inverj,j]} \left( \xi + \zeta \right) - \tau^{[\inverj,j]} \left( \xi \right)
  &\!\!\!\!=
  \mathop{\sum}\limits_{k \in \mathbb{N}} a^{[\inverj,j]}_{(\xi)k} \He_k \left( \zeta \right) \;\;\; \forall \zeta \in \mathbb{R}_+ \\
  &\!\!\!\!=
  \mathop{\sum}\limits_{k=0}^\infty a^{[\inverj,j]}_{(\xi)k} \mathop{\sum}\limits_{l=0}^{k} \binom{k}{l} \Henum_{k-l} \zeta^l \\
  &\!\!\!\!=
  \mathop{\sum}\limits_{k=1}^\infty a^{[\inverj,j]}_{(\xi)k} \mathop{\sum}\limits_{l=1}^{k} \binom{k}{l} \Henum_{k-l} \zeta^l \\
 \end{array}
\label{eq:hermexpand}
\end{equation}
(in the final step we use that $\centau^{[\inverj,j]} (0;\xi) = 0$) with coefficients:
\[
 \begin{array}{rl}
  a^{[\inverj,j]}_{(\xi)k} 
  &\!\!\!\!=
  \frac{1}{k!\sqrt{2\pi}} \int_{-\infty}^\infty \centau^{[\inverj,j]} \left( \zeta ; \xi \right) e^{-\frac{\zeta^2}{2}} \He_k \left( \zeta \right) d\zeta \\
 \end{array}
\]
which converges on a strip $\mathbb{S}_{\rho_{(\xi)}^{[\inverj,j]}} = \{ z \in 
\mathbb{C} : -\rho_{(\xi)}^{[\inverj,j]} < {\rm Im} (z) < 
\rho_{(\xi)}^{[\inverj,j]} \}$ of width $\rho_{(\xi)}^{[\inverj,j]}$ about the 
real axis in the complex plane, where:
\[
 \begin{array}{rl}
  \rho_{(\xi)}^{[\inverj,j]} = -\mathop{\rm lim \; sup}\limits_{k \to \infty} \frac{1}{\sqrt{2k+1}} \log \left( \left| \frac{a^{[\inverj,j]}_{(\xi)k}}{\sqrt{k!\sqrt{\pi}}} \right| \right)
 \end{array}
\]

\subsection{Rectified Activation Functions} \label{sec:raf_fun}

Recall that the rectified activation functions are defined as:
\[
{\!\!\!\!\!\!\!\!{
 \begin{array}{rl}
  \abstau_\eta^{[\inverj,j]} \left( \zeta ; \xi, \xi' \right)
  &\!\!\!\!=
  \mathop{\sum}\limits_{k=1}^\infty
  \frac{a_{(\xi)k}^{[\inverj,j]} a_{(\xi')k}^{[\inverj,j]}}{\eta^{2k}}
  \mathop{\sum}\limits_{l=1}^k {\binom{k}{l}}^2 \Henum_{k-l}^2 \zeta^l \\
 \end{array}
}\!\!\!\!\!\!\!\!}
\]
where $\eta \in (0,1)$ is fixed.  
To understand the convergence of this function, observe that:
\[
 \begin{array}{rl}
  \abstau_\eta^{[\inverj,j]} \left( \zeta ; \xi, \xi' \right)
  &\!\!\!\!\leq
  \mathop{\max}\limits_{\xi'' \in \{\xi,\xi'\}}
  \mathop{\sum}\limits_{k=1}^\infty
  \mathop{\sum}\limits_{l=1}^k
  \left|
  \frac{a_{(\xi'')k}^{[\inverj,j]}}{\eta^k}
  \binom{k}{l}
  \Henum_{k-l}
  \sqrt{\zeta^l}
  \right|^2
 \end{array}
\]
which is the $2$-norm of a sequence.  Hence:
\[
 \begin{array}{rl}
  \abstau_\eta^{[\inverj,j]} \left( \zeta ; \xi, \xi' \right)
  &\!\!\!\!\leq
  \mathop{\max}\limits_{\xi'' \in \{\xi,\xi'\}}
  \left(
  \mathop{\sum}\limits_{k=1}^\infty
  \mathop{\sum}\limits_{l=1}^k
  \left|
  \frac{\left| a_{(\xi'')k}^{[\inverj,j]} \right|}{\eta^k}
  \binom{k}{l}
  \left| \Henum_{k-l} \right|
  \sqrt{\zeta^l}
  \right|
  \right)^2
 \end{array}
\]
Thus is suffices to study the convergence of:
\[
 \begin{array}{rl}
  \gamma^{[\inverj,j]} \left( \lambda ; \xi \right)
  &\!\!\!\!=
  \mathop{\sum}\limits_{k=1}^\infty
  \frac{\left| a_{(\xi)k}^{[\inverj,j]} \right|}{\eta^k}
  \mathop{\sum}\limits_{l=1}^k
  \binom{k}{l}
  \left| \Henum_{k-l} \right|
  \lambda^l
 \end{array}
\]
which in turn bounds:
\[
 \begin{array}{rl}
  \abstau_\eta^{[\inverj,j]} \left( \zeta ; \xi, \xi' \right)
  &\!\!\!\!\leq
  \max
  \left\{
  \gamma^{[\inverj,j]} \left( \sqrt{\zeta} ; \xi  \right)^2,
  \gamma^{[\inverj,j]} \left( \sqrt{\zeta} ; \xi' \right)^2
  \right\}
 \end{array}
\]

Using (\ref{eq:replusimhermite}):
\[
{\!\!\!\!{
 \begin{array}{rl}
  \gamma^{[\inverj,j]} \left( \lambda ; \xi \right)
  &\!\!\!\!=
  {\rm Re} \left(
  \overline{\gamma}^{[\inverj,j]} \left( {\tt i}\lambda ; \xi \right)
  \right)
  +
  {\rm Im} \left(
  \overline{\gamma}^{[\inverj,j]} \left( {\tt i}\lambda ; \xi \right)
  \right)
 \end{array}
}\!\!\!\!}
\]
where:
\[
 \begin{array}{rl}
  \gamma^{[\inverj,j]} \left( \lambda ; \xi \right)
  &\!\!\!\!=
  \mathop{\sum}\limits_{k=1}^\infty
  \frac{\left| a_{(\xi)k}^{[\inverj,j]} \right|}{\eta^k}
  \mathop{\sum}\limits_{l=1}^k
  \binom{k}{l}
  \Henum_{k-l}
  \lambda^l
 \end{array}
\]
which, using (\ref{eq:comphermroc}), is convergent for:
\[
 \begin{array}{l}
  \left| \lambda \right|
  <
  L_{{(\xi)}}^{[\inverj,j]}
  <
  -\mathop{\rm lim \; sup}\limits_{k \to \infty} \frac{1}{\sqrt{2k+1}} \left( \ln \left| \frac{a_{(\xi)k}^{[\inverj,j]}}{\eta^k} \right| \right) \\
 \end{array}
\]
whereon: 
\[
 \begin{array}{l}
  \left| \gamma^{[\inverj,j]} \left( \lambda ; \xi \right) \right|
  \leq
  M_{{(\xi)}}^{[\inverj,j]}
  =
  \gamma^{[\inverj,j]} \left( L_{{(\xi)}}^{[\inverj,j]} ; \xi \right)
 \end{array}
\]
and we conclude that $\abstau_\eta^{[\inverj,j]} (\zeta;\xi,\xi')$ is convergent 
for:
\[
 \begin{array}{rl}
  \left| \zeta \right|
  \leq
  \roctau_{{(\xi)\eta}}^{[\inverj,j]2}
  =
  \min
  \left\{
  -\mathop{\rm lim \; sup}\limits_{k \to \infty} \frac{1}{\sqrt{2k+1}} \ln \left| \frac{a_{(\xi)k}^{[\inverj,j]}}{\eta^k} \right|,
  -\mathop{\rm lim \; sup}\limits_{k \to \infty} \frac{1}{\sqrt{2k+1}} \ln \left| \frac{a_{(\xi)k}^{[\inverj,j]}}{\eta^k} \right|
  \right\}^2
 \end{array}
\]
whereon:
\[
 \begin{array}{l}
  \abstau_\eta^{[\inverj,j]} \left( \zeta;\xi,\xi' \right)
  \leq
  \abstaumax_{{(\xi)\eta}}^{[\inverj,j]2}
  =
  \abstau_\eta^{[\inverj,j]} \left( \roctau_{{(\xi)\eta}}^{[\inverj,j]2} ; \xi, \xi' \right)
 \end{array}
\]

The envelope is convergent for:
\[
 \begin{array}{l}
  \left| \zeta \right|
  <
  \roctau_{(\xpostm)\eta}^{[\inverj,j]2}
  =
  \mathop{\rm inf}\limits_{|\xi|,|\xi'| \leq \xpostm} \roctau_{{(\xi,\xi')\eta}}^{[\inverj,j]2} \\
 \end{array}
\]
whereon: 
\[
 \begin{array}{l}
  \abstaumax_{(\xpostm)\eta}^{[\inverj,j]2}
  =
  \abstau_\eta^{[\inverj,j]} \left ( \roctau_{(\xpostm)\eta}^{[\inverj,j]2}; \xpostm \right)
 \end{array}
\]

Finally:
\[
 \begin{array}{rl}
  \absT_\eta \left( \zeta \right)
  &\!\!\!\!=
  \mathop{\sum}\limits_{k=1}^\infty
  \eta^{2k}
  \mathop{\sum}\limits_{l=1}^k
  \zeta^l
  =
  \frac{\zeta}{1-\zeta}
  \mathop{\sum}\limits_{k=1}^\infty
  \eta^{2k}
  \left( 1 - \zeta^k \right)
  =
  \frac{\zeta}{1-\zeta}
  \left(
  \mathop{\sum}\limits_{k=1}^\infty
  \eta^{2k}
  -
  \mathop{\sum}\limits_{k=1}^\infty
  (\zeta \eta^2)^k
  \right) \\
  &\!\!\!\!=
  \frac{\zeta}{1-\zeta}
  \left(
  \frac{\eta^2}{1-\eta^2}
  -
  \frac{\zeta\eta^2}{1-\zeta\eta^2}
  \right)
 \end{array}
\]
is convergent $\forall |\zeta| < \rocT_\eta^{2} < 1$, with max 
value of $\absTmax_\eta^{2} = \frac{\rocT_\eta^2}{1-\rocT_\eta^2} 
(\frac{\eta^2}{1-\eta^2}-\frac{\eta^2\rocT_\eta^2}{1-\eta^2\rocT_\eta^2})$ 
thereon.

\subsection{ReLU Activation Function} \label{sec:hermxformact}

In this section we derive the Hermite-polynomial expansion of the centered 
ReLU activation function:
\[
 %
\]

We find it convenient to work in terms of the physicists Hermite 
polynomials $H_k$ to suit \citep{Gra1}.  Using this:
\[
{{
%
}}
\]
Using the recursion and derivative properties, for $k > 1$:
\[
 \begin{array}{rl}
  \zeta H_k \left( \zeta \right)
  &\!\!\!\!=
  \frac{1}{2} H_{k+1} \left( \zeta \right) + \frac{1}{2} H'_{k} \left( \zeta \right) \\
  &\!\!\!\!=
  \frac{1}{2} H_{k+1} \left( \zeta \right) + k H_{k-1} \left( \zeta \right) \\
 \end{array}
\]
and hence, using \citep[(7.373)]{Gra1}:
\[
 \begin{array}{rl}
  a_{(\xi)k}^{[{\rm ReLU}]}
  &\!\!\!\!=
  \frac{k+1}{\sqrt{\pi}} \frac{1}{\sqrt{2}^{k+1} (k+1)!} \int_{-\frac{\xi}{\sqrt{2}}}^\infty e^{-\zeta^2} H_{k+1} \left( \zeta \right) d\zeta
  +
  \frac{1}{\sqrt{\pi}} \frac{1}{\sqrt{2}^{k-1} (k-1)!} \int_{-\frac{\xi}{\sqrt{2}}}^\infty e^{-\zeta^2} H_{k-1} \left( \zeta \right) d\zeta
  + \ldots \\ &\;\;\;\; \ldots
  +
  \frac{1}{\sqrt{\pi}} \frac{1}{\sqrt{2}^k k!} \xi \int_{-\frac{\xi}{\sqrt{2}}}^\infty e^{-\zeta^2} H_k \left( \zeta \right) d\zeta
  -
  \frac{1}{\sqrt{\pi}} \frac{1}{\sqrt{2}^k k!} \left[ \xi \right]_+ \int_{-\infty}^\infty e^{-\zeta^2} H_k \left( \zeta \right) d\zeta \\

  &\!\!\!\!=
  \frac{k+1}{\sqrt{\pi}} \frac{1}{\sqrt{2}^{k+1} (k+1)!} \int_{-\frac{\xi}{\sqrt{2}}}^\infty e^{-\zeta^2} H_{k+1} \left( \zeta \right) d\zeta
  +
  \frac{1}{\sqrt{\pi}} \frac{1}{\sqrt{2}^{k-1} (k-1)!} \int_{-\frac{\xi}{\sqrt{2}}}^\infty e^{-\zeta^2} H_{k-1} \left( \zeta \right) d\zeta
  + \ldots \\ &\;\;\;\; \ldots
  +
  \frac{1}{\sqrt{\pi}} \frac{1}{\sqrt{2}^k k!} \xi \int_{-\frac{\xi}{\sqrt{2}}}^\infty e^{-\zeta^2} H_k \left( \zeta \right) d\zeta
  -
  \delta_{k,2p} \frac{1}{\sqrt{\pi}} \frac{1}{\sqrt{2}^{2p} (2p)!} \left[ \xi \right]_+ \int_{-\infty}^\infty e^{-\zeta^2} H_{2p} \left( \zeta \right) d\zeta \\

  &\!\!\!\!=
  \frac{k+1}{\sqrt{\pi}} \frac{1}{\sqrt{2}^{k+1} (k+1)!} \int_{-\frac{\xi}{\sqrt{2}}}^\infty e^{-\zeta^2} H_{k+1} \left( \zeta \right) d\zeta
  +
  \frac{1}{\sqrt{\pi}} \frac{1}{\sqrt{2}^{k-1} (k-1)!} \int_{-\frac{\xi}{\sqrt{2}}}^\infty e^{-\zeta^2} H_{k-1} \left( \zeta \right) d\zeta
  + \ldots \\ &\;\;\;\; \ldots
  +
  \frac{1}{\sqrt{\pi}} \frac{1}{\sqrt{2}^k k!} \xi \int_{-\frac{\xi}{\sqrt{2}}}^\infty e^{-\zeta^2} H_k \left( \zeta \right) d\zeta
 \end{array}
\]
Using \citep[(7.373)]{Gra1} we have:
\[
 \begin{array}{rl}
  a_{(\xi)k}^{[{\rm ReLU}]}
  &\!\!\!\!=
  \frac{1}{\sqrt{\pi}}
  \frac{1}{\sqrt{2}^{k+1} (k+1)!}
  (k+1) 
  \left( 
  e^{-\frac{\xi^2}{2}} H_k \left( -\frac{\xi}{\sqrt{2}} \right) 
  -
  e^{-\frac{\infty^2}{2}} H_k \left( \frac{\infty}{\sqrt{2}} \right)
  \right)
  +
  \ldots \\ &\;\;\;\; \ldots
  \frac{1}{\sqrt{\pi}}
  \frac{1}{\sqrt{2}^{k-1} (k-1)!}
  \left( 
  e^{-\frac{\xi^2}{2}} H_{k-2} \left( -\frac{\xi}{\sqrt{2}} \right) 
  -
  e^{-\frac{\infty^2}{2}} H_{k-2} \left( \frac{\infty}{\sqrt{2}} \right)
  \right)
  +
  \ldots \\ &\;\;\;\; \ldots
  \frac{1}{\sqrt{\pi}}
  \frac{1}{\sqrt{2}^k k!} \xi
  \left( 
  e^{-\frac{\xi^2}{2}} H_{k-1} \left( -\frac{\xi}{\sqrt{2}} \right) 
  -
  e^{-\frac{\infty^2}{2}} H_{k-1} \left( \frac{\infty}{\sqrt{2}} \right)
  \right)
\\

  &\!\!\!\!\!\!\!\!\!\!\!\!\!\!\!\!\!\!\!\!=
  \frac{1}{\sqrt{2\pi}}
  e^{-\frac{\xi^2}{2}}
  \left(
  \frac{k+1}{\sqrt{2}^{k} (k+1)!}
  H_k \left( -\frac{\xi}{\sqrt{2}} \right) 
  +
  \frac{1}{\sqrt{2}^{k-2} (k-1)!}
  H_{k-2} \left( -\frac{\xi}{\sqrt{2}} \right) 
  +
  \frac{1}{\sqrt{2}^{k-1} k!} \xi
  H_{k-1} \left( -\frac{\xi}{\sqrt{2}} \right) 
  \right)
 \end{array}
\]
If $k = 2p$ and $p>0$ then, noting that $H_k(0) = \sqrt{2}^k \Henum_k$:
\[
{\!\!\!\!\!\!\!\!\!\!\!\!\!\!\!\!\!\!\!\!\!\!\!\!{
 \begin{array}{rl}
  a_{(\xi)2p}^{[{\rm ReLU}]}
  &\!\!\!\!=
  \frac{1}{\sqrt{2\pi}}
  e^{-\frac{\xi^2}{2}}
  \left(
  \frac{1}{\sqrt{2}^{2p} (2p+1)!}
  (2p+1) H_{2p} \left( -\frac{\xi}{\sqrt{2}} \right) 
  +
  \frac{1}{\sqrt{2}^{2p-2} (2p-1)!}
  H_{2p-2} \left( -\frac{\xi}{\sqrt{2}} \right) 
  +
  \frac{1}{\sqrt{2}^{2p-1} (2p)!} \xi
  H_{2p-1} \left( -\frac{\xi}{\sqrt{2}} \right) 
  \right) \\
  &\!\!\!\!=
  \frac{1}{\sqrt{2\pi}}
  e^{-\frac{\xi^2}{2}}
  \left(
  \frac{1}{\sqrt{2}^{2p} (2p+1)!}
  (2p+1) H_{2p} \left( \frac{\xi}{\sqrt{2}} \right) 
  +
  \frac{1}{\sqrt{2}^{2p-2} (2p-1)!}
  H_{2p-2} \left( \frac{\xi}{\sqrt{2}} \right) 
  -
  \frac{1}{\sqrt{2}^{2p-1} (2p)!} \xi
  H_{2p-1} \left( \frac{\xi}{\sqrt{2}} \right) 
  \right) \\
  &\!\!\!\!=
  \frac{(-1)^{p+1}}{\sqrt{2\pi} (2p-1) 2^p p!}
  e^{-\frac{\xi^2}{2}}
  \left(
  \frac{(-1)^{p+1} (2p-1) p!}{(2p)!}
  H_{2p} \left( \frac{\xi}{\sqrt{2}} \right) 
  +
  \frac{(-1)^{p+1}2 p!}{(2(p-1))!}
  H_{2p-2} \left( \frac{\xi}{\sqrt{2}} \right) 
  -
  \frac{(-1)^{p+1} (p-1)!}{\sqrt{2}(2(p-1))!} \xi
  H_{2p-1} \left( \frac{\xi}{\sqrt{2}} \right) 
  \right) \\

  &\!\!\!\!\left(=
  \frac{(-1)^{p+1}}{\sqrt{2\pi} (2p-1) 2^p p!}
  e^{-\frac{\xi^2}{2}}
  \left(
  \frac{(-1)^{p+1} (2p-1) p!}{(2p)!}
  H_{2p} \left( 0 \right) 
  +
  \frac{(-1)^{p+1}2 p!}{(2(p-1))!}
  H_{2p-2} \left( 0 \right) 
  \right), \mbox{ if } \xi = 0 \right) \\
  &\!\!\!\!\left(=
  \frac{(-1)^{p+1}}{\sqrt{2\pi} (2p-1) 2^p p!}
  e^{-\frac{\xi^2}{2}}
  \left(
  \frac{(-1)^{p+1} (2p-1) p!}{(2p)!}
  2^p \Henum_{2p}
  +
  \frac{(-1)^{p+1}2 p!}{(2(p-1))!}
  2^{p-1} \Henum_{2p-2}
  \right), \mbox{ if } \xi = 0 \right) \\
  &\!\!\!\!\left(=
  \frac{(-1)^{p+1}}{\sqrt{2\pi} (2p-1) 2^p p!}
  e^{-\frac{\xi^2}{2}}
  \left(
  \frac{(-1)^{p+1} (2p-1) p!}{(2p)!}
  2^p \frac{(-1)^p (2p)!}{2^p p!}
  +
  \frac{(-1)^{p+1}2 p!}{(2(p-1))!}
  2^{p-1} \frac{(-1)^{p+1} (2p-2)!}{(p-1)! 2^{p-1}}
  \right), \mbox{ if } \xi = 0 \right) \\
\end{array}
}}
\]
If $k = 2p+1$ and $p>0$ then:
\[
{\!\!\!\!\!\!\!\!\!\!\!\!\!\!\!\!\!\!\!\!\!\!\!\!{
 \begin{array}{rl}
  a_{(\xi)2p+1}^{[{\rm ReLU}]}
  &\!\!\!\!=
  \frac{1}{\sqrt{2\pi}}
  e^{-\frac{\xi^2}{2}}
  \left(
  \frac{1}{\sqrt{2}^{2p+1} (2p+2)!}
  (2p+2) H_{2p+1} \left( -\frac{\xi}{\sqrt{2}} \right) 
  +
  \frac{1}{\sqrt{2}^{2p-1} (2p)!}
  H_{2p-1} \left( -\frac{\xi}{\sqrt{2}} \right) 
  +
  \frac{1}{\sqrt{2}^{2p-2} (2p-1)!} \xi
  H_{2p-2} \left( -\frac{\xi}{\sqrt{2}} \right) 
  \right) \\
  &\!\!\!\!=
  \frac{1}{\sqrt{2\pi}}
  e^{-\frac{\xi^2}{2}}
  \left(
  -\frac{1}{\sqrt{2}^{2p+1} (2p+2)!}
  (2p+2) H_{2p+1} \left( \frac{\xi}{\sqrt{2}} \right) 
  -
  \frac{1}{\sqrt{2}^{2p-1} (2p)!}
  H_{2p-1} \left( \frac{\xi}{\sqrt{2}} \right) 
  +
  \frac{1}{\sqrt{2}^{2p-2} (2p-1)!} \xi
  H_{2p-2} \left( \frac{\xi}{\sqrt{2}} \right) 
  \right) \\
  &\!\!\!\!=
  \frac{1}{\sqrt{2\pi}}
  e^{-\frac{\xi^2}{2}}
  \left(
  -\frac{\sqrt{2}}{2^{p+1} (2p+1)!}
  H_{2p+1} \left( \frac{\xi}{\sqrt{2}} \right) 
  -
  \frac{\sqrt{2}}{2^{p} (2p)!}
  H_{2p-1} \left( \frac{\xi}{\sqrt{2}} \right) 
  +
  \frac{1}{2^{p-1} (2p-1)!} \xi
  H_{2p-2} \left( \frac{\xi}{\sqrt{2}} \right) 
  \right) \\

  &\!\!\!\!\left( = 0 \mbox{ if } \xi = 0 \right) \\
 \end{array}
}}
\]

For the cases $k=0,1$ we use the result:
\[
 \begin{array}{rl}
  \int_a^b x^m e^{-x^2} dx =
  \frac{1}{2}
  \Gamma \left( \frac{m+1}{2}, a^2 \right)
  -
  \frac{1}{2}
  \Gamma \left( \frac{m+1}{2}, b^2 \right) 
 \end{array}
\]
and so:
\[
 \begin{array}{rl}
  \int_a^\infty x^m e^{-x^2} dx =
  \frac{1}{2}
  \Gamma \left( \frac{m+1}{2}, a^2 \right)
 \end{array}
\]
In the case $k=0$:
\[
 \begin{array}{rl}
  a_{(\xi)0}^{[{\rm ReLU}]}
  &\!\!\!\!=
  \sqrt{\frac{2}{\pi}}
  \int_{-\frac{\xi}{\sqrt{2}}}^\infty \zeta e^{-\zeta^2} d\zeta
  +
  \frac{1}{\sqrt{\pi}} \xi
  \int_{-\frac{\xi}{\sqrt{2}}}^\infty e^{-\zeta^2} d\zeta
  -
  \left[ \xi \right]_+ \\

  &\!\!\!\!=
  \frac{1}{\sqrt{2\pi}}
  \Gamma \left( 1, \frac{\xi^2}{2} \right)
  +
  \frac{1}{2\sqrt{\pi}} \xi
  \Gamma \left( \frac{1}{2}, \frac{\xi^2}{2} \right)
  -
  \left[ \xi \right]_+ \\

  &\!\!\!\!\left( = \frac{1}{\sqrt{2\pi}} \mbox{ if } \xi = 0 \right) \\
 \end{array}
\]
and in the case $k=1$:
\[
 \begin{array}{rl}
  a_{(\xi)1}^{[{\rm ReLU}]}
  &\!\!\!\!=
  \frac{2}{\sqrt{\pi}}
  \int_{-\frac{\xi}{\sqrt{2}}}^\infty \zeta^2 e^{-\zeta^2} d\zeta
  +
  \sqrt{\frac{2}{\pi}} \xi
  \int_{-\frac{\xi}{\sqrt{2}}}^\infty \zeta e^{-\zeta^2} d\zeta \\

  &\!\!\!\!=
  \frac{1}{\sqrt{\pi}}
  \Gamma \left( \frac{3}{2}, \frac{\xi^2}{2} \right)
  +
  \frac{1}{\sqrt{2\pi}} \xi
  \Gamma \left( 1, \frac{\xi^2}{2} \right) \\

  &\!\!\!\!\left(= \frac{1}{2} \mbox{ if } \xi = 0 \right) \\
 \end{array}
\]

Next we derive the magnitude functions for the ReLU.  Using integration by 
parts, we see that:
\[
 \begin{array}{rl}
  \frac{1}{\sqrt{2\pi}} \int_{c}^x \frac{1}{\zeta^2} \left( e^{\frac{1}{2} \zeta^2} - 1 \right) d\zeta
  &\!\!\!\!= \frac{1}{\sqrt{2\pi}} \frac{1}{\sqrt{2}} \int_{c}^x \frac{2}{\zeta^2} \left( e^{\frac{1}{2} \zeta^2} - 1 \right) d\frac{\zeta}{\sqrt{2}} \\
  &\!\!\!\!= \frac{1}{\sqrt{2\pi}} \frac{1}{\sqrt{2}} \int_{\frac{c}{\sqrt{2}}}^{\frac{x}{\sqrt{2}}} \frac{1}{\zeta^2} \left( e^{\zeta^2} - 1 \right) d\zeta \\
  &\!\!\!\!= -\frac{1}{\sqrt{2\pi}} \frac{1}{x} \left( e^{\frac{1}{2} x^2} - 1 \right) + \frac{1}{\sqrt{2\pi}} \frac{1}{c} \left( e^{\frac{1}{2} c^2} - 1 \right) + \frac{1}{\sqrt{\pi}} \int_{\frac{c}{\sqrt{2}}}^{\frac{x}{\sqrt{2}}} e^{\zeta^2} d\zeta \\
  &\!\!\!\!= -\frac{1}{\sqrt{2\pi}} \frac{1}{x} \left( e^{\frac{1}{2} x^2} - 1 \right) + \frac{1}{\sqrt{2\pi}} \frac{1}{c} \left( e^{\frac{1}{2} c^2} - 1 \right) + \frac{1}{2} \frac{2}{\sqrt{\pi}} \int_{\frac{c}{\sqrt{2}}}^{\frac{x}{\sqrt{2}}} e^{\zeta^2} d\zeta \\
  &\!\!\!\!= -\frac{1}{\sqrt{2\pi}} \frac{1}{x} \left( e^{\frac{1}{2} x^2} - 1 \right) + \frac{1}{2} {\rm erfi} \left( \frac{x}{\sqrt{2}} \right)
           - \frac{1}{2} \left( {\rm erfi} \left( \frac{c}{\sqrt{2}} \right) - \frac{1}{\sqrt{2\pi}} \frac{2}{c} \left( e^{\frac{1}{2} c^2} - 1 \right) \right) \\
 \end{array}
\]
So:
\[
 \begin{array}{rl}
  \mathop{\sum}\limits_{k=1}^\infty \left| a^{[{\rm ReLU}]}_k \right| x^k
  &\!\!\!\!= \frac{1}{2} x + \frac{1}{\sqrt{2\pi}} \mathop{\sum}\limits_{p=1}^\infty \frac{x^{2p}}{(2p-1) 2^p p!} \\
  &\!\!\!\!= \frac{1}{2} x + \frac{1}{\sqrt{2\pi}} x \mathop{\sum}\limits_{p=1}^\infty \frac{x^{2p-1}}{(2p-1) 2^p p!} \\
  &\!\!\!\!= \frac{1}{2} x + \frac{1}{\sqrt{2\pi}} x \int_c^x \left( \frac{\partial}{\partial \zeta} \mathop{\sum}\limits_{p=1}^\infty \frac{\zeta^{2p-1}}{(2p-1) 2^p p!} \right) d\zeta \\
  &\!\!\!\!= \frac{1}{2} x + \frac{1}{\sqrt{2\pi}} x \int_c^x \left( \mathop{\sum}\limits_{p=1}^\infty \frac{\zeta^{2p-2}}{2^p p!} \right) d\zeta \\
  &\!\!\!\!= \frac{1}{2} x + \frac{1}{2\sqrt{2\pi}} x \int_c^x \left( \mathop{\sum}\limits_{p=1}^\infty \frac{1}{p!} \left( \frac{1}{2} \zeta^2 \right)^{p-1} \right) d\zeta \\
  &\!\!\!\!= \frac{1}{2} x + \frac{1}{\sqrt{2\pi}} x \int_c^x \frac{1}{\zeta^2} \left( \mathop{\sum}\limits_{p=1}^\infty \frac{1}{p!} \left( \frac{1}{2} \zeta^2 \right)^p \right) d\zeta \\
  &\!\!\!\!= \frac{1}{2} x + \frac{1}{\sqrt{2\pi}} x \int_c^x \frac{1}{\zeta^2} \left( e^{\frac{1}{2} \zeta^2} - 1 \right) d\zeta \\
  &\!\!\!\!= \frac{1}{2} x \left( {\rm erfi} \left( \frac{x}{\sqrt{2}} \right) + 1 - {\rm erfi} \left( \frac{c}{\sqrt{2}} \right) + \frac{1}{\sqrt{2\pi}} \frac{2}{c} \left( e^{\frac{1}{2} c^2} - 1 \right) \right) + \frac{1}{\sqrt{2\pi}} \left( 1 - e^{\frac{1}{2} x^2} \right) \\
 \end{array}
\]
Select $c$ so that the first derivative is $\frac{1}{2}x$:
\[
 \begin{array}{rl}
  -{\rm erfi} \left( \frac{c}{\sqrt{2}} \right) + \frac{1}{\sqrt{2\pi}} \frac{2}{c} \left( e^{\frac{1}{2} c^2} - 1 \right) = 0 \mbox{ if } c = 0
 \end{array}
\]
Hence:
\begin{equation}
 \begin{array}{rl}
  \abssig^{[{\rm ReLU}]} \left( x \right)
  &\!\!\!\!\triangleq \mathop{\sum}\limits_{k=1}^\infty \left| a^{[{\rm ReLU}]}_k \right| \left( 1 + x \right)^k - \mathop{\sum}\limits_{k=1}^\infty \left| a^{[{\rm ReLU}]}_k \right| \\
  &\!\!\!\!= \frac{1}{2} \left( 1+x \right) \left( {\rm erfi} \left( \frac{1+x}{\sqrt{2}} \right) + 1 \right) + \frac{1}{\sqrt{2\pi}} \left( 1 - e^{\frac{1}{2} \left( 1+x \right)^2} \right) -
             \frac{1}{2}                    \left( {\rm erfi} \left( \frac{1  }{\sqrt{2}} \right) + 1 \right) - \frac{1}{\sqrt{2\pi}} \left( 1 - e^{\frac{1}{2}           }            \right) \\

  &\!\!\!\!= \frac{1}{2} x \left( {\rm erfi} \left( \frac{1+x}{\sqrt{2}} \right) + 1 \right) + \frac{1}{\sqrt{2\pi}} \left( e^{\frac{1}{2} } - e^{\frac{1}{2} \left( 1+x \right)^2} \right) +
             \frac{1}{2}   \left( {\rm erfi} \left( \frac{1+x}{\sqrt{2}} \right) - {\rm erfi} \left( \frac{1}{\sqrt{2}} \right) \right) \\
 \end{array}
\label{eq:sigmabar}
\end{equation}

\begin{figure}
\begin{center}
\includegraphics[width=0.45\textwidth]{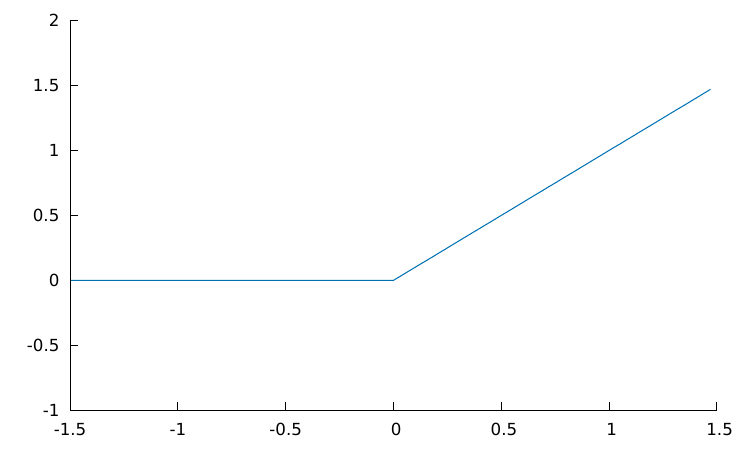}
\includegraphics[width=0.45\textwidth]{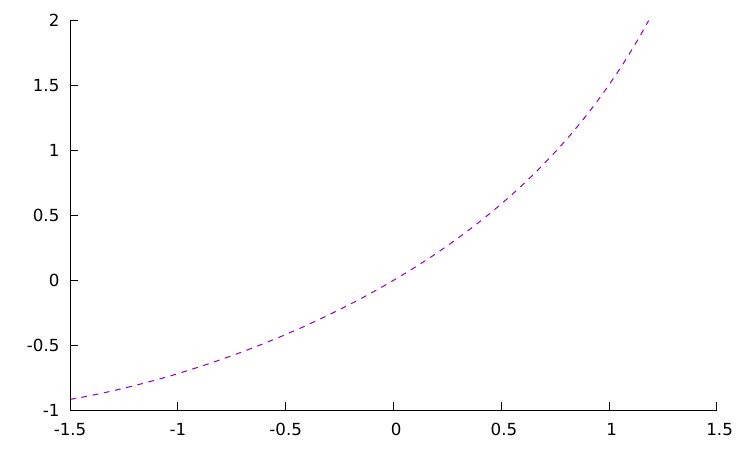}
\includegraphics[width=0.45\textwidth]{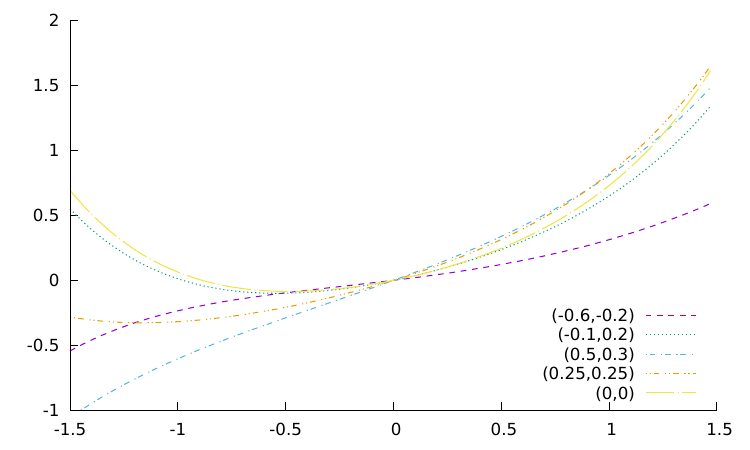}
\includegraphics[width=0.45\textwidth]{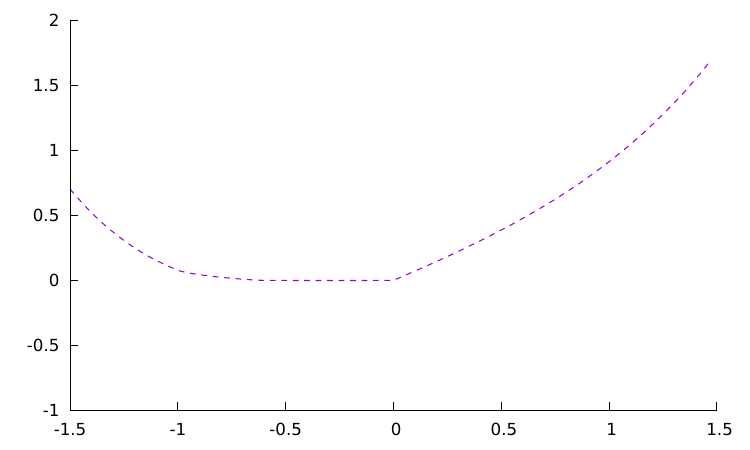}
\end{center}
 \caption{The ReLU magnitude activation and associated 
          functions - ReLU activation $\tau^{[{\rm ReLU}]}$ (top left), 
          magnitude $\abssig^{[{\rm ReLU}]}$ (top right), rectified 
          activation $\abstau_\eta^{[{\rm ReLU}]}$ for various $\xi,\xi'$ pairs 
          (bottom left) and its envelope (bottom right) for $\xi,\xi' \in 
          [-2,2]$.  Note that $\eta = 0.75$ for all plots.}
\label{fig:rectactrelu}
\end{figure}

\section{Proof of the Global Dual Model} \label{append:dualform}

Here we prove the validity of the global dual model presented in the 
paper.  Recall that the dual model has the form (Theorem \ref{th:netexpand}, 
equation (\ref{eq:ourdualmodel})):
\begin{equation}
  \begin{array}{rl}
   {\bf f} \left( {\bf x}; \allofW \right) 
   &\!\!\!\!= \leftbf \Postver{\bg{\Psi}} \left( {\allofW} \right), \Postver{\bg{\phi}} \left( {\bf x} \right) \rightbf_{\postver{\bf g}} \\
  \end{array}
\label{eq:ourgoalglobal}
\end{equation}
where, as per Figure \ref{fig:featmapdef}, 
$\Postver{\bg{\Psi}} = \Postver{\bg{\Psi}}^{[D-1]}$, 
$\Postver{\bg{\phi}} = \Postver{\bg{\phi}}^{[D-1]}$, 
$\Postver{\bf g} = \Postver{\bf g}^{[D-1]}$ and, given the base case 
$\Postver{\bg{\Psi}}^{[-1]} ({\allofW}) = {\bf 1}_n$, 
$\Postver{\bg{\phi}}^{[-1]} ({\bf x}) = {\bf x}$, 
$\postver{\bf g}^{[-1]} = {\bf 1}_{n}$, the recursive definition of the feature 
maps and metric is proposed:
\begin{equation}
 \begin{array}{rl}
  \Prever{\bg{\Psi}}_{:i_{\inverj}}^{[\inverj,j]} \left( {\allofW} \right)
  &\!\!\!\!= \left[
             \left| a_{(0)k}^{[\inverj,j]} \right|^{\frac{1}{2}}\!
             \left[
             {\binom{k}{l}}^{\frac{1}{2}}\!
             \left(
             \sqrt{\frac{\shada^2+1}{\abssig^{[\inverj,j]-1} \left( \prever{\phi}^{[\inverj,j]2} \right)}}
             \Postver{\bg{\Psi}}_{:i_{\inverj}}^{[\inverj]} \left( {\allofW} \right)
             \right)^{\otimes l}
             \right]_{1 \leq l \leq k}
             \right]_{k \geq 1} \\
  \Prever{\bg{\phi}}^{[\inverj,j]} \left( {\bf x} \right)
  &\!\!\!\!= \left[
             a_{(0)k}^{[\inverj,j]\left<\frac{1}{2}\right>}\!
             \left[
             {\binom{k}{l}}^{\frac{1}{2}}\!
             \left(
             \sqrt{\frac{\abssig^{[\inverj,j]-1} \left( \prever{\phi}^{[\inverj,j]2} \right)}{\shada^2+1}}
             \Postver{\bg{\phi}}^{[\inverj]} \left( {\bf x} \right)
             \right)^{\otimes l}
             \right]_{1 \leq l \leq k}
             \right]_{k \geq 1} \\
  \prever{\bf g}^{[\inverj,j]}
  &\!\!\!\!= \left[
             \left[
             \Henum_{k-l}
             \postver{\bf g}^{[\inverj]\otimes l}
             \right]_{1 \leq l \leq k}
             \right]_{k \geq 1}
 \end{array}
\label{eq:edgecalc}
\end{equation}
$\forall \inverj \in \Inver{\mathbb{P}}^{[j]}$ 
(the feature map transforms associated with the edges of the graph) and:
\begin{equation}
 \begin{array}{rl}
  \Postver{\bg{\Psi}}^{[j]} \left( {\allofW} \right)
  &\!\!\!\!= \left[ \begin{array}{c}
                    {\bf b}^{[j]\tsp} + {\bg{\upsilon}}_\tau^{[j]\tsp} \\
                    \inver{p}^{[j]} \mathop{\rm diag}\limits_{{{\inverj \in \Inver{\mathbb{P}}^{[j]}}}} \left( \prever{\phi}^{[\inverj,j]} \Prever{\bg{\Psi}}^{[\inverj,j]} \left( {\allofW} \right) \right) {\bf W}^{[j]} \\
             \end{array} \right] \\

  \Postver{\bg{\phi}}^{[j]} \left( {\bf x} \right)
  &\!\!\!\!= \left[ \begin{array}{c}
                    \shada \\
                    \frac{1}{\inver{p}^{[j]}} \left[ \frac{1}{\prever{\phi}^{[\inverj,j]}} {\Prever{\bg{\Phi}}}^{[\inverj,j]} \left( {\bf x} \right) \right]_{\inverj \in \Inver{\mathbb{P}}^{[j]}} \\
             \end{array} \right] \\

  \postver{\bf g}^{[j]}
  &\!\!\!\!= \left[ \begin{array}{c}
                    1 \\
                    \left[ \prever{\bf g}^{[\inverj,j]} \right]_{\inverj \in \Inver{\mathbb{P}}^{[j]}} \\
             \end{array} \right]
 \end{array}
\label{eq:nodecalc}
\end{equation}
(the feature map transforms associated with the nodes of the graph) where 
${\bg{\upsilon}}_\tau^{[j]} = \sum_{\inverj \in \Inver{\mathbb{P}}^{[j]}} 
\frac{\tau^{[\inverj,j]} (0)}{\shada} {\bf W}^{[\inverj,j]\tsp} {\bf 
1}_{H^{[\inverj]}}$.  Our approach to demonstrating that this is true is to 
prove that, given some input ${\bf x} \in \mathbb{X}$ then, for all edges 
$(\inverj \to j)$:
\begin{equation}
  \begin{array}{rl}
   \prever{\bf x}^{[\inverj,j]} - {\bf 1}_{H^{[\inverj]}} \tau^{[\inverj,j]} \left( 0 \right)
   &\!\!\!\!=
   \leftbf \Prever{\bg{\Psi}}^{[\inverj,j]} \left( {\allofW} \right), \Prever{\bg{\phi}}^{[\inverj,j]} \left( {\bf x} \right) \rightbf_{\prever{\bf g}^{[\inverj,j]}} \\
  \end{array}
\label{eq:edgefact}
\end{equation}
and likewise for all nodes $j$:
\begin{equation}
  \begin{array}{rl}
   \postver{\bf x}^{[j]}
   &\!\!\!\!=
   \leftbf \Postver{\bg{\Psi}}^{[j]} \left( {\allofW} \right), \Postver{\bg{\phi}}^{[j]} \left( {\bf x} \right) \rightbf_{\postver{\bf g}^{[j]}} \\
  \end{array}
\label{eq:nodefact}
\end{equation}

{\bf Base case:} By the definition of the base case, we have:
\[
  \begin{array}{rl}
   \postver{\bf x}^{[-1]}
   &\!\!\!\!=
   \leftbf \Postver{\bg{\Psi}}^{[-1]} \left( {\allofW} \right), \Postver{\bg{\phi}}^{[-1]} \left( {\bf x} \right) \rightbf_{\postver{\bf g}^{[-1]}}
   =
   {\bf x} \\
  \end{array}
\]

{\bf Node case:} Assume (\ref{eq:edgefact}) is true.  Then, using 
(\ref{eq:nodecalc}), we have that:
\[
  \begin{array}{rl}
   \leftbf \Postver{\bg{\Psi}}^{[j]} \left( {\allofW} \right), \Postver{\bg{\phi}}^{[j]} \left( {\bf x} \right) \rightbf_{\postver{\bf g}^{[j]}}

   &\!\!\!\!=
   \shada {\bf b}^{[j]}
   +
   \mathop{\sum}\limits_{\inverj \in \Inver{\mathbb{P}}^{[j]}} 
   \left(
   {\bf W}^{[\inverj,j]\tsp} {\bf 1}_{H^{[\inverj]}} \tau^{[\inverj,j]} \left( 0 \right)
   +
   {\bf W}^{[\inverj,j]\tsp} \prever{\bf x}^{[\inverj,j]}
   -
   {\bf W}^{[\inverj,j]\tsp} {\bf 1}_{H^{[\inverj]}} \tau^{[\inverj,j]} \left( 0 \right)
   \right) \\

   &\!\!\!\!=
   \shada {\bf b}^{[j]}
   +
   \mathop{\sum}\limits_{\inverj \in \Inver{\mathbb{P}}^{[j]}}
   {\bf W}^{[\inverj,j]\tsp} \prever{\bf x}^{[\inverj,j]} \\

   &\!\!\!\!=
   \postver{\bf x}^{[j]} \\
  \end{array}
\]

{\bf Edge case:} Assume (\ref{eq:nodefact}) is true.  Then, using 
(\ref{eq:edgecalc}), we have that:
\[
 \begin{array}{rl}
  \leftbf \Prever{\bg{\Psi}}^{[\inverj,j]} \left( {\allofW} \right), \Prever{\bg{\phi}}^{[\inverj,j]} \left( {\bf x} \right) \rightbf_{\prever{\bf g}^{[\inverj,j]}}
   &\!\!\!\!=
   \left[
  \leftbf \Prever{\bg{\Psi}}_{:i_\inverj}^{[\inverj,j]} \left( {\allofW} \right), \Prever{\bg{\phi}}^{[\inverj,j]} \left( {\bf x} \right) \rightbf_{\prever{\bf g}^{[\inverj,j]}}
   \right]_{i_{\inverj}} \\

   &\!\!\!\!=
   \left[
   \mathop{\sum}\limits_{k \geq 1}
   a_{(0)k}^{[\inverj,j]}
   \mathop{\sum}\limits_{1 \leq l \leq k}
   \binom{k}{l} \Henum_{k-l}
   \leftbf \Postver{\bg{\Psi}}_{:i_\inverj}^{[j]} \left( {\allofW} \right)^{\otimes l}, \Postver{\bg{\phi}}^{[j]} \left( {\bf x} \right)^{\otimes l} \rightbf_{\postver{\bf g}^{[j]\otimes l}}
   \right]_{i_{\inverj}} \\

   &\!\!\!\!=
   \left[
   \mathop{\sum}\limits_{k \geq 1}
   a_{(0)k}^{[\inverj,j]}
   \mathop{\sum}\limits_{1 \leq l \leq k}
   \binom{k}{l} \Henum_{k-l}
   \leftbf \Postver{\bg{\Psi}}_{:i_\inverj}^{[j]} \left( {\allofW} \right), \Postver{\bg{\phi}}^{[j]} \left( {\bf x} \right) \rightbf_{\postver{\bf g}^{[j]}}^l
   \right]_{i_{\inverj}} \\

   &\!\!\!\!=
   \left[
   \mathop{\sum}\limits_{k \geq 1}
   a_{(0)k}^{[\inverj,j]}
   \mathop{\sum}\limits_{1 \leq l \leq k}
   \binom{k}{l} \Henum_{k-l}
   \postver{x}_{i_\inverj}^{[\inverj]l}
   \right]_{i_{\inverj}} \\

   &\!\!\!\!=
   \left[
   \mathop{\sum}\limits_{k \geq 0}
   a_{(0)k}^{[\inverj,j]}
   \mathop{\sum}\limits_{0 \leq l \leq k}
   \binom{k}{l} \Henum_{k-l}
   \postver{x}_{i_\inverj}^{[\inverj]l}
   -
   \mathop{\sum}\limits_{k \geq 0}
   a_{(0)k}^{[\inverj,j]}
   \mathop{\sum}\limits_{0 \leq l \leq k}
   \binom{k}{l} \Henum_{k-l}
   0^l 
   \right]_{i_{\inverj}} \\

   &\!\!\!\!=
   \left[
   \tau^{[\inverj,j]} \left( \postver{x}_{i_\inverj}^{[\inverj]} \right) - \tau^{[\inverj,j]} \left( 0 \right)
   \right]_{i_{\inverj}} \\

   &\!\!\!\!=
   \prever{\bf x}^{[\inverj,j]} - {\bf 1}_{H^{[\inverj]}} \tau^{[\inverj,j]} \left( 0 \right) \\
 \end{array}
\]

The desired result (\ref{eq:ourgoalglobal}) then follows by identifying the 
output node $j=D-1$.

\subsection{Norm-Bounds for the Global Dual Model}

Our proof of the norm-bounds of the global model follows the same model as our 
proof of the validity of said model.  We want to prove the bounds
\begin{equation}
 \begin{array}{rl}
  \left\| \Prever{\bg{\phi}}^{[\inverj,j]} \left( {\bf x} \right) \right\|_2^2
  &\!\!\!\!\in
  \left[
  \prever{\phi}_{\mindownarrow}^{[\inverj,j]2} = {\abssig}^{[\inverj,j]} \left( \frac{{\abssig}^{[\inverj,j]-1} \left( \prever{\phi}^{[\inverj,j]2} \right)}{\shada^2+1} \postver{\phi}_{\mindownarrow}^{[\inverj]2} \right),
  \prever{\phi}^{[\inverj,j]2}
  \right] \\

  \left\| \Prever{\bg{\Psi}}^{[\inverj,j]} \left( {\allofW} \right) \right\|_2^2
  &\!\!\!\!\leq
  \prever{\psi}^{[\inverj,j]2} = {\abssig}^{[\inverj,j]} \left( \frac{\shada^2+1}{{\abssig}^{[\inverj,j]-1} \left( \prever{\phi}^{[\inverj,j]2} \right)} \postver{\psi}^{[\inverj]2} \right) \\

  \left\| \Prever{\bg{\Psi}}^{[\inverj,j]} \left( {\allofW} \right) \right\|_{\Henum[\tau]}^2
  &\!\!\!\!\leq
  \undertilde{\prever{\psi}}^{[\inverj,j]2} = 
              {\!\begin{array}{c}
              \!\!\!\!\mathop{\sup}\limits_{{{\postver{\phi}_{\mindownarrow}^{[\inverj]} \leq \postver{\phi}^{[\inverj]} \leq \postver{\phi}^{[\inverj]}}\atop{-\undertilde{\postver{\psi}}^{[\inverj]} \leq \undertilde{\postver{\psi}}^{[\inverj]} \leq \undertilde{\postver{\psi}}^{[\inverj]} }}}
              \end{array}\!}
              \left\{
              \frac{
              \tau^{[\inverj,j]} \left( \postver{\phi}^{[\inverj]} \undertilde{\postver{\psi}}^{[\inverj]} \right)^2
              }{
              {\abssig}^{[\inverj,j]} \left( \frac{{\abssig}^{[\inverj,j]-1} \left( \prever{\phi}^{[\inverj,j]2} \right)}{\shada^2+1} \postver{\phi}^{[\inverj]2} \right)
              }
              \right\} \\
 \end{array}
\label{eq:preverboundwant}
\end{equation}
\begin{equation}
 \begin{array}{rl}
  \left\| \Postver{\bg{\phi}}^{[j]} \left( {\bf x} \right) \right\|_2^2
  &\!\!\!\!\in
  \left[ 
  \postver{\phi}_{\mindownarrow}^{[j]2} = \shada^2 + \frac{1}{\inver{p}^{[j]}} \mathop{\sum}\limits_{\inverj\in\Inver{\mathbb{P}}^{[j]}} \frac{\prever{\phi}_{\mindownarrow}^{[\inverj,j]2}}{\prever{\phi}^{[\inverj,j]2}},
  \postver{\phi}^{[j]2} = \shada^2+1
  \right] \\

  \left\| \Postver{\bg{\Psi}}^{[j]} \left( {\allofW} \right) \right\|_2^2
  &\!\!\!\!\leq
  \postver{\psi}^{[j]2}
  =
  \left( \shadb^{[j]} + \frac{\shadx^{[j]} \left| \tau^{[\inverj,j]} (0) \right|}{\shada} \right)^2
  +
  \inver{p}^{[j]} \shadx^{[j]2}
  \prever{\phi}^{[j]2} \prever{\psi}^{[j]2} \\

  \left\| \Postver{\bg{\Psi}}^{[j]} \left( {\allofW} \right) \right\|_{\Henum[\tau]}^2
  &\!\!\!\!\leq
  \undertilde{\postver{\psi}}^{[j]2}
  =
  \left( \shadb^{[j]} + \frac{\shadx^{[j]} \left| \tau^{[\inverj,j]} (0) \right|}{\shada} \right)^2
  +
  \inver{p}^{[j]} \shadx^{[j]2}
  \prever{\phi}^{[j]2} \undertilde{\prever{\psi}}^{[j]2} \\
 \end{array}
\label{eq:postverboundwant}
\end{equation}
with the base-cases 
$\| \Postver{\bg{\phi}}^{[-1]} ({\bf x}) \|_2^2 \in [ \postver{\phi}_{\mindownarrow}^{[-1]2} = 0, \postver{\phi}^{[-1]2} = 1 ]$, 
$\| \Postver{\bg{\Psi}}^{[-1]} (\allofW) \|_2^2 \leq \postver{\psi}^{[-1]2} = 1$ 
and 
$\| \Postver{\bg{\Psi}}^{[-1]} (\allofW) \|_{\Henum[\tau]}^2 \leq \undertilde{\postver{\psi}}^{[-1]2} = 1$.  
We proceed as follows:

{\bf Base case:} By the definition of the base case, using our assumptions, we 
have:
\[
  \begin{array}{rl}
   \left\| \Postver{\bg{\phi}}^{[-1]} \left( {\bf x} \right) \right\|_2^2
   &\!\!\!\!=
   \left\| {\bf x} \right\|_2^2 \leq 1 = \postver{\phi}^{[-1]2}
  \end{array}
\]
\[
  \begin{array}{rl}
   \left\| \Postver{\bg{\Psi}}^{[-1]} \left( \allofW \right) \right\|_2^2
   &\!\!\!\!=
   \left\| {\bf I}_n \right\|_2^2 \leq 1 = \postver{\psi}^{[-1]2}
  \end{array}
\]
\[
  \begin{array}{rl}
   \left\| \Postver{\bg{\Psi}}^{[-1]} \left( \allofW \right) \right\|_{\Henum[\tau]}^2
   &\!\!\!\!=
   \mathop{\rm sup}\limits_{{\bf x} \in \mathbb{X}} 
   \left\| \leftbf \frac{\Postver{\bg{\phi}}^{[-1]} \left( {\bf x} \right)}{\left\| \Postver{\bg{\phi}}^{[-1]} \left( {\bf x} \right) \right\|_2}, \Postver{\bg{\Psi}}^{[-1]} \left( \allofW \right) \rightbf_{\postver{\bf g}} \right\|_2^2
   =
   \mathop{\rm sup}\limits_{{\bf x} \in \mathbb{X}} \left\| \frac{\Postver{\bg{\phi}}^{[-1]} \left( {\bf x} \right)}{\left\| \Postver{\bg{\phi}}^{[-1]} \left( {\bf x} \right) \right\|_2} \right\|_2^2
   \leq 1 = \undertilde{\postver{\psi}}^{[-1]2}
  \end{array}
\]

{\bf Node case:} Assume (\ref{eq:preverboundwant}) is true.  Then, using our assumptions, 
we have that:
\[
  \begin{array}{rl}
   \left\| \Postver{\bg{\phi}}^{[j]} \left( {\bf x} \right) \right\|_2^2
   &\!\!\!\!=
   \left\|
             \left[ \begin{array}{c}
                    \shada \\
                    \frac{1}{\sqrt{\inver{p}^{[j]}}}
                    \left[ \frac{1}{\prever{\phi}^{[\inverj,j]}} {\Prever{\bg{\phi}}}^{[\inverj,j]} \left( {\bf x} \right) \right]_{\inverj \in \Inver{\mathbb{P}}^{[j]}} \\
             \end{array} \right]
   \right\|_2^2 \\

   &\!\!\!\!=
   \shada^2
   +
   \frac{1}{\sqrt{\inver{p}^{[j]}}}
   \mathop{\sum}\limits_{\inverj \in \Inver{\mathbb{P}}^{[j]}} 
   \frac{1}{\prever{\phi}^{[\inverj,j]}} \left\| {\Prever{\bg{\phi}}}^{[\inverj,j]} \left( {\bf x} \right) \right\|_2^2 \\

   &\!\!\!\!\in
   \left[ 
   \shada^2+\frac{1}{\inver{p}^{[j]}} \mathop{\sum}\limits_{\inverj\in\Inver{\mathbb{P}}^{[j]}} \frac{\prever{\phi}_{\mindownarrow}^{[\inverj,j]2}}{\prever{\phi}^{[\inverj,j]2}} = \postver{\phi}_{\mindownarrow}^{[j]2},
   \shada^2+1 = \postver{\phi}^{[j]2}
   \right] \\
  \end{array}
\]
\[
  \begin{array}{rl}
   \left\| \Postver{\bg{\Psi}}^{[j]} \left( \allofW \right) \right\|_2^2
   &\!\!\!\!=
   \left\|
             \left[ \begin{array}{c}
                    {\bf b}^{[j]\tsp} + {\bg{\upsilon}}_\tau^{[j]\tsp} \\
                    \sqrt{\inver{p}^{[j]}}
                    \mathop{\rm diag}\limits_{{{\inverj \in \Inver{\mathbb{P}}^{[j]}}}} \left( \prever{\phi}^{[\inverj,j]} \Prever{\bg{\Psi}}^{[\inverj,j]} \left( {\allofW} \right) \right) {\bf W}^{[j]} \\
             \end{array} \right]
   \right\|_2^2 \\

   &\!\!\!\!\leq
   \left( \left\| {\bf b}^{[j]} \right\|_2 + \left\| {\bg{\upsilon}}_\tau^{[j]} \right\|_2 \right)^2
   +
   \inver{p}^{[j]}
   \mathop{\max}\limits_{\inverj \in \Inver{\mathbb{P}}^{[j]}}
   \prever{\phi}^{[\inverj,j]} \left\| \Prever{\bg{\Psi}}^{[\inverj,j]} \left( {\allofW} \right) {\bf W}^{[\inverj,j]} \right\|_2^2 \\

   &\!\!\!\!\leq
   \left( \shadb^{[j]} + \frac{\shadx^{[j]} \left| \tau^{[\inverj,j]} (0) \right|}{\shada} \right)^2
   +
   \inver{p}^{[j]}
   \mathop{\max}\limits_{\inverj \in \Inver{\mathbb{P}}^{[j]}}
   \prever{\phi}^{[\inverj,j]} \Prever{{\psi}}^{[\inverj,j]2} \shadx^{[\inverj,j]2} \\

   &\!\!\!\!\leq
   \left( \shadb^{[j]} + \frac{\shadx^{[j]} \left| \tau^{[\inverj,j]} (0) \right|}{\shada} \right)^2
   +
   \inver{p}^{[j]}
   \prever{\phi}^{[j]} \Prever{{\psi}}^{[j]2} \shadx^{[j]2}
   = \postver{\psi}^{[j]2} \\
  \end{array}
\]
\[
  \begin{array}{rl}
   \left\| \Postver{\bg{\Psi}}^{[j]} \left( \allofW \right) \right\|_{\Henum[\tau]}^2
   &\!\!\!\!=
   \left\|
             \left[ \begin{array}{c}
                    {\bf b}^{[j]\tsp} + {\bg{\upsilon}}_\tau^{[j]\tsp} \\
                    \sqrt{\inver{p}^{[j]}}
                    \mathop{\rm diag}\limits_{{{\inverj \in \Inver{\mathbb{P}}^{[j]}}}} \left( \prever{\phi}^{[\inverj,j]} \Prever{\bg{\Psi}}^{[\inverj,j]} \left( {\allofW} \right) \right) {\bf W}^{[j]} \\
             \end{array} \right]
   \right\|_{\Henum[\tau]}^2 \\

   &\!\!\!\!\leq
   \left( \left\| {\bf b}^{[j]} \right\|_2 + \left\| {\bg{\upsilon}}_\tau^{[j]} \right\|_2 \right)^2
   +
   \inver{p}^{[j]}
   \mathop{\max}\limits_{\inverj \in \Inver{\mathbb{P}}^{[j]}}
   \prever{\phi}^{[\inverj,j]} \left\| \Prever{\bg{\Psi}}^{[\inverj,j]} \left( {\allofW} \right) {\bf W}^{[\inverj,j]} \right\|_{\Henum[\tau]}^2 \\

   &\!\!\!\!\leq
   \left( \shadb^{[j]} + \frac{\shadx^{[j]} \left| \tau^{[\inverj,j]} (0) \right|}{\shada} \right)^2
   +
   \inver{p}^{[j]}
   \mathop{\max}\limits_{\inverj \in \Inver{\mathbb{P}}^{[j]}}
   \prever{\phi}^{[\inverj,j]} \undertilde{\Prever{{\psi}}}^{[\inverj,j]2} \shadx^{[\inverj,j]2} \\

   &\!\!\!\!\leq
   \left( \shadb^{[j]} + \frac{\shadx^{[j]} \left| \tau^{[\inverj,j]} (0) \right|}{\shada} \right)^2
   +
   \inver{p}^{[j]}
   \prever{\phi}^{[j]} \undertilde{\Prever{{\psi}}}^{[j]2} \shadx^{[j]2}
   = \undertilde{\postver{\psi}}^{[j]2} \\
  \end{array}
\]

{\bf Edge case:} Assume (\ref{eq:postverboundwant}) is true.  Then, using 
our assumptions and the definition (\ref{eq:defmagfn}), and using that the 
magnitude function is increasing on $\mathbb{R}_+$, we have that:
\[
  \begin{array}{rl}
   \left\| \Prever{\bg{\phi}}^{[\inverj,j]} \left( {\bf x} \right) \right\|_2^2
   &\!\!\!\!=
   \left\|
             \left[
             a_{(0)k}^{[\inverj,j]\left<\frac{1}{2}\right>}\!
             \left[
             {\binom{k}{l}}^{\frac{1}{2}}\!
             \left(
             \sqrt{\frac{\abssig^{[\inverj,j]-1} \left( \prever{\phi}^{[\inverj,j]2} \right)}{\shada^2+1}}
             \Postver{\bg{\phi}}^{[\inverj]} \left( {\bf x} \right)
             \right)^{\otimes l}
             \right]_{1 \leq l \leq k}
             \right]_{k \geq 1}
   \right\|_2^2 \\

   &\!\!\!\!=
   \mathop{\sum}\limits_{k \geq 1}
             \left| a_{(0)k}^{[\inverj,j]} \right|
   \mathop{\sum}\limits_{1 \leq l \leq k}
             \binom{k}{l}
             \left(
             \frac{\abssig^{[\inverj,j]-1} \left( \prever{\phi}^{[\inverj,j]2} \right)}{\shada^2+1}
             \left\| \Postver{\bg{\phi}}^{[\inverj]} \left( {\bf x} \right) \right\|_2^2
             \right)^{l} \\

   &\!\!\!\!=
   \abssig^{[\inverj,j]}
             \left(
             \frac{\abssig^{[\inverj,j]-1} \left( \prever{\phi}^{[\inverj,j]2} \right)}{\shada^2+1}
             \left\| \Postver{\bg{\phi}}^{[\inverj]} \left( {\bf x} \right) \right\|_2^2
             \right) \\

   &\!\!\!\!\in
   \left[
   \abssig^{[\inverj,j]}
             \left(
             \frac{\abssig^{[\inverj,j]-1} \left( \prever{\phi}^{[\inverj,j]2} \right)}{\shada^2+1}
             \postver{\phi}_{\mindownarrow}^{[\inverj]2}
             \right),
   \abssig^{[\inverj,j]}
             \left(
             \frac{\abssig^{[\inverj,j]-1} \left( \prever{\phi}^{[\inverj,j]2} \right)}{\shada^2+1}
             \postver{\phi}^{[\inverj]2}
             \right)
   \right] \\

   &\!\!\!\!\in
   \left[
   \abssig^{[\inverj,j]}
             \left(
             \frac{\abssig^{[\inverj,j]-1} \left( \prever{\phi}^{[\inverj,j]2} \right)}{\shada^2+1}
             \postver{\phi}_{\mindownarrow}^{[\inverj]2}
             \right) = \prever{\phi}_{\mindownarrow}^{[\inverj,j]2},
   \prever{\phi}^{[\inverj,j]2}
   \right] \\
  \end{array}
\]
\[
  \begin{array}{rl}
   \left\| \Prever{\bg{\Psi}}^{[\inverj,j]} \left( \allofW \right) \right\|_2^2
   &\!\!\!\!=
   \mathop{\max}\limits_{i_{\inverj}}
   \left\|
             \left[
             \left| a_{(0)k}^{[\inverj,j]} \right|^{\frac{1}{2}}\!
             \left[
             {\binom{k}{l}}^{\frac{1}{2}}\!
             \left(
             \sqrt{\frac{\shada^2+1}{\abssig^{[\inverj,j]-1} \left( \prever{\phi}^{[\inverj,j]2} \right)}}
             \Postver{\bg{\Psi}}_{:i_{\inverj}}^{[\inverj]} \left( {\allofW} \right)
             \right)^{\otimes l}
             \right]_{1 \leq l \leq k}
             \right]_{k \geq 1}
   \right\|_2^2 \\

   &\!\!\!\!=
   \mathop{\max}\limits_{i_{\inverj}}
   \abssig^{[\inverj,j]}
             \left(
             \frac{\shada^2+1}{\abssig^{[\inverj,j]-1} \left( \prever{\phi}^{[\inverj,j]2} \right)}
             \left\| \Postver{\bg{\Psi}}_{:i_{\inverj}}^{[\inverj]} \left( {\allofW} \right) \right\|_2^2
             \right) \\

   &\!\!\!\!=
   \abssig^{[\inverj,j]}
             \left(
             \frac{\shada^2+1}{\abssig^{[\inverj,j]-1} \left( \prever{\phi}^{[\inverj,j]2} \right)}
             \left\| \Postver{\bg{\Psi}}^{[\inverj]} \left( {\allofW} \right) \right\|_2^2
             \right) \\

   &\!\!\!\!\leq
   \abssig^{[\inverj,j]}
             \left(
             \frac{\shada^2+1}{\abssig^{[\inverj,j]-1} \left( \prever{\phi}^{[\inverj,j]2} \right)}
             \Postver{{\psi}}^{[\inverj]2}
             \right) \\
  \end{array}
\]
\[
  \begin{array}{rl}
  \left\| \Prever{\bg{\Psi}}^{[\inverj,j]} \left( {\allofW} \right) \right\|_{\Henum[\tau]}^2
  &\!\!\!\!\leq
  \undertilde{\prever{\psi}}^{[\inverj,j]2} = 
              {\!\begin{array}{c}
              \!\!\!\!\mathop{\sup}\limits_{{{\postver{\phi}_{\mindownarrow}^{[\inverj]} \leq \postver{\phi}^{[\inverj]} \leq \postver{\phi}^{[\inverj]}}\atop{-\undertilde{\postver{\psi}}^{[\inverj]} \leq \undertilde{\postver{\psi}}^{[\inverj]} \leq \undertilde{\postver{\psi}}^{[\inverj]} }}}
              \end{array}\!}
              \left\{
              \frac{
              \tau^{[\inverj,j]} \left( \postver{\phi}^{[\inverj]} \undertilde{\postver{\psi}}^{[\inverj]} \right)^2
              }{
              {\abssig}^{[\inverj,j]} \left( \frac{{\abssig}^{[\inverj,j]-1} \left( \prever{\phi}^{[\inverj,j]2} \right)}{\shada^2+1} \postver{\phi}^{[\inverj]2} \right)
              }
              \right\} \\
  \end{array}
\]
where the final bound in this sequence is simply the definition of the norm in 
question with the range of the supremum expanded to the known bound on this 
range.

The desired result follows by identifying the output node $j=D-1$.  We note 
that the bounds $\prever{\phi}^{[\inverj,j]} \in \mathbb{R}_+$ may be chosen 
arbitrarily here.

\section{Proof of the Local Dual Model} \label{append:dualformlocal}

We now repeat the proof from the previous section \ref{append:dualform} for 
the local model.  Recall that the dual model has the form (Theorem 
\ref{th:netexpand}, equation (\ref{eq:ourdualmodel})):
\begin{equation}
 \begin{array}{rl}
  \changein {\bf f} \left( {\bf x}; \changein \allofW \right) 
  &\!\!\!\!= \leftbf \Postver{\bg{\Psi}}_{\changein} \left( \changein {\allofW} \right), \Postver{\bg{\phi}}_\changein \left( {\bf x} \right) \rightbf_{\Postver{\bf G}_\changein \left( {\bf x} \right)} \\
  &\!\!\!\!= \left[ \leftbf \Postver{\bg{\Psi}}_{\changein :i_{D-1}} \left( \changein {\allofW} \right), \Postver{\bg{\phi}}_\changein \left( {\bf x} \right) \rightbf_{\postver{\bf G}_{\changein :i_{D-1}} \left( {\bf x} \right)} \right]_{i_{D-1}} \\
 \end{array}
\label{eq:ourgoallocal}
\end{equation}
where, as per Figure \ref{fig:changein_feat}, 
$\Postver{\bg{\Psi}}_\changein = \Postver{\bg{\Psi}}_\changein^{[D-1]}$, 
$\Postver{\bg{\phi}}_\changein = \Postver{\bg{\phi}}_\changein^{[D-1]}$, 
$\Postver{\bf G}_\changein = \Postver{\bf G}_\changein^{[D-1]}$ and, 
given the base case 
$\Postver{\bg{\Psi}}_\changein^{[-1]} (\changein {\allofW}) = {\bf 0}_{0 \times n}$, 
$\Postver{\bg{\phi}}_\changein^{[-1]} ({\bf x}) = {\bf 0}_0$ 
$\Postver{\bf G}_\changein^{[-1]} ({\bf x}) = {\bf 1}_{0 \times n}$ 
the recursive definition of the feature maps is proposed as:
\begin{equation}
\begin{array}{rl}
  \Prever{\bg{\phi}}_{\changein}^{[\inverj,j]} \left( {\bf x} \right)
  &\!\!\!\!= \left[
             \frac{1}{\eta^{k}}
             \left[
             \left(
             \sqrt{\frac{\roctau_{(\tilde{\xpostm})\eta}^{[\inverj,j]2}}{\shada^2+1}}
             \Postver{\bg{\phi}}_{\changein}^{[\inverj]} \left( {\bf x} \right)
             \right)^{\otimes l}
             \right]_{1 \leq l \leq k}
             \right]_{k \geq 1} \\

  \Prever{\bg{\Psi}}_{\changein :i_{\inverj}}^{[\inverj,j]} \left( {\allofW} \right)
  &\!\!\!\!= \left[
             {\eta^{k}}
             \left[
             \left(
             \sqrt{\frac{\shada^2+1}{\roctau_{(\tilde{\xpostm})\eta}^{[\inverj,j]2}}}
             \Postver{\bg{\Psi}}_{\changein :i_{\inverj}}^{[\inverj]} \left( {\allofW} \right)
             \right)^{\otimes l}
             \right]_{1 \leq l \leq k}
             \right]_{k \geq 1} \\

  \Prever{\bf G}_{\changein :i_{\inverj}}^{[\inverj,j]} \left( {\bf x} \right)
  &\!\!\!\!= \left[
             a_{\big(\postver{x}_{i_{\inverj}}^{[\inverj]}\big)k}^{[\inverj,j]}
             \left[
             \binom{k}{l} \Henum_{k-l}
             \Postver{\bf G}_{\changein :i_{\inverj}}^{[\inverj]} \left( {\bf x} \right)^{\otimes l}
             \right]_{1 \leq l \leq k}
             \right]_{k \geq 1} \\
 \end{array}
\label{eq:edgecalclocal}
\end{equation}
$\forall \inverj \in \Inver{\mathbb{P}}^{[j]}$ 
(the feature map transforms associated with the edges of the graph) and:
\begin{equation}
 %
\label{eq:nodecalclocal}
\end{equation}
(the feature map transforms associated with the nodes of the graph).  Our 
approach to demonstrating that this is true is to prove that, given some input 
${\bf x} \in \mathbb{X}$ then, for all edges $(\inverj \to j)$:
\begin{equation}
  \begin{array}{rl}
   \changein \prever{\bf x}^{[\inverj,j]}
   &\!\!\!\!=
   \leftbf \Prever{\bg{\Psi}}_\changein^{[\inverj,j]} \left( \changein {\allofW} \right), \Prever{\bg{\phi}}_\changein^{[\inverj,j]} \left( {\bf x} \right) \rightbf_{\Prever{\bf G}_\changein^{[\inverj,j]} \left( {\bf x} \right)} \\
  \end{array}
\label{eq:edgefactlocal}
\end{equation}
and likewise for all nodes $j$:
\begin{equation}
  \begin{array}{rl}
   \changein \postver{\bf x}^{[j]}
   &\!\!\!\!=
   \leftbf \Postver{\bg{\Psi}}_\changein^{[j]} \left( \changein {\allofW} \right), \Postver{\bg{\phi}}_\changein^{[j]} \left( {\bf x} \right) \rightbf_{\Postver{\bf G}_\changein^{[j]} \left( {\bf x} \right)} \\
  \end{array}
\label{eq:nodefactlocal}
\end{equation}

{\bf Base case:} By the definition of the base case, we have:
\[
  \begin{array}{rl}
   \changein \postver{\bf x}^{[-1]}
   &\!\!\!\!=
   \leftbf \Postver{\bg{\Psi}}_\changein^{[-1]} \left( \changein {\allofW} \right), \Postver{\bg{\phi}}_\changein^{[-1]} \left( {\bf x} \right) \rightbf_{\Postver{\bf G}_\changein^{[-1]} \left( {\bf x} \right)}
   =
   {\bf 0}_n \\
  \end{array}
\]

{\bf Node case:} Assume (\ref{eq:edgefactlocal}) is true.  Then, using 
(\ref{eq:nodecalclocal}), we have that:
\[
  \begin{array}{l}
   \leftbf \Postver{\bg{\Psi}}_\changein^{[j]} \left( \changein {\allofW} \right), \Postver{\bg{\phi}}_\changein^{[j]} \left( {\bf x} \right) \rightbf_{\Postver{\bf G}_\changein^{[j]} \left( {\bf x} \right)}

   =
   \shada \changein {\bf b}^{[j]}
   +
   \mathop{\sum}\limits_{\inverj \in \Inver{\mathbb{P}}^{[j]}}
   \left(
   \changein {\bf W}^{[\inverj,j]\tsp} \prever{\bf x}^{[\inverj,j]}
   +
   {\bf W}^{[\inverj,j]\tsp} \changein \prever{\bf x}^{[\inverj,j]}
   +
   \changein {\bf W}^{[\inverj,j]\tsp} \changein \prever{\bf x}^{[\inverj,j]}
   \right) \\

   =
   \shada \left( {\bf b}^{[j]} + \changein {\bf b}^{[j]} \right)
   +
   \mathop{\sum}\limits_{\inverj \in \Inver{\mathbb{P}}^{[j]}}
   \left( {\bf W}^{[\inverj,j]} + \changein {\bf W}^{[\inverj,j]} \right)^\tsp \left( \prever{\bf x}^{[\inverj,j]} + \changein \prever{\bf x}^{[\inverj,j]} \right)
   -
   \shada {\bf b}^{[j]}
   -
   \mathop{\sum}\limits_{\inverj \in \Inver{\mathbb{P}}^{[j]}}
   {\bf W}^{[\inverj,j]\tsp} \prever{\bf x}^{[\inverj,j]} \\

   =
   \left( \postver{\bf x}^{[j]} + \changein \postver{\bf x}^{[j]} \right)
   -
   \postver{\bf x}^{[j]}
   =
   \changein \postver{\bf x}^{[j]} \\
  \end{array}
\]

{\bf Edge case:} Assume (\ref{eq:nodefactlocal}) is true.  Then, using 
(\ref{eq:edgecalclocal}) and (\ref{eq:hermexpand}), we have that:
\[
 \begin{array}{rl}
  \leftbf \Prever{\bg{\Psi}}_\changein^{[\inverj,j]} \left( \changein {\allofW} \right), \Prever{\bg{\phi}}_\changein^{[\inverj,j]} \left( {\bf x} \right) \rightbf_{\Prever{\bf G}_\changein^{[\inverj,j]} \left( {\bf x} \right)}
   &\!\!\!\!=
   \left[
   \mathop{\sum}\limits_{k \geq 1}
   a_{\big(\postver{x}_{i_{\inverj}}^{[\inverj]}\big)k}^{[\inverj,j]2}
   \mathop{\sum}\limits_{1 \leq l \leq k}
   \binom{k}{l}
   \Henum_{k-l}
   \leftbf \Postver{\bg{\Psi}}_{\changein :i_{\inverj}}^{[\inverj]} \left( \changein {\allofW} \right), \Postver{\bg{\phi}}_{\changein}^{[\inverj]} \left( {\bf x} \right) \rightbf_{\Prever{\bf G}_\changein^{[\inverj,j]} \left( {\bf x} \right)}^l
   \right]_{i_{\inverj}} \\

   &\!\!\!\!=
   \left[
   \mathop{\sum}\limits_{k \geq 1}
   a_{\big(\postver{x}_{i_{\inverj}}^{[\inverj]}\big)k}^{[\inverj,j]}
   \mathop{\sum}\limits_{1 \leq l \leq k}
   \binom{k}{l}
   \Henum_{k-l}
   \changein \postver{x}_{i_\inverj}^{[j]l}
   \right]_{i_{\inverj}} \\

   &\!\!\!\!=
   \left[
   \centau^{[\inverj,j]} \left( \changein \postver{x}_{i_{\inverj}}^{[j]} ; \postver{x}_{i_{\inverj}}^{[\inverj]} \right)
   \right]_{i_{\inverj}} \\

   &\!\!\!\!=
   \left[
   \tau^{[\inverj,j]} \left( \postver{x}_{i_{\inverj}}^{[\inverj]} + \changein \postver{x}_{i_{\inverj}}^{[j]} \right)
   -
   \tau^{[\inverj,j]} \left( \postver{x}_{i_{\inverj}}^{[\inverj]} \right)
   \right]_{i_{\inverj}} \\

   &\!\!\!\!=
   \changein \prever{\bf x}^{[\inverj,j]} \\
 \end{array}
\]

The desired result (\ref{eq:ourgoallocal}) then follows by identifying the 
output node $j=D-1$.

\subsection{Norm-Bounds for the Local Dual Model}

Our proof of the norm-bounds of the local model follows the same model as our 
proof of the validity of said model.  We want to prove the bounds
\begin{equation}
 \begin{array}{rl}
  \left\| \Prever{\bg{\phi}}_\changein^{[\inverj,j]} \left( {\bf x} \right) \odot \Prever{\bf G}_{\changein:i_\inverj}^{[\inverj,j]} \left( {\bf x} \right) \right\|_2^2
  &\!\!\!\!\leq
  \prever{\phi}_{\changein}^{[\inverj,j]2} = \abstaumax_{(\tilde{\xpostm}^{[\inverj,j]})\eta}^{[\inverj,j]2}
  \forall i_\inverj \\

  \left\| \Prever{\bg{\Psi}}_\changein^{[\inverj,j]} \left( \changein {\allofW} \right) \right\|_2^2
  &\!\!\!\!\leq
  \prever{\psi}_{\changein}^{[\inverj,j]2}
  =
  \absT_\eta \left( \frac{\shada^2+1}{\roctau_{(\tilde{\xpostm}^{[\inverj,j]})\eta}^{[\inverj,j]2}} \postver{\psi}_{\changein  }^{[\inverj]2} \right) \\
 \end{array}
\label{eq:preverboundwantlocal}
\end{equation}
\begin{equation}
 \begin{array}{rl}
  \left\| \Postver{\bg{\phi}}_{\changein}^{[j]} \left( {\bf x} \right) \odot \Postver{\bf G}_{\changein:i_j}^{[j]} \left( {\bf x} \right) \right\|_2^2
  &\!\!\!\!\leq
  \postver{\phi}_{\changein }^{[j]2}
  =
  \shada^2+1 \forall i_j \\

  \left\| \Postver{\bg{\Psi}}_{\changein}^{[j]} \left( {\allofW} \right) \right\|_2^2
  &\!\!\!\!\leq
  \postver{\psi}_{\changein}^{[j]2}
  =
   \shadb_\changein^{[j]2}
   + 
   3\inver{p}^{[j]}
   \left(
   \shadx_\changein^{[j]2}
   \tilde{\xpostm}^{[j]2}
   +
   2\shadx^{[j]2}
   \prever{\psi}_{\changein}^{[j]2}
   \right) \\
 \end{array}
\label{eq:postverboundwantlocal}
\end{equation}
with the base-cases 
$\| \Postver{\bg{\phi}}_\changein^{[-1]} ({\bf x}) \odot \Postver{\bf G}_{\changein:i_{-1}}^{[-1]} ({\bf x}) \|_2^2 \leq \postver{\phi}_\changein^{[-1]2} = 0$ 
and 
$\| \Postver{\bg{\Psi}}_\changein^{[-1]} (\changein \allofW) \|_2^2 \leq \postver{\psi}_\changein^{[-1]2} = 0$.  
We proceed as follows:

{\bf Base case:} By the definition of the base case, using our assumptions, we 
have:
\[
  \begin{array}{rl}
   \left\| \Postver{\bg{\phi}}_\changein^{[-1]} \left( {\bf x} \right) \odot \Postver{\bf G}_{\changein:i_{-1}}^{[-1]} \left( {\bf x} \right) \right\|_2^2
   &\!\!\!\!=
   0 = \postver{\phi}^{[-1]2} \forall i_{-1}
  \end{array}
\]
\[
  \begin{array}{rl}
   \left\| \Postver{\bg{\Psi}}_\changein^{[-1]} \left( \changein \allofW \right) \right\|_2^2
   &\!\!\!\!=
   0 = \postver{\psi}^{[-1]2}
  \end{array}
\]

{\bf Node case:} Assume (\ref{eq:preverboundwantlocal}) is true.  Then, using our assumptions, 
we have that:
\[
  \begin{array}{rl}
   \left\| \Postver{\bg{\phi}}_\changein^{[j]} \left( {\bf x} \right) \odot \Postver{\bf G}_{\changein:i_{j}}^{[j]} \left( {\bf x} \right) \right\|_2^2
   &\!\!\!\!=
   \shada^2
   +
   \frac{1}{3\inver{p}^{[j]}}
   \mathop{\sum}\limits_{\inverj \in \Inver{\mathbb{P}}^{[j]}}
   \frac{1}{\tilde{\xpostm}^{[\inverj,j]}}
   \left\| {\bf x}^{[\inverj,j]} \right\|_2^2
   +
   \frac{2}{3\inver{p}^{[j]}}
   \mathop{\sum}\limits_{\inverj \in \Inver{\mathbb{P}}^{[j]}}
   \frac{1}{\prever{\psi}_{\changein}^{[\inverj,j]}}
   \left\| \Prever{\bg{\phi}}_\changein^{[\inverj,j]} \left( {\bf x} \right) \odot \Prever{\bf G}_{\changein:i_{\inverj}}^{[\inverj,j]} \left( {\bf x} \right) \right\|_2^2 \\

   &\!\!\!\!\leq
   \shada^2+1
   =
   \postver{\phi}_{\changein }^{[j]2} \\
  \end{array}
\]
\[
  \begin{array}{rl}
   \left\| \Postver{\bg{\Psi}}_\changein^{[j]} \left( \changein \allofW \right) \right\|_2^2
   &\!\!\!\!=
   \left\| \changein {\bf b}^{[j]\tsp} \right\|_2^2
   +
   3\inver{p}^{[j]}
   \mathop{\max}\limits_{\inverj \in \Inver{\mathbb{P}}^{[j]}}
   \tilde{\xpostm}^{[\inverj,j]2}
   \left\| \changein {\bf W}^{[j]} \right\|_2^2
   +
   3\inver{p}^{[j]}
   \mathop{\max}\limits_{\inverj \in \Inver{\mathbb{P}}^{[j]}}
   \prever{\psi}_{\changein}^{[\inverj,j]2}
   \left\| {\bf W}^{[j]} \right\|_2^2
   +
   \ldots \\ & \hfill \ldots
   3\inver{p}^{[j]}
   \mathop{\max}\limits_{\inverj \in \Inver{\mathbb{P}}^{[j]}}
   \prever{\psi}_{\changein}^{[\inverj,j]2}
   \left\| \changein {\bf W}^{[j]} \right\|_2^2 \\

   &\!\!\!\!\leq
   \shadb_\changein^{[j]2}
   + 
   3\inver{p}^{[j]}
   \mathop{\max}\limits_{\inverj \in \Inver{\mathbb{P}}^{[j]}}
   \tilde{\xpostm}^{[\inverj,j]2}
   \shadx_\changein^{[j]2}
   +
   3\inver{p}^{[j]}
   \mathop{\max}\limits_{\inverj \in \Inver{\mathbb{P}}^{[j]}}
   \prever{\psi}_{\changein}^{[\inverj,j]2}
   \shadx^{[j]2}
   +
   3\inver{p}^{[j]}
   \mathop{\max}\limits_{\inverj \in \Inver{\mathbb{P}}^{[j]}}
   \prever{\psi}_{\changein}^{[\inverj,j]2}
   \shadx_\changein^{[j]2} \\

   &\!\!\!\!=
   \shadb_\changein^{[j]2}
   + 
   3\inver{p}^{[j]}
   \tilde{\xpostm}^{[j]2}
   \shadx_\changein^{[j]2}
   +
   6\inver{p}^{[j]}
   \prever{\psi}_{\changein}^{[j]2}
   \shadx^{[j]2} \\

   &\!\!\!\!=
   \shadb_\changein^{[j]2}
   + 
   3\inver{p}^{[j]}
   \left(
   \shadx_\changein^{[j]2}
   \tilde{\xpostm}^{[j]2}
   +
   2\shadx^{[j]2}
   \prever{\psi}_{\changein}^{[j]2}
   \right)
   =
   \postver{\psi}_{\changein}^{[j]2} \\
  \end{array}
\]

{\bf Edge case:} Assume (\ref{eq:postverboundwantlocal}) is true.  Then, using 
our assumptions and the definition (\ref{eq:rectactfn}) of the rectified 
activation function and it's increasing (on $\mathbb{R}_+$) envelope, we have 
that:
\[
  \begin{array}{rl}
   \left\| \Prever{\bg{\phi}}_{\changein}^{[\inverj,j]} \left( {\bf x} \right) \odot \Prever{\bf G}_{\changein:i_{\inverj}}^{[\inverj,j]} \left( {\bf x} \right) \right\|_2^2
   &\!\!\!\!=
   \left\|
             \left[
             \frac{a_{\big(\postver{x}_{i_{\inverj}}^{[\inverj]}\big)k}^{[\inverj,j]2}}{\eta^k}
             \left[
             \binom{k}{l} \Henum_{k-l}
   \left( \frac{\roctau_{(\tilde{\xpostm}^{[\inverj,j]})\eta}^{[\inverj,j]2}}{\shada^2+1} \right)^{\frac{l}{2}}
             \Postver{\bg{\phi}}_{\changein}^{[\inverj]} \left( {\bf x} \right)^{\otimes l}
             \odot
             \Postver{\bf G}_{\changein :i_\inverj}^{[\inverj]} \left( {\bf x} \right)^{\otimes l}
             \right]_{1 \leq l \leq k}
             \right]_{k \geq 1}
   \right\|_2^2 \\

   &\!\!\!\!=
   \left\|
             \left[
             \frac{a_{\big(\postver{x}_{i_{\inverj}}^{[\inverj]}\big)k}^{[\inverj,j]}}{\eta^k}
             \left[
             \binom{k}{l} \Henum_{k-l}
             \left(
   \left( \frac{\roctau_{(\tilde{\xpostm}^{[\inverj,j]})\eta}^{[\inverj,j]2}}{\shada^2+1} \right)^{\frac{l}{2}}
             \Postver{\bg{\phi}}_{\changein}^{[\inverj]} \left( {\bf x} \right)
             \odot
             \Postver{\bf G}_{\changein :i_\inverj}^{[\inverj]} \left( {\bf x} \right)
             \right)^{\otimes l}
             \right]_{1 \leq l \leq k}
             \right]_{k \geq 1}
   \right\|_2^2 \\

   &\!\!\!\!=
   \mathop{\sum}\limits_{k \geq 1}
   \frac{a_{\big(\postver{x}_{i_{\inverj}}^{[\inverj]}\big)k}^{[\inverj,j]2}}{\eta^{2k}}
   \mathop{\sum}\limits_{1 \leq l \leq k}
   \binom{k}{l}^2
   \Henum_{k-l}^2
   \left(
   \frac{\roctau_{(\tilde{\xpostm}^{[\inverj,j]})\eta}^{[\inverj,j]2}}{\shada^2+1}
   \left\|
             \Postver{\bg{\phi}}_{\changein}^{[\inverj]} \left( {\bf x} \right)
             \odot
             \Postver{\bf G}_{\changein :i_\inverj}^{[\inverj]} \left( {\bf x} \right)
   \right\|_2^2
  \right)^l \\
\end{array}
\]
\begin{equation}
  \begin{array}{rl}
   \left\| \Prever{\bg{\Psi}}_\changein^{[\inverj,j]} \left( \changein \allofW \right) \right\|_2^2
   &\!\!\!\!=
   \mathop{\max}\limits_{i_\inverj}
   \left\|
             \left[
             \eta^k
             \left[
   \left( \frac{\shada^2+1}{\roctau_{(\tilde{\xpostm}^{[\inverj,j]})\eta}^{[\inverj,j]2}} \right)^{\frac{l}{2}}
             \left(
             \Postver{\bg{\Psi}}_{\changein :i_{\inverj}}^{[\inverj]} \left( \changein {\allofW} \right)
             \right)^{\otimes l}
             \right]_{1 \leq l \leq k}
             \right]_{k \geq 1}
   \right\|_2^2 \\

   &\!\!\!\!=
   \mathop{\max}\limits_{i_\inverj}
   \absT_\eta \left(
             \frac{\shada^2+1}{\roctau_{(\tilde{\xpostm}^{[\inverj,j]})\eta}^{[\inverj,j]2}}
             \left\| \Postver{\bg{\Psi}}_{\changein :i_{\inverj}}^{[\inverj]} \left( \changein {\allofW} \right) \right\|_2^2
             \right) \\

   &\!\!\!\!=
   \absT_\eta \left(
             \frac{\shada^2+1}{\roctau_{(\tilde{\xpostm}^{[\inverj,j]})\eta}^{[\inverj,j]2}}
             \left\| \Postver{\bg{\Psi}}_\changein^{[\inverj]} \left( \changein {\allofW} \right) \right\|_2^2
             \right) \\

   &\!\!\!\!\leq
   \absT_\eta \left(
             \frac{\shada^2+1}{\roctau_{(\tilde{\xpostm}^{[\inverj,j]})\eta}^{[\inverj,j]2}}
             \Postver{{\psi}}_\changein^{[\inverj]2}
             \right)
   =
   \prever{\psi}_{\changein}^{[\inverj,j]2} \\
  \end{array}
\label{eq:absTconvscenario}
\end{equation}

The desired result follows by identifying the output node $j=D-1$.

\subsection{Convergence Region of Local Dual Model}

Here we prove the bounds on the change in weights and biases for which the 
local dual model holds as state in theorem \ref{th:netexpand_changein}.  The 
goal is to place bounds on $\shadx_\changein^{[j]}, \shadb_\changein^{[j]}$ 
to ensure that:
\[
 \begin{array}{l}
  \frac{\shada^2+1}{\roctau_{(\tilde{\xpostm}^{[\inverj,j]})\eta}^{[\inverj,j]2}}
  \left\| \Postver{\bg{\Psi}}_\changein^{[\inverj]} \left( \changein {\allofW} \right) \right\|_2^2
  \leq
  \frac{\shada^2+1}{\roctau_{(\tilde{\xpostm}^{[\inverj,j]})\eta}^{[\inverj,j]2}}
  \Postver{{\psi}}_\changein^{[\inverj]2}
  \leq \rocT_\eta^2
 \end{array}
\]
$\forall \inverj \in \Inver{\mathbb{P}}^{[j]}$, $\forall j$ in 
(\ref{eq:absTconvscenario}).  This result suffices to ensure that 
$\| \Prever{\bg{\Psi}}_\changein^{[\inverj,j]} (\changein \allofW) 
\|_{\prever{\bf h}^{[\inverj,j]*}}^2 \leq \absTmax_\eta^2$, and subsequently 
that both feature maps are convergent, allowing us to conclude that the 
model is well-defined (convergent) and enabling e.g. our bound on Rademacher 
complexity to be derived.

We aim to find the largest possible $\shadx_\changein^{[j]}, \shadb_\changein^{[j]}$ 
such that:
\[
 \begin{array}{l}
  \frac{\shada^2+1}{\roctau_{(\tilde{\xpostm}^{[\inverj,j]})\eta}^{[\inverj,j]2}}
  \Postver{{\psi}}_\changein^{[\inverj]2}
  \leq
  \rocT_\eta^2
 \end{array}
\]
$\forall j:\inverj \in \Inver{\mathbb{P}}^{[j]}$. So we require that:
\begin{equation}
 \begin{array}{l}
  \Postver{{\psi}}_\changein^{[\inverj]2}
  \leq
  \mathop{\min}\limits_{j:\inverj \in \Inver{\mathbb{P}}^{[j]}}
  \frac{\roctau_{(\tilde{\xpostm}^{[\inverj,j]})\eta}^{[\inverj,j]2}}{\shada^2+1}
  \rocT_\eta^2
 \end{array}
\label{eq:psilimitbase}
\end{equation}
Recall that:
\[
  \begin{array}{rl}
   \postver{\psi}_{\changein}^{[\inverj]2}
   =
   \shadb_\changein^{[\inverj]2}
   + 
   3\inver{p}^{[\inverj]}
   \left(
   \shadx_\changein^{[\inverj]2}
   \tilde{\xpostm}^{[\inverj]2}
   +
   2\shadx^{[\inverj]2}
   \prever{\psi}_{\changein}^{[\inverj]2}
   \right)
  \end{array}
\]
Thus our condition becomes:
\[
  \begin{array}{rl}
   \left(
   \shadb_\changein^{[\inverj]2}
   + 
   3\inver{p}^{[\inverj]}
   \tilde{\xpostm}^{[\inverj]2}
   \shadx_\changein^{[\inverj]2}
   \right)
   +
   6\inver{p}^{[\inverj]}
   \shadx^{[\inverj]2}
   \prever{\psi}_{\changein}^{[\inverj]2}
   \leq
   \Postver{{\psi}}_\changein^{[\inverj]2}
  \end{array}
\]

Selecting $d^{[\inverj]} \in (0,1)$ $\forall j$ we can simplify the requirement 
to:
\[
  \begin{array}{rl}
   \left(
   \shadb_\changein^{[\inverj]2}
   + 
   3\inver{p}^{[\inverj]}
   \tilde{\xpostm}^{[\inverj]2}
   \shadx_\changein^{[\inverj]2}
   \right)
   \leq
   d^{[\inverj]}
   \Postver{{\psi}}_\changein^{[\inverj]2}
  \end{array}
\]
and:
\[
  \begin{array}{rl}
   6\inver{p}^{[\inverj]}
   \shadx^{[\inverj]2}
   \prever{\psi}_{\changein}^{[\inverj]2}
   \leq
   \left( 1 - d^{[\inverj]} \right)
   \mathop{\min}\limits_{j:\inverj \in \Inver{\mathbb{P}}^{[j]}}
   \frac{\roctau_{(\tilde{\xpostm}^{[\inverj,j]})\eta}^{[\inverj,j]2}}{\shada^2+1}
   \rocT_\eta^2
  \end{array}
\]
Re-ordering, we find:
\[
  \begin{array}{rl}
   6\inver{p}^{[\inverj]}
   \shadx^{[\inverj]2}
   \absT_\eta \left(
             \frac{\shada^2+1}{\roctau_{(\tilde{\xpostm}^{[\inverj',\inverj]})\eta}^{[\inverj',\inverj]2}}
             \Postver{{\psi}}_\changein^{[\inverj']2}
             \right)
   \leq
   \left( 1 - d^{[\inverj]} \right)
   \Postver{{\psi}}_\changein^{[\inverj]2}
   \;\;
   \forall \inverj' \in \Inver{\mathbb{P}}^{[\inverj]}
  \end{array}
\]
and after cleaning up and re-indexing:
\[
  \begin{array}{rl}
   \Postver{{\psi}}_\changein^{[j]2}
   \leq
   \frac{\roctau_{(\tilde{\xpostm}^{[j,\inverj]})\eta}^{[j,\inverj]2}}{\shada^2+1}
   \absT_\eta^{-1}
   \left(
   \frac{1}{6\inver{p}^{[\inverj]} \shadx^{[\inverj]2}}
   \left( 1 - d^{[\inverj]} \right)
   \Postver{{\psi}}_\changein^{[\inverj]2}
   \right)
   \;\;
   \forall \inverj : j \in \Inver{\mathbb{P}}^{[\inverj]}
  \end{array}
\]
so, overall:
\[
 \begin{array}{l}
  \Postver{{\psi}}_\changein^{[j]2}
  \leq
  \mathop{\min}\limits_{\inverj:j \in \Inver{\mathbb{P}}^{[\inverj]}}
  \frac{\roctau_{(\tilde{\xpostm}^{[j,\inverj]})\eta}^{[j,\inverj]2}}{\shada^2+1}
  \left\{
  \rocT_\eta^2,
   \absT_\eta^{-1}
   \left(
   \frac{1}{6\inver{p}^{[\inverj]} \shadx^{[\inverj]2}}
   \left( 1 - d^{[\inverj]} \right)
   \Postver{{\psi}}_\changein^{[\inverj]2}
   \right)
  \right\}
 \end{array}
\]
and so sufficient conditions are:
\[
  \begin{array}{rl}
   \left(
   \shadb_\changein^{[j]2}
   + 
   3\inver{p}^{[j]}
   \tilde{\xpostm}^{[j]2}
   \shadx_\changein^{[j]2}
   \right)
   \leq
   d^{[j]}
   u^{[j]2}
  \end{array}
\]
where:
\[
 \begin{array}{l}
  u^{[j]2}
  =
  \mathop{\min}\limits_{\inverj:j \in \Inver{\mathbb{P}}^{[\inverj]}}
  \frac{\roctau_{(\tilde{\xpostm}^{[j,\inverj]})\eta}^{[j,\inverj]2}}{\shada^2+1}
  \left\{
  \rocT_\eta^2,
   \absT_\eta^{-1}
   \left(
   \frac{1}{6\inver{p}^{[\inverj]} \shadx^{[\inverj]2}}
   \left( 1 - d^{[\inverj]} \right)
   \Postver{{\psi}}_\changein^{[\inverj]2}
   \right)
  \right\}
 \end{array}
\]
and the final result follows by setting $d^{[j]} = \frac{1}{2}$ $\forall j$.

\subsection{RKHS Form of Local Model} \label{sup:itshilbert}

Our goal in this section is to derive the LiNK kernel from the Local dual 
model.  For reason that will become apparent we find it convenient to work 
with a slight variant of the local dual feature map with minor re-scaling, 
namely:
\begin{equation}
 \begin{array}{rl}
  \Prever{\bg{\phi}}_{\changein}^{[\inverj,j]} \left( {\bf x} \right)
  &\!\!\!\!= \left[
             \left[
             \left(
             \sqrt{\frac{\roctau_{(\tilde{\xpostm})\eta}^{[\inverj,j]2}}{\shada^2+1}}
             \Postver{\bg{\phi}}_{\changein}^{[\inverj]} \left( {\bf x} \right)
             \right)^{\otimes l}
             \right]_{1 \leq l \leq k}
             \right]_{k \geq 1} \\

  \Prever{\bg{\Psi}}_{\changein :i_{\inverj}}^{[\inverj,j]} \left( {\allofW} \right)
  &\!\!\!\!= \left[
             \left[
             \left(
             \sqrt{\frac{\shada^2+1}{\roctau_{(\tilde{\xpostm})\eta}^{[\inverj,j]2}}}
             \Postver{\bg{\Psi}}_{\changein :i_{\inverj}}^{[\inverj]} \left( {\allofW} \right)
             \right)^{\otimes l}
             \right]_{1 \leq l \leq k}
             \right]_{k \geq 1} \\

  \Prever{\bf G}_{\changein :i_{\inverj}}^{[\inverj,j]} \left( {\bf x} \right)
  &\!\!\!\!= \left[
             a_{\big(\postver{x}_{i_{\inverj}}^{[\inverj]}\big)k}^{[\inverj,j]}
             \left[
             \binom{k}{l} \Henum_{k-l}
             \Postver{\bf G}_{\changein :i_{\inverj}}^{[\inverj]} \left( {\bf x} \right)^{\otimes l}
             \right]_{1 \leq l \leq k}
             \right]_{k \geq 1} \\
 \end{array}%
\end{equation}
We have already demonstrated that the necessary conditions for functions in 
our space to like in an RKHS, so it remains derive the kernel representing 
this space.  To this end we recall that Reisz representor theory implies that 
$\forall {\bf x} \in \mathbb{X}$ $\exists$ unique ${\bf K}_{\bf x} \in 
\mathcal{F} \times \mathbb{R}^m$ such that $\left< {\bf f} ({\bf x}), {\bf v} 
\right> = \left< {\bf f}, {\bf K}_{\bf x} {\bf v} \right>_{\rkhs{}}$ 
$\forall {\bf v} \in \mathbb{R}^m$; from which the kernel may be obtained 
using:
\[
 \begin{array}{l}
  {\bf K} \left( {\bf x},{\bf x}' \right)
  =
  \Big[ \left< {\bf K}_{\bf x} {\bg{\delta}}{}_{(i_{D-1})}^{[D-1]}, {\bf K}_{{\bf x}'} {\bg{\delta}}{}_{(i'_{D-1})}^{[D-1]} \right>_{\rkhs{}} \Big]{}_{{{}\atop{\!i_{D-1},i'_{D-1}}}}
 \end{array}
\]
where ${\bg{\delta}}_{(k)}^{[j]} = [\delta_{k,i_j}]_{i_j}$.  Thus our first 
task is to find such a ${\bf K}_{\bf x}$.  Recall:
\[
 \begin{array}{rl}
  \changein {\bf f} \left( {\bf x} \right)
  =
  \left(
  \Postver{\bg{\Psi}}_\changein^{[D-1]} \left( {\allofW} \right)
  \odot
  \Postver{\bf G}_{\changein}^{[D-1]} \left( {\bf x} \right)
  \right)^\tsp
  \Postver{\bg{\phi}}_\changein^{[D-1]} \left( {\bf x} \right)
 \end{array}
\]
and so:
\[
 \begin{array}{rl}
  \left< \changein {\bf f} \left( {\bf x} \right), {\bf v} \right>
  &\!\!\!\!=
  \changein {\bf f} \left( {\bf x} \right)^\tsp {\bf v} \\

  &\!\!\!\!=
  \Postver{\bg{\phi}}_\changein^{[D-1]} \left( {\bf x} \right)^\tsp
  \left(
  \Postver{\bg{\Psi}}_\changein^{[D-1]} \left( {\allofW} \right)
  \odot
  \Postver{\bf G}_{\changein}^{[D-1]} \left( {\bf x} \right)
  \right)
  {\bf v} \\

  &\!\!\!\!=
  \sum_{k,i'_{D-1}}
  \Postver{{\phi}}_{\changein k}^{[D-1]} \left( {\bf x} \right)
  \left(
  \Postver{{\Psi}}_{\changein k,i'_{D-1}}^{[D-1]} \left( {\allofW} \right)
  \Postver{G}_{\changein k,i'_{D-1}}^{[D-1]} \left( {\bf x} \right)
  \right)
  v_{i'_{D-1}} \\
  &\!\!\!\!=
  \sum_{k,i'_{D-1}}
  \Postver{{\Psi}}_{\changein k,i'_{D-1}}^{[D-1]} \left( {\allofW} \right)
  \left(
  \Postver{{\phi}}_{\changein k}^{[D-1]} \left( {\bf x} \right)
  \Postver{G}_{\changein k,i'_{D-1}}^{[D-1]} \left( {\bf x} \right)
  v_{i'_{D-1}}
  \right) \\
  &\!\!\!\!=
  \sum_{k,i'_{D-1}}
  \Postver{{\Psi}}_{\changein k,i'_{D-1}}^{[D-1]} \left( {\allofW} \right)
  \left(
  \sum_{i''_{D-1}}
  \delta_{i'_{D-1},i''_{D-1}}
  \Postver{{\phi}}_{\changein k}^{[D-1]} \left( {\bf x} \right)
  \Postver{G}_{\changein k,i''_{D-1}}^{[D-1]} \left( {\bf x} \right)
  v_{i''_{D-1}}
  \right) \\
  &\!\!\!\!=
  \left< {\bf f}, {\bf K}_{\bf x} {\bf v} \right>
 \end{array}
\]
where we have denoted:
\[
 \begin{array}{rl}
  {\bf f}
  &\!\!\!\!=
  \left[
  \Postver{{\Psi}}_{\changein k,i_{D-1}}^{[D-1]} \left( {\allofW} \right)
  \right]_{(k,i_{D-1})} \\

  {\bf K}_{\bf x}
  &\!\!\!\!=
  \left[
  \delta_{i_{D-1},i'_{D-1}}
  \Postver{{\phi}}_{\changein k}^{[D-1]} \left( {\bf x} \right)
  \Postver{G}_{\changein k,i'_{D-1}}^{[D-1]} \left( {\bf x} \right)
  \right]_{(k,i_{D-1}),i'_{D-1}} \\
 \end{array}
\]
We may therefore proceed to derive the kernel as follows, letting 
$\left< \cdot, \cdot \right>_{\rkhs{}}$ be the standard inner product:
\[
 \begin{array}{rl}
  K_{i_{D-1},i'_{D-1}} \left( {\bf x}, {\bf x}' \right)
  &\!\!\!\!=
  \Bigg[ \Bigg<
  \left[
  \delta_{\tilde{i}'_{D-1},\tilde{i}''_{D-1}}
  \Postver{{\phi}}_{\changein k}^{[D-1]} \left( {\bf x} \right)
  \Postver{G}_{\changein k,\tilde{i}''_{D-1}}^{[D-1]} \left( {\bf x} \right)
  \right]_{(k,\tilde{i}'_{D-1}),\tilde{i}''_{D-1}}
  \left[
  \delta_{i_{D-1},\tilde{i}''_{D-1}}
  \right]_{\tilde{i}''_{D-1}}, \\
  &
  \left[
  \delta_{\tilde{i}'_{D-1},\tilde{i}''_{D-1}}
  \Postver{{\phi}}_{\changein k}^{[D-1]} \left( {\bf x}' \right)
  \Postver{G}_{\changein k,\tilde{i}''_{D-1}}^{[D-1]} \left( {\bf x}' \right)
  \right]_{(k,\tilde{i}'_{D-1}),\tilde{i}''_{D-1}}
  \left[
  \delta_{i'_{D-1},\tilde{i}''_{D-1}}
  \right]_{\tilde{i}''_{D-1}}
  \Bigg>_{\rkhs{}} \Bigg]_{i_{D-1},i'_{D-1}} \\

  &\!\!\!\!=
  \Bigg[ \Bigg<
  \left[
  \delta_{\tilde{i}'_{D-1},i_{D-1}}
  \Postver{{\phi}}_{\changein k}^{[D-1]} \left( {\bf x} \right)
  \Postver{G}_{\changein k,i_{D-1}}^{[D-1]} \left( {\bf x} \right)
  \right]_{(k,\tilde{i}'_{D-1})}, \\
  &
  \left[
  \delta_{\tilde{i}'_{D-1},i'_{D-1}}
  \Postver{{\phi}}_{\changein k}^{[D-1]} \left( {\bf x}' \right)
  \Postver{G}_{\changein k,i'_{D-1}}^{[D-1]} \left( {\bf x}' \right)
  \right]_{(k,\tilde{i}'_{D-1})}
  \Bigg>_{\rkhs{}} \Bigg]_{i_{D-1},i'_{D-1}} \\

  &\!\!\!\!=
  \left[ \mathop{\sum}\limits_{k,\tilde{i}'_{D-1}}
  \delta_{\tilde{i}'_{D-1},i_{D-1}}
  \Postver{{\phi}}_{\changein k}^{[D-1]} \left( {\bf x} \right)
  \Postver{G}_{\changein k,i_{D-1}}^{[D-1]} \left( {\bf x} \right) 
  \delta_{\tilde{i}'_{D-1},i'_{D-1}}
  \Postver{{\phi}}_{\changein k}^{[D-1]} \left( {\bf x}' \right)
  \Postver{G}_{\changein k,i'_{D-1}}^{[D-1]} \left( {\bf x}' \right)
  \right]_{i_{D-1},i'_{D-1}} \\
\end{array}
\]
where:
\[
 \begin{array}{rl}
  \Postver{K}_{i_j}^{[j]} \left( {\bf x}, {\bf x}' \right)
  &\!\!\!\!=
  \left<
  \Postver{\bg{\phi}}_{\changein}^{[j]} \left( {\bf x} \right)
  \odot
  \Postver{\bf G}_{\changein :i_{j}}^{[j]} \left( {\bf x} \right),
  \Postver{\bg{\phi}}_{\changein}^{[j]} \left( {\bf x}' \right)
  \odot
  \Postver{\bf G}_{\changein :i_{j}}^{[j]} \left( {\bf x}' \right)
  \right> \\

  &\!\!\!\!=
  \shada^2
  +
  \frac{1}{\inver{p}^{[j]}}
  \mathop{\sum}\limits_{\inverj \in \Inver{\mathbb{P}}^{[j]}}
  \frac{1}{\tilde{\xpostm}^{[\inverj,j]2}}
  \left< \prever{\bf x}^{[\inverj,j]}, \prever{\bf x}^{\prime[\inverj,j]} \right>
  +
  \ldots \\ &\;\;\;\;\ldots +
  \frac{1}{\inver{p}^{[j]}}
  \mathop{\sum}\limits_{\inverj \in \Inver{\mathbb{P}}^{[j]}}
  \frac{1}{H^{[\inverj]}}
  \frac{1}{\prever{\psi}_{\changein}^{[\inverj,j]}}
  \mathop{\sum}\limits_{i_\inverj}
  \left<
  \Prever{\bg{\phi}}_{\changein}^{[\inverj,j]} \left( {\bf x} \right)
  \odot
  \Prever{\bf G}_{\changein :i_{\inverj}}^{[\inverj,j]} \left( {\bf x} \right),
  \Prever{\bg{\phi}}_{\changein}^{[\inverj,j]} \left( {\bf x}' \right)
  \odot
  \Prever{\bf G}_{\changein :i_{\inverj}}^{[\inverj,j]} \left( {\bf x}' \right)
  \right> \\

  &\!\!\!\!=
  \shada^2
  +
  \frac{1}{\inver{p}^{[j]}}
  \mathop{\sum}\limits_{\inverj \in \Inver{\mathbb{P}}^{[j]}}
  \frac{1}{\tilde{\xpostm}^{[\inverj,j]2}}
  \left< \prever{\bf x}^{[\inverj,j]}, \prever{\bf x}^{\prime[\inverj,j]} \right>
  +
  \ldots \\
\end{array}
\]
So, noting that this is independent of $i_j$, we find that ${\bf f} \in 
\rkhs{\bf K}$, where ${\bf K} = {\bf I} K_{\rm LiNK}$, where 
$K_{\rm LiNK} = K_{\rm LiNK}^{[D-1]}$ is defined recursively:
\[
 \begin{array}{rl}
  \Postver{K}_{\rm LiNK}^{[j]} \left( {\bf x}, {\bf x}' \right)
  &\!\!\!\!=
  \shada^2
  +
  \frac{1}{\inver{p}^{[j]}}
  \mathop{\sum}\limits_{\inverj \in \Inver{\mathbb{P}}^{[j]}}
  \frac{1}{\tilde{\xpostm}^{[\inverj,j]2}}
  \Sigma^{[\inverj,j]} \left( {\bf x}, {\bf x}' \right)
  +
  \frac{1}{\inver{p}^{[j]}}
  \mathop{\sum}\limits_{\inverj \in \Inver{\mathbb{P}}^{[j]}}
  \frac{1}{\prever{\psi}_{\changein}^{[\inverj,j]}}
  \frac{1}{H^{[\inverj]}}
  \mathop{\sum}\limits_{i_\inverj}
   \abstau_\eta^{[\inverj,j]}
   \left(
   \frac{\roctau_{(\tilde{\xpostm})\eta}^{[\inverj,j]2}}{\shada^2+1}
   \Postver{K}_{\rm LiNK}^{[\inverj]} \left( {\bf x}, {\bf x}' \right)
   ; \postver{x}_{i_{\inverj}}^{[\inverj]}, \postver{x}_{i_{\inverj}}^{\prime[\inverj]}
   \right) \\
 \end{array}
\]
with $\Postver{K}_{\rm LiNK}^{[-1]} ({\bf x},{\bf x}') = 0$.

Recall that:
\[
 \begin{array}{rl}
  \abstau_\eta^{[\inverj,j]} \left( \zeta ; \xi, \xi' \right)
  &\!\!\!\!=
  \mathop{\sum}\limits_{k=1}^\infty
  \frac{a_{(\xi)k}^{[\inverj,j]} a_{(\xi')k}^{[\inverj,j]}}{\eta^{2k}}
  \mathop{\sum}\limits_{l=1}^k {\binom{k}{l}}^2 \Henum_{k-l}^2 \zeta^l \\
 \end{array}
\]
We aim to express:
\[
 \begin{array}{rl}
   \abstau_\eta^{[\inverj,j]} \left( \zeta ; \xi, \xi' \right)
  =
  \mathop{\sum}\limits_q
   \abstau_{\eta,q}^{[\inverj,j]} \left( \zeta ; \xi, \xi' \right)
 \end{array}
\]
where $\abstau_{\eta,q}^{[\inverj,j]} (\zeta;\xi,\xi') = b(\xi,\xi')\zeta^q$.  As 
noted previously, if:
\[
 \begin{array}{rll}
  f \left( x \right)
  &\!\!\!\!=
  \mathop{\sum}\limits_{k=0}^\infty
  a_{k}
  \mathop{\sum}\limits_{l=0}^{k}
  \binom{k}{l}
  \Henum_{k-l}
  x^{l} \\
 \end{array}
\]
then:
\[
 \begin{array}{rll}
  f^{(q)} \left( x \right)
  &\!\!\!\!=
  \mathop{\sum}\limits_{k=0}^\infty
  \frac{(k+q)!}{k!}
  a_{k+q}
  \mathop{\sum}\limits_{l=0}^{k-q}
  \binom{k}{l}
  \Henum_{k-l}
  x^{l} \\
 \end{array}
\]

Working toward the general case:
\[
 \begin{array}{rl}
  \abstau_{\eta,1}^{[\inverj,j]} \left( \zeta ; \xi, \xi' \right)
  &\!\!\!\!=
  \zeta
  \mathop{\sum}\limits_{k=1}^\infty
  \frac{a_{(\xi)k}^{[\inverj,j]} a_{(\xi')k}^{[\inverj,j]}}{\eta^{2k}}
  {\binom{k}{1}}^2 \Henum_{k-1}^2 \\
  &\!\!\!\!=
  \zeta
  \mathop{\sum}\limits_{k,k'=0}^\infty
  \delta_{k,k'}
  \frac{a_{(\xi)k+1}^{[\inverj,j]}}{\eta^{k+1}}
  {\binom{k+1}{1}} 
  \Henum_{k}
  \frac{a_{(\xi')k'+1}^{[\inverj,j]}}{\eta^{k'+1}}
  {\binom{k'+1}{1}} 
  \Henum_{k'} \\
 \end{array}
\]
and:
\[
 \begin{array}{rl}
  \abstau_{\eta,2}^{[\inverj,j]} \left( \zeta ; \xi, \xi' \right)
  &\!\!\!\!=
  \zeta^2
  \mathop{\sum}\limits_{k=2}^\infty
  \frac{a_{(\xi)k}^{[\inverj,j]} a_{(\xi')k}^{[\inverj,j]}}{\eta^{2k}}
  {\binom{k}{2}}^2 \Henum_{k-2}^2 \\
  &\!\!\!\!=
  \zeta^2
  \mathop{\sum}\limits_{k,k'=0}^\infty
  \delta_{k,k'}
  \frac{a_{(\xi)k+2}^{[\inverj,j]}}{\eta^{k+2}}
  {\binom{k+2}{2}}
  \Henum_{k}
  \frac{a_{(\xi')k'+2}^{[\inverj,j]}}{\eta^{k'+2}}
  {\binom{k'+2}{2}}
  \Henum_{k'} \\
 \end{array}
\]
and so on, so:
\[
 %
\]
where $\cdot$ is the Cauchy product and in the final step we have used 
Mertens' theorem for Cauchy products, and:
\[
 \begin{array}{rll}
  \centau^{[\inverj,j]}_\eta \left( \zeta; \xi \right)
  &\!\!\!\!=
  \mathop{\sum}\limits_{k=1}^\infty
  \frac{a^{[\inverj,j]}_{(\xi)k}}{\eta^k}
  \mathop{\sum}\limits_{l=1}^{k}
  \binom{k}{l}
  \Henum_{k-l}
  \zeta^{l} \\
 \end{array}
\]
Finally, we see that, in the limit $\eta \to 1$:
\[
 \begin{array}{rl}
  \abstau_{\eta,q}^{[\inverj,j]} \left( \zeta ; \xi, \xi' \right)
  &\!\!\!\!=
  \frac{1}{q!}
  \centau^{[\inverj,j](q)} \left( 0; \xi \right)
  \frac{1}{q!}
  \centau^{[\inverj,j](q)} \left( 0; \xi' \right)
  \zeta^q \\
 \end{array}
\]
which leads to the simplified form of the LiNK:
\[
 \begin{array}{rl}
  \Postver{K}_{\rm LiNK}^{[j]} \left( {\bf x}, {\bf x}' \right)
  &\!\!\!\!=
  \shada^2
  +
  \frac{1}{\inver{p}^{[j]}}
  \mathop{\sum}\limits_{\inverj \in \Inver{\mathbb{P}}^{[j]}}
  \frac{1}{\tilde{\xpostm}^{[\inverj,j]2}}
  \Sigma^{[\inverj,j]} \left( {\bf x}, {\bf x}' \right)
  +
  \ldots \\ & \ldots +
  \frac{1}{\inver{p}^{[j]}}
  \mathop{\sum}\limits_{\inverj \in \Inver{\mathbb{P}}^{[j]}}
  \frac{1}{\prever{\psi}_{\changein}^{[\inverj,j]}}
  \frac{1}{H^{[\inverj]}}
  \mathop{\sum}\limits_{q=1}^\infty
  \mathop{\sum}\limits_{i_\inverj}
   \frac{1}{q!}
   \centau^{[\inverj,j](q)} \left( 0; \postver{x}_{i_{\inverj}}^{[\inverj]} \right)
   \frac{1}{q!}
   \centau^{[\inverj,j](q)} \left( 0; \postver{x}_{i_{\inverj}}^{\prime[\inverj]} \right)
   \left(
   \frac{\roctau_{(\tilde{\xpostm})\eta}^{[\inverj,j]2}}{\shada^2+1}
   \Postver{K}_{\rm LiNK}^{[\inverj]} \left( {\bf x}, {\bf x}' \right)
   \right)^q \\
 \end{array}
\]
with $\Postver{K}_{\rm LiNK}^{[-1]} ({\bf x},{\bf x}') = 0$.  With some cleanup:
\[
 %
\]
with $\Postver{K}_{\rm LiNK}^{[-1]} ({\bf x},{\bf x}') = 0$.

\section{Rademacher Complexity Bounds - Proof of Theorems \ref{th:radbounded}, \ref{th:radbounded_lip} and \ref{th:radbounded_local}} \label{supp:radbound}

In this supplementary we prove theorems relating to the Rademacher complexity 
of neural networks for the global and local models.

{\bf Theorem \ref{th:radbounded}}
 {\em The set $\mathcal{F} = \{ f (\cdot; \allofW) : \mathbb{R}^n\to\mathbb{R} 
 | \allofW\in\mathbb{W} \}$ of networks (\ref{eq:yall_main}) 
 satisfying our assumptions has Rademacher complexity bounded by 
 $\mathcal{R}_N (\mathcal{F}) \leq \frac{1}{\sqrt{N}} {\postver{\phi}} \undertilde{\postver{\psi}}$ 
 (definitions as per Figure \ref{fig:featmapdef}).}
\newline
\begin{proof}
Let $\mathcal{F}$ be the set of attainable neural networks (scalar output) and 
$\epsilon$ a Rademacher random variable.  Let ${\bf x} \sim \nu$.  Then the 
Rademacher complexity is bounded as:
\[
 \begin{array}{rl}
  \mathcal{R}_N \left( \mathcal{F} \right)
  &\!\!\!\!\triangleq
  \mathbb{E}_\nu \mathbb{E}_\epsilon \left[ \mathop{\sup}\limits_{f \in \mathcal{F}} \frac{1}{N} \sum_k \epsilon_k f \left( {\bf x}_k \right) \right] \\
  &\!\!\!\!=^{\rm{Global\;Dual}}
  \frac{1}{N} \mathbb{E}_\nu \mathbb{E}_\epsilon \left[ \mathop{\sup}\limits_{\allofW \in \mathbb{W}} \sum_k \epsilon_k 
  \leftbf \Postver{\bg{\Psi}} \left( {\allofW} \right), \Postver{\bg{\phi}} \left( {\bf x}_k \right) \rightbf_{\bf g}
  \right] \\
  &\!\!\!\!=
  \frac{1}{N} \mathbb{E}_\nu \mathbb{E}_\epsilon \left[ \mathop{\sup}\limits_{\allofW \in \mathbb{W}} 
  \leftbf \Postver{\bg{\Psi}} \left( {\allofW} \right), \sum_k \epsilon_k \Postver{\bg{\phi}} \left( {\bf x}_k \right) \rightbf_{\bf g}
  \right] \\
  &\!\!\!\!\leq^{\rm{Operator\;norm}}
  \frac{1}{N} \mathbb{E}_\nu \mathbb{E}_\epsilon \left[  \mathop{\sup}\limits_{\allofW \in \mathbb{W}} 
  \left\| \Postver{\bg{\Psi}} \left( {\allofW} \right) \right\|_{\Henum[\tau]} \sqrt{\left\| \sum_k \epsilon_k \Postver{\bg{\phi}} \left( {\bf x}_k \right) \right\|_2^2} 
  \right] \\
  &\!\!\!\!\leq^{\rm{Norm\;Bound}}
  \frac{\undertilde{\postver{\psi}}}{N} \mathbb{E}_\nu \left[ \mathbb{E}_\epsilon 
  \sqrt{\left\| \sum_k \epsilon_k \Postver{\bg{\phi}} \left( {\bf x}_k \right) \right\|_2^2}
  \right] \\
  &\!\!\!\!\leq^{\rm{Jensen}}
  \frac{\undertilde{\postver{\psi}}}{N} \mathbb{E}_\nu \left[ 
  \sqrt{\mathbb{E}_\epsilon \left\| \sum_k \epsilon_k \Postver{\bg{\phi}} \left( {\bf x}_k \right) \right\|_2^2}
  \right] \\
  &\!\!\!\!=^{\{\mathbb{E}_\epsilon \epsilon_k \epsilon_l = \delta_{k,l}\}}
  \frac{\undertilde{\postver{\psi}}}{N} \mathbb{E}_\nu \left[ 
  \sqrt{\sum_k \left\| \Postver{\bg{\phi}} \left( {\bf x}_k \right) \right\|_2^2}
  \right] \\
  &\!\!\!\!\leq^{\rm{Norm\;Bound}}
  \frac{\undertilde{\postver{\psi}}}{N} \mathbb{E}_\nu \left[ 
  \sqrt{N{\postver{\phi}}^2}
  \right] 
  =
  \frac{{\postver{\phi}} \undertilde{\postver{\psi}}}{\sqrt{N}}
 \end{array}
\]
\end{proof}

{\bf Theorem \ref{th:radbounded_lip}}
 {\em Let $\mathcal{F} = \{ f (\cdot; \allofW) : \mathbb{R}^n\to\mathbb{R} 
 | \allofW\in\mathbb{W} \}$ be the set of unbiased networks 
 (\ref{eq:yall_main}) with $L$-Lipschitz activations satisfying our 
 assumptions.  Then 
 $\mathcal{R}_N (\mathcal{F}) \leq \max_{\mathcal{S} \in \mathbb{S}} 
 \prod_{j \in \mathcal{S}} L^2 \inver{p}^{[j]} \shadx^{[j]}$, where $\mathbb{S}$ 
 is the set of all input-output paths in the network graph.}

\begin{proof}
Observe from Figure \ref{fig:featmapdef} that:
\[
 \begin{array}{rl}
  \prever{\phi}^{[\inverj,j]2} \undertilde{\prever{\psi}}^{[\inverj,j]2}
  &\!\!\!\!\leq
  \frac{
  \postver{\phi}^{[\inverj]2}
  }{
  {\abssig}^{[\inverj,j]} \left( {\abssig}^{[\inverj,j]-1} \left( \prever{\phi}^{[\inverj,j]2} \right) \postver{\phi}^{[\inverj]2} \right)
  }
  L^2 \prever{\phi}^{[\inverj,j]2} \undertilde{\postver{\psi}}^{[\inverj]2} \\
 \end{array}
\]
Note that the numerator of the factional part is linearly increasing while the 
denominator is superlinearly increasing, so we may pessimise this bound as:
\[
 \begin{array}{rl}
  \prever{\phi}^{[\inverj,j]2} \undertilde{\prever{\psi}}^{[\inverj,j]2}
  &\!\!\!\!\leq
  \frac{
  \postver{\phi}^{[\inverj]2}
  }{
  {\abssig}^{[\inverj,j]} \left( {\abssig}^{[\inverj,j]-1} \left( \prever{\phi}^{[\inverj,j]2} \right) \postver{\phi}^{[\inverj]2} \right)
  }
  L^2 \prever{\phi}^{[\inverj,j]2} \undertilde{\postver{\psi}}^{[\inverj]2} \\
  &\!\!\!\!\leq
  \mathop{\lim}\limits_{\postver{\phi}^{[\inverj]} \to 0}
  \frac{
  \postver{\phi}^{[\inverj]2}
  }{
  {\abssig}^{[\inverj,j]} \left( {\abssig}^{[\inverj,j]-1} \left( \prever{\phi}^{[\inverj,j]2} \right) \postver{\phi}^{[\inverj]2} \right)
  }
  L^2 \prever{\phi}^{[\inverj,j]2} \undertilde{\postver{\psi}}^{[\inverj]2} 
  =
  \frac{1}{a_0^{[\inverj,j]} {\abssig}^{[\inverj,j]-1} \left( \prever{\phi}^{[\inverj,j]2} \right)}
  L^2 \prever{\phi}^{[\inverj,j]2}\undertilde{\postver{\psi}}^{[\inverj]2} \\
 \end{array}
\]
As $\prever{\phi}^{[\inverj,j]2}$ is a free parameter here we can let 
$\prever{\phi}^{[\inverj,j]2} \to 0$, in which limit 
$\prever{\phi}^{[\inverj,j]2} \undertilde{\prever{\psi}}^{[\inverj,j]2} 
\leq L^2 \undertilde{\postver{\psi}}^{[\inverj]2} \leq 
L^2 p^{[\inverj]} \shadx^{[\inverj]2} \sum_{\inver{\inverj} \in \Inver{\mathbb{P}}^{[\inverj]}} 
\prever{\phi}^{[\inver{\inverj},\inverj]2} \undertilde{\prever{\psi}}^{[\inver{\inverj},\inverj]2}$.  
The result follows recursively, then upper bounding with the pathwise product.
\end{proof}

{\bf Theorem \ref{th:radbounded_local}}
 {\em The set $\mathcal{F}_\changein = \{\changein f ( \cdot ; \changein 
 \allofW) : \mathbb{R}^n \to \mathbb{R} | \changein \allofW \in 
 \mathbb{W}_\changein \}$ of change in neural-network operation satisfying 
 (\ref{eq:stepbnddef}) has Rademacher complexity $\mathcal{R}_N (\mathcal{F}) 
 \leq\frac{1}{\sqrt{N}} {\postver{\phi}}_{\changein} 
 {\postver{\psi}}_{\changein}$ (defined in Figure \ref{fig:changein_feat}).}
\newline
\begin{proof}
Let $\mathcal{F}_\changein$ be the set of attainable changes in neural networks 
(scalar output) and $\epsilon$ a Rademacher random variable.  Let ${\bf x} \sim 
\nu$.  Then the Rademacher complexity is bounded as:
\[
 \begin{array}{rl}
  \mathcal{R}_N \left( \mathcal{F}_\changein \right)
  &\!\!\!\!\triangleq
  \mathbb{E}_\nu \mathbb{E}_\epsilon \left[ \mathop{\sup}\limits_{\changein f \in \mathcal{F}} \frac{1}{N} \sum_k \epsilon_k \changein f \left( {\bf x}_k \right) \right] \\
  &\!\!\!\!=^{\rm{Local\;Dual}}
  \frac{1}{N} \mathbb{E}_\nu \mathbb{E}_\epsilon \left[ \mathop{\sup}\limits_{\changein \allofW \in \mathbb{W}_\changein} \sum_k \epsilon_k 
  \leftbf \Postver{\bg{\Psi}}_\changein \left( \changein {\allofW} \right), \Postver{\bg{\phi}}_\changein \left( {\bf x}_k \right) \rightbf_{\bf g}
  \right] \\
  &\!\!\!\!=
  \frac{1}{N} \mathbb{E}_\nu \mathbb{E}_\epsilon \left[ \mathop{\sup}\limits_{\changein \allofW \in \mathbb{W}_\changein} 
  \leftbf \Postver{\bg{\Psi}}_\changein \left( \changein {\allofW} \right), \sum_k \epsilon_k \Postver{\bg{\phi}}_\changein \left( {\bf x}_k \right) \rightbf_{\bf g}
  \right] \\
  &\!\!\!\!\leq^{\rm{Cauchy\;Schwarz}}
  \frac{1}{N} \mathbb{E}_\nu \mathbb{E}_\epsilon \left[  \mathop{\sup}\limits_{\changein \allofW \in \mathbb{W}_\changein} 
  \left\| \Postver{\bg{\Psi}}_\changein \left( \changein {\allofW} \right) \right\|_2 \sqrt{\left\| \sum_k \epsilon_k \Postver{\bg{\phi}}_\changein \left( {\bf x}_k \right) \right\|_2^2} 
  \right] \\
  &\!\!\!\!\leq^{\rm{Norm\;Bound}}
  \frac{{\postver{\psi}_\changein}}{N} \mathbb{E}_\nu \left[ \mathbb{E}_\epsilon 
  \sqrt{\left\| \sum_k \epsilon_k \Postver{\bg{\phi}}_\changein \left( {\bf x}_k \right) \right\|_2^2}
  \right] \\
  &\!\!\!\!\leq^{\rm{Jensen}}
  \frac{{\postver{\psi}}_{\changein}}{N} \mathbb{E}_\nu \left[ 
  \sqrt{\mathbb{E}_\epsilon \left\| \sum_k \epsilon_k \Postver{\bg{\phi}}_\changein \left( {\bf x}_k \right) \right\|_2^2}
  \right] \\
  &\!\!\!\!=^{\{\mathbb{E}_\epsilon \epsilon_k \epsilon_l = \delta_{k,l}\}}
  \frac{{\postver{\psi}}_{\changein}}{N} \mathbb{E}_\nu \left[ 
  \sqrt{\sum_k \left\| \Postver{\bg{\phi}}_\changein \left( {\bf x}_k \right) \right\|_2^2}
  \right] \\
  &\!\!\!\!\leq^{\rm{Norm\;Bound}}
  \frac{{\postver{\psi}}_{\changein}}{N} \mathbb{E}_\nu \left[ 
  \sqrt{N{\postver{\phi}}_{\changein}^2}
  \right] 
  =
  \frac{{\postver{\phi}}_{\changein} {\postver{\psi}}_{\changein}}{\sqrt{N}}
 \end{array}
\]
\end{proof}

\section{Derivation of the Exact (LeNK) Representor Theory} \label{sup:reptheory}

In this section we present our derivation of the LeNK-based representor 
theory.  Our approach to this proof is to start from the first layer (recall 
that we are only concerned with layer-wise networks) and then systematically 
feed-forward through the network to the output, at every stage deriving a 
representor theory for that layer in terms of an intermediate kernel that 
builds off the construction for the previous layer.  Then the final 
representor theory is simply the intermediate representor theory for the 
output of layer $D-1$.  To demonstrate the ``kernel-ness'' at each layer 
we construct an explicit feature map from which the kernel (for that layer) 
derives using the kernel trick.  Before proceeding we note the following:
\begin{enumerate}
 \item The feature map necessarily includes the weights and biases, leading 
       to {\em kernel warping} as discussed in the body of the paper.
 \item The utility of the feature maps is that the demonstrated (and allow 
       us to derive) the LeNK.  Beyond that their complexity is not a concern.
 \item For purely practical reasons we have chosen to use the indexed tensor 
       style notation here.  This leads to significant notational clutter, but 
       remove potential ambiguities (at least for the authors) that we found 
       resulted from premature ``cleaning up'' of the expressions involved.  
       Note that we have cleaned up the final formulae in the body of the 
       paper (post-derivation) to aid readability and interperability.
 \item The layerwise assumption appears to be necessary for kernelisation.  
       Roughly speaking, in more general case the feedback phase ``picks up'' 
       terms from all paths from output node to input node, which subsequently 
       get fed-forward through all paths, resulting in terms on the 
       feed-forward process not ``pairing up'' appropriately with elements 
       in the nodes (ie. the weights on path $A$ influence ${\bg{\alpha}}^{[0]}$, 
       and this effects the change in weights on path $B$, where there are no 
       corresponding weights with which to connect them through the kernel 
       trick).
 \item The non-regularized assumption is required to avoid untenable levels of 
       complexity.  We note that it is not difficult to include the effects of 
       regularisation for a single layer network, but for deeper networks the 
       number of terms involved grows with each layer and quickly becomes 
       unmanageable.
\end{enumerate}

To begin, note that by the definition of back-propagation (gradient descent) 
in the un-regularized case:
\[
 \begin{array}{rl}
  \postver{\bg{\alpha}}_{\{l\}}^{[D-1]}
  &\!\!\!\!=
  -\eta
  \frac{\partial}{\partial \postver{\bf x}_{\{l\}}^{[D-1]}} R \left( \postver{\bf x}_{\{l\}}^{[D-1]} - {\bf y}_{\{l\}} \right) \\

  \postver{\bg{\alpha}}_{\{l\}}^{[j]}
  &\!\!\!\!=
  \mathop{\sum}\limits_{\inverj : j \in \Inver{\mathbb{P}}^{[\inverj]}}
  \tau^{[j,\inverj](1)} \left( \postver{\bf x}_{\{l\}}^{[j]} \right)
  \odot
  \prever{\bg{\alpha}}_{\{l\}}^{[j,\inverj]} 
  \;\;\;\; \forall j \ne D-1 \\

  \prever{\bg{\alpha}}_{\{l\}}^{[\inverj,j]}
  &\!\!\!\!=
  {\bf W}^{[\inverj,j]}
  \postver{\bg{\alpha}}_{\{l\}}^{[j]}
  \;\;\;\; \forall \inverj \in \Inver{\mathbb{P}}^{[j]} \\
 \end{array}
\]
and:
\[
 \begin{array}{rl}
  \changein {\bf W}^{[\inverj,j]}
  &\!\!\!\!=
  \sum_l
  \prever{\bf x}_{\{l\}}^{[\inverj,j]}
  \postver{\bg{\alpha}}_{\{l\}}^{[j]\tsp} \\

  \changein {\bf b}^{[j]\tsp}
  &\!\!\!\!=
  \sum_l
  \shada
  \postver{\bg{\alpha}}_{\{l\}}^{[j]\tsp} \\
 \end{array}
\]

Hence:
\[
{{
 \begin{array}{rl}
  \changein \postver{\bf x}^{[j]}
  &\!\!\!\!=
  \shada \changein {\bf b}^{[j]}
  +
  \mathop{\sum}\limits_{\inverj \in \Inver{\mathbb{P}}^{[j]}}
  \changein {\bf W}^{[\inverj,j]\tsp}
  \prever{\bf x}^{[\inverj,j]}
  +
  \mathop{\sum}\limits_{\inverj \in \Inver{\mathbb{P}}^{[j]}}
  {\bf W}^{[\inverj,j]\tsp}
  \changein \prever{\bf x}^{[\inverj,j]}
  +
  \mathop{\sum}\limits_{\inverj \in \Inver{\mathbb{P}}^{[j]}}
  \changein {\bf W}^{[\inverj,j]\tsp}
  \changein \prever{\bf x}^{[\inverj,j]} \\
 \end{array}
}}
\]
and so, recalling that we are assuming a layer-wise network (ie. 
$\Inver{\mathbb{P}}^{[j]} = \{j-1\}$ $\forall j>0$) with 
$\tau^{[-1,0]} (\zeta) = \zeta$, we have:
\[
 \begin{array}{rl}
  \changein \postver{\bf x}^{[j]}
  &\!\!\!\!=
  \left[ \begin{array}{c}
  \changein {\bf W}^{[j-1,j]} \\
  \changein {\bf b}^{[j]^\tsp} \\
  \end{array} \right]^\tsp
  \left[ \begin{array}{c}
  \prever{\bf x}^{[j-1,j]} \\
  \shada \\
  \end{array} \right]
  +
  \left[ \begin{array}{c}
  {\bf W}^{[j-1,j]} \\
  {\bf b}^{[j]^\tsp} \\
  \end{array} \right]^\tsp
  \left[ \begin{array}{c}
  \changein \prever{\bf x}^{[j-1,j]} \\
  0 \\
  \end{array} \right]
  +
  \left[ \begin{array}{c}
  \changein {\bf W}^{[j-1,j]} \\
  \changein {\bf b}^{[j]^\tsp} \\
  \end{array} \right]^\tsp
  \left[ \begin{array}{c}
  \changein \prever{\bf x}^{[j-1,j]} \\
  0 \\
  \end{array} \right] \;\; \forall j > 0
 \end{array}
\]
with the recursive base-case:
\[
 \begin{array}{rl}
  \changein \postver{\bf x}^{[0]}
  &\!\!\!\!=
  \left[ \begin{array}{c}
  \changein {\bf W}^{[-1,0]} \\
  \changein {\bf b}^{[0]^\tsp} \\
  \end{array} \right]^\tsp
  \left[ \begin{array}{c}
  \prever{\bf x}^{[-1,0]} \\
  \shada \\
  \end{array} \right] \\
 \end{array}
\]

To avoid even more notational clutter than that which unavoidably follows, we 
define $\allofW_+ = \allofW + \changein \allofW$.  Using Einstein the 
summation convention:
\[
 \begin{array}{rl}
  \changein \postver{x}_{i_0}^{[0]}
  &\!\!\!\!=
  \sum_{l}
  \Postver{K}^{[0]}{}_{i_0}{}^{i_0^1} \left( {\bf x}, {\bf x}_{\{l\}}; \allofW_+, \allofW \right)
  \postver{{\alpha}}_{\{l\}i_0^1}^{[j]} \\
 \end{array}
\]
where:
\[
 \begin{array}{rl}
  \Sigma^{[0]} \left( {\bf x}', {\bf x}^\backprime \right)
  &\!\!\!\!=
  \shada^2 + \prever{\bf x}^{\prime[-1,0]\tsp} \prever{\bf x}^{\backprime[-1,0]}
  =
  \Prever{\bg{\varphi}}^{[0]} \left( {\bf x}' \right)^\tsp
  \Prever{\bg{\varphi}}^{[0]} \left( {\bf x}^\backprime \right) \\

  \Postver{K}^{[0]}{}_{i_0}{}^{i'_0} \left( {\bf x}', {\bf x}^\backprime; \allofW', \allofW^\backprime \right)
  &\!\!\!\!=
  \delta_{i_0}{}^{i'_0} \Sigma^{[0]} \left( {\bf x}', {\bf x}^\backprime \right)
  =
  \left<
  \Postver{\bf F}^{[0]}{}_{:i_0 } \left( {\bf x}';           \allofW'           \right),
  \Postver{\bf F}^{[0]}{}^{:i'_0} \left( {\bf x}^\backprime; \allofW^\backprime \right)
  \right> \\
 \end{array}
\]
and:
\[
 \begin{array}{rl}
  \Prever{\bg{\varphi}}^{[0]} \left( {\bf x}' \right)
  &\!\!\!\!=
  \left[ \begin{array}{c} \shada \\ \prever{\bf x}^{\prime[-1,0]} \\ \end{array} \right] \\

  \Postver{\bf F}^{[0]}_{:i_0} \left( {\bf x}'; \allofW^\prime \right)
  &\!\!\!\!=
  \left[ \delta_{i_0,i'_0} \right]_{i'_0} \otimes \Prever{\bg{\varphi}}^{[0]} \left( {\bf x}' \right) \\
 \end{array}
\]

Moving forward:
\[
{\!\!\!\!\!\!\!\!\!\!\!\!\!\!\!\!\!\!\!\!{
 \begin{array}{rl}
  \changein \prever{x}^{[0,1]}_{i_0}
  &\!\!\!\!=
  \mathop{\sum}\limits_{l_1}
  {{\tau}}^{[0,1](1)} \left( \postver{x}_{i_0}^{[0]} \right)
  \Postver{K}^{[0]}{}_{i_0}{}^{i_0^1} \left( {\bf x}, {\bf x}_{\{l_1\}}; \allofW_+, \allofW \right)
  \postver{{\alpha}}_{\{l_1\}i_0^1}^{[0]}

  + \ldots \\ \ldots &\!\!\!\!+
  \frac{1}{2!}
  \mathop{\sum}\limits_{l_1,l_2}
  {{\tau}}^{[0,1](2)} \left( \postver{x}_{i_0}^{[0]} \right)
  \Postver{K}^{[0]}{}_{i_0}{}^{i_0^1} \left( {\bf x}, {\bf x}_{\{l_1\}}; \allofW_+, \allofW \right)
  \Postver{K}^{[0]}{}_{i_0}{}^{i_0^2} \left( {\bf x}, {\bf x}_{\{l_2\}}; \allofW_+, \allofW \right)
  \postver{{\alpha}}_{\{l_1\}i_0^1}^{[0]}
  \postver{{\alpha}}_{\{l_2\}i_0^2}^{[0]}

  + \ldots \\ \ldots &\!\!\!\!+
  \frac{1}{3!}
  \mathop{\sum}\limits_{l_1,l_2,l_3}
  {{\tau}}^{[0,1](3)} \left( \postver{x}_{i_0}^{[0]} \right)
  \Postver{K}^{[0]}{}_{i_0}{}^{i_0^1} \left( {\bf x}, {\bf x}_{\{l_1\}}; \allofW_+, \allofW \right)
  \Postver{K}^{[0]}{}_{i_0}{}^{i_0^2} \left( {\bf x}, {\bf x}_{\{l_2\}}; \allofW_+, \allofW \right)
  \Postver{K}^{[0]}{}_{i_0}{}^{i_0^3} \left( {\bf x}, {\bf x}_{\{l_3\}}; \allofW_+, \allofW \right)
  \ldots \\ & \ldots \;\;\;\;
  \postver{{\alpha}}_{\{l_1\}i_0^1}^{[0]}
  \postver{{\alpha}}_{\{l_2\}i_0^2}^{[0]}
  \postver{{\alpha}}_{\{l_3\}i_0^3}^{[0]}

  + \ldots \\ \ldots &\!\!\!\!+
  \ldots 
 \end{array}
}}
\]
we recall:
\[
 \begin{array}{rl}
  \postver{\bg{\alpha}}_{\{l\}}^{[0]}
  &\!\!\!\!=
  {{\tau}}^{[0,1](1)} \left( \postver{\bf x}_{\{l\}}^{[0]} \right)
  \odot
  \prever{\bg{\alpha}}_{\{l\}}^{[0,1]} \\
 \end{array}
\]
and so:
\[
{{
 \begin{array}{rl}
  \changein \prever{x}_{i_0}^{[0,1]}
  &\!\!\!\!=
  \mathop{\sum}\limits_{k \in \mathbb{Z}_+}
  \frac{1}{k!}
  \mathop{\sum}\limits_{l_1,l_2,\ldots,l_k}
  \prever{K}^{[0,1]}{}_{i_0}{}^{i_0^1,\ldots,i_0^k}
  \left(
  \left\{ \featarg{\bf x}{k} \right\},
  \left\{
  \featarg{{\bf x}_{\{l_1\}}}{1},
  \featarg{{\bf x}_{\{l_2\}}}{1},
  \ldots,
  \featarg{{\bf x}_{\{l_k\}}}{1}
  \right\};
  \allofW_+, \allofW
  \right)
  \prever{{\alpha}}_{\{l_1\}i_0^1}^{[0,1]}
  \prever{{\alpha}}_{\{l_2\}i_0^2}^{[0,1]}
  \ldots
  \prever{{\alpha}}_{\{l_k\}i_0^k}^{[0,1]}
 \end{array}
}}
\]
where:
\[
{{
 \begin{array}{l}
  \prever{K}^{[0,1]}{}_{i'_0,i''_0,\ldots,i''''_0}{}^{i_0^{\backprime},i_0^{\backprime\backprime},\ldots,i_0^{\backprime\backprime\backprime\backprime}}
  \left(
  \left\{ \featarg{{\bf x}'}{k'}, \featarg{{\bf x}''}{k''}, \ldots, \featarg{{\bf x}''''}{k''''} \right\},
  \left\{ \featarg{{\bf x}^\backprime}{k^\backprime}, \featarg{{\bf x}^{\backprime\backprime}}{k^{\backprime\backprime}}, \ldots, \featarg{{\bf x}^{\backprime\backprime\backprime\backprime}}{k^{\backprime\backprime\backprime\backprime}} \right\}; 
  \allofW', \allofW^\backprime
  \right)
  \ldots \\ \;\;\;\;\;\;\;\;\;\;\;\; =
  \left<
  \Prever{\bf F}^{[0,1]}{}_{:i'_0,i''_0,\ldots,i''''_0}
  \left(
  \left\{ \featarg{{\bf x}'}{k'}, \featarg{{\bf x}''}{k''}, \ldots, \featarg{{\bf x}''''}{k''''} \right\};
  \allofW'
  \right),
  \Prever{\bf F}^{[0,1]}{}^{:i_0^{\backprime},i_0^{\backprime\backprime},\ldots,i_0^{\backprime\backprime\backprime\backprime}}
  \left(
  \left\{ \featarg{{\bf x}^\backprime}{k^\backprime}, \featarg{{\bf x}^{\backprime\backprime}}{k^{\backprime\backprime}}, \ldots, \featarg{{\bf x}^{\backprime\backprime\backprime\backprime}}{k^{\backprime\backprime\backprime\backprime}} \right\};
  \allofW^\backprime
  \right)
  \right>
 \end{array}
}}
\]
(we require that $k' + k'' + \ldots + k'''' = k^\backprime + k^{\backprime\backprime} + \ldots + k^{\backprime\backprime\backprime\backprime}$) and:
\[
{{
 \begin{array}{r}
  \Prever{\bf F}^{[0,1]}{}_{:i'_0,i''_0,\ldots,i''''_0}
  \left(
  \left\{ \featarg{{\bf x}'}{k'}, \featarg{{\bf x}''}{k''}, \ldots, \featarg{{\bf x}''''}{k''''} \right\};
  \allofW'
  \right)
  =
  {{\tau}}^{[0,1](k')  } \left( \postver{x}^{\prime[0]}{}_{i'_0} \right) \Postver{\bf F}^{[0]}{}_{:i'_0} \left( {\bf x}'; \allofW' \right)^{\otimes k'}
  \otimes
  \ldots \;\;\;\;\;\;\;\;\;\;\;\; \\ \ldots
  {{\tau}}^{[0,1](k'') } \left( \postver{x}^{\prime\prime[0]}{}_{i''_0} \right) \Postver{\bf F}^{[0]}{}_{:i''_0} \left( {\bf x}''; \allofW' \right)^{\otimes k''}
  \otimes
  \ldots
  \otimes
  {{\tau}}^{[0,1](k'''') } \left( \postver{x}^{\prime\prime\prime\prime[0]}{}_{i''''_0} \right) \Postver{\bf F}^{[0]}{}_{:i''''_0} \left( {\bf x}''''; \allofW' \right)^{\otimes k''''}
 \end{array}
}}
\]
from which we observe that:
\[
{{
 \begin{array}{l}
  \prever{K}^{[0,1]}{}_{i_0}{}^{i_0^1,\ldots,i_0^k}
  \left(
  \left\{ \featarg{\bf x}{k} \right\},
  \left\{ \featarg{{\bf x}_{\{l_1\}}}{1}, \ldots, \featarg{{\bf x}_{\{l_k\}}}{1} \right\};
  \allofW_+, \allofW
  \right)
  = \ldots \\ \;\;\;\;\;\; \ldots =
  {{\tau}}^{[0,1](k)} \left( \postver{x}^{[0]}{}_{{i}_0^0} \right)
  \postver{K}^{[0]}{}_{i_0}{}^{i_0^1} \left( {\bf x}, {\bf x}_{\{l_1\}}; \allofW_+, \allofW \right)
  \postver{K}^{[0]}{}_{i_0}{}^{i_0^2} \left( {\bf x}, {\bf x}_{\{l_2\}}; \allofW_+, \allofW \right)
  \ldots
  \postver{K}^{[0]}{}_{i_0}{}^{i_0^k} \left( {\bf x}, {\bf x}_{\{l_k\}}; \allofW_+, \allofW \right)
  \ldots \\ \;\;\;\;\;\;\;\;\;\;\;\; \ldots
  {{\tau}}^{[0,1](1)} \left( \postver{x}_{\{l_1\}}^{[0]}{}_{{i}_0^1} \right)
  {{\tau}}^{[0,1](1)} \left( \postver{x}_{\{l_2\}}^{[0]}{}_{{i}_0^2} \right)
  \ldots
  {{\tau}}^{[0,1](1)} \left( \postver{x}_{\{l_k\}}^{[0]}{}_{{i}_0^k} \right)
 \end{array}
}}
\]
as required (the general case can be similarly expanded but the form is 
non-trivial as it the kernel splitting depends on the indices $k',k'',\ldots, 
k''''$ and $k^{\backprime},k^{\backprime\backprime},\ldots,k^{\backprime 
\backprime\backprime\backprime}$).

Recall that:
\[
 \begin{array}{rl}
  {\Prever{\bg{\alpha}}}_{\{l\}}^{[0,1]}
  &\!\!\!\!=
  {\bf W}^{[0,1]}
  {\postver{\bg{\alpha}}}_{\{l\}}^{[1]}
 \end{array}
\]
It follows that:
\[
{{
 \begin{array}{rl}
  \left[ \begin{array}{c}
  \changein {\bf W}_{:i_1}^{[0,1]} \\
  \changein b_{i_1}^{[1]} \\
  \end{array} \right]^\tsp
  \left[ \begin{array}{c}
  \prever{\bf x}^{[0,1]} \\
  \shada \\
  \end{array} \right]
  &\!\!\!\!= 
  \sum_l
  \postver{{\alpha}}_{\{l\}i_1}^{[1]}
  \left[ \begin{array}{c}
  \prever{\bf x}^{[0,1]}_{\{l\}} \\
  \shada \\
  \end{array} \right]^\tsp
  \left[ \begin{array}{c}
  \prever{\bf x}^{[0,1]} \\
  \shada \\
  \end{array} \right] \\

  &\!\!\!\!= 
  \sum_l
  \Sigma^{[1]} \left( {\bf x}, {\bf x}_{\{l\}} \right)
  \Postver{{\alpha}}_{\{l\}i_1}^{[1]} \\
 \end{array}
}}
\]
and:
\[
{{
 \begin{array}{rl}
  \left[ \begin{array}{c}
  {\bf W}_{:i_1}^{[0,1]} \\
  b_{i_1}^{[1]} \\
  \end{array} \right]^\tsp
  \left[ \begin{array}{c}
  \changein \prever{\bf x}^{[0,1]} \\
  0 \\
  \end{array} \right]
  &\!\!\!\!= 
  {\bf W}^{[1]\tsp}_{:i_1}
  \changein \prever{\bf x}^{[0,1]} \\

  &\!\!\!\!\!\!\!\!\!\!\!\!\!\!\!\!\!\!\!\!\!\!\!\!\!\!\!\!\!\!\!\!= 
  \mathop{\sum}\limits_{k \geq 1}
  \frac{1}{k!}
  \mathop{\sum}\limits_{l_1,\ldots,l_k}
  R{}_{i_1}{}^{i_1^1,i_1^2,\ldots,i_1^k}
  \left(
  \left\{ \featarg{{\bf x}}{k} \right\},
  \left\{ \featarg{{\bf x}_{\{l_1\}}}{1}, \featarg{{\bf x}_{\{l_2\}}}{1}, \ldots \featarg{{\bf x}_{\{l_k\}}}{1} \right\}
  \right)
  \postver{{\alpha}}_{\{l_1\}i_1^1}^{[1]}
  \postver{{\alpha}}_{\{l_2\}i_1^2}^{[1]}
  \ldots
  \postver{{\alpha}}_{\{l_k\}i_1^k}^{[1]}
 \end{array}
}}
\]
where:
\[
{{
 \begin{array}{r}
  R{}_{i'_1,i''_1,\ldots,i''''_1}{}^{i_1^{\backprime},i_1^{\backprime\backprime},\ldots,i_1^{\backprime\backprime\backprime\backprime}}
  \left(
  \left\{ \featarg{{\bf x}'}{k'}, \featarg{{\bf x}''}{k''}, \ldots, \featarg{{\bf x}''''}{k''''} \right\},
  \left\{ \featarg{{\bf x}^{\backprime}}{k^{\backprime}}, \featarg{{\bf x}^{\backprime\backprime}}{k^{\backprime\backprime}}, \ldots, \featarg{{\bf x}^{\backprime\backprime\backprime\backprime}}{k^{\backprime\backprime\backprime\backprime}} \right\},
  \right)
  \ldots \;\;\;\;\;\;\;\;\;\;\;\;\;\;\;\;\;\;\;\;\;\;\;\;\;\;\;\;\;\;\;\; \\
 \begin{array}{rl}
  &\!\!\!\!=
  {W}^{[0,1]}{}^{i'_0}{}_{i'_1}
  {W}^{[0,1]}{}^{i''_0}{}_{i''_1}
  \ldots
  {W}^{[0,1]}{}^{i''''_0}{}_{i''''_1}
  {W}^{[0,1]}{}^{i_0^{\backprime}}{}_{i_1^{\backprime}}
  {W}^{[0,1]}{}^{i_0^{\backprime\backprime}}{}_{i_1^{\backprime\backprime}}
  \ldots
  {W}^{[0,1]}{}^{i_0^{\backprime\backprime\backprime\backprime}}{}_{i_1^{\backprime\backprime\backprime\backprime}}
  \ldots \\ &\ldots
  \prever{K}^{[0,1]}{}_{i'_0,i''_0,\ldots,i''''_0}{}^{i_0^{\backprime},i_0^{\backprime\backprime},\ldots,i_0^{\backprime\backprime\backprime\backprime}}
  \left(
  \left\{ \featarg{{\bf x}'}{k'}, \featarg{{\bf x}''}{k''}, \ldots, \featarg{{\bf x}''''}{k''''} \right\},
  \left\{ \featarg{{\bf x}^{\backprime}}{k^{\backprime}}, \featarg{{\bf x}^{\backprime\backprime}}{k^{\backprime\backprime}}, \ldots, \featarg{{\bf x}^{\backprime\backprime\backprime\backprime}}{k^{\backprime\backprime\backprime\backprime}} \right\};
  \allofW_+, \allofW
  \right) \\
  &\!\!\!\!=
  \left<
  \Postver{\bf G}_K^{[1]}{}_{i'_1,i''_1,\ldots,i''''_1}
  \left(
  \left\{ \featarg{{\bf x}'}{k'}, \featarg{{\bf x}''}{k''}, \ldots, \featarg{{\bf x}''''}{k''''} \right\}
  \right),
  \Postver{\bf G}_R^{[1]}{}_{i_1^{\backprime},i_1^{\backprime\backprime},\ldots,i_1^{\backprime\backprime\backprime\backprime}}
  \left(
  \left\{ \featarg{{\bf x}^{\backprime}}{k^{\backprime}}, \featarg{{\bf x}^{\backprime\backprime}}{k^{\backprime\backprime}}, \ldots, \featarg{{\bf x}^{\backprime\backprime\backprime\backprime}}{k^{\backprime\backprime\backprime\backprime}} \right\}
  \right)
  \right>
 \end{array} \\
 \end{array}
}}
\]
is a tensor-valued kernel, for which:
\[
{{
 \begin{array}{l}
  \Postver{\bf G}_K^{[1]}{}_{i'_1,i''_1,\ldots,i''''_1}
  \left(
  \left\{ \featarg{{\bf x}'}{k'}, \featarg{{\bf x}''}{k''}, \ldots, \featarg{{\bf x}''''}{k''''} \right\}
  \right)
  = \ldots \\ \;\;\;\; \ldots =
  {W}^{[0,1]}{}^{i'_0}{}_{i'_1}
  {W}^{[0,1]}{}^{i''_0}{}_{i''_1}
  \ldots
  {W}^{[0,1]}{}^{i''''_0}{}_{i''''_1}
  \Prever{\bf F}^{[0,1]}{}_{i'_0,i''_0,\ldots,i''''_0}
  \left(
  \left\{ \featarg{{\bf x}'}{k'}, \featarg{{\bf x}''}{k''}, \ldots, \featarg{{\bf x}''''}{k''''} \right\}; \allofW_+
  \right) \\

  \Postver{\bf G}_R^{[1]}{}_{i'_1,i''_1,\ldots,i''''_1}
  \left(
  \left\{ \featarg{{\bf x}'}{k'}, \featarg{{\bf x}''}{k''}, \ldots, \featarg{{\bf x}''''}{k''''} \right\}
  \right)
  = \ldots \\ \;\;\;\; \ldots =
  {W}^{[0,1]}{}^{i'_0}{}_{i'_1}
  {W}^{[0,1]}{}^{i''_0}{}_{i''_1}
  \ldots
  {W}^{[0,1]}{}^{i''''_0}{}_{i''''_1}
  \Prever{\bf F}^{[0,1]}{}_{i'_0,i''_0,\ldots,i''''_0}
  \left(
  \left\{ \featarg{{\bf x}'}{k'}, \featarg{{\bf x}''}{k''}, \ldots, \featarg{{\bf x}''''}{k''''} \right\}; \allofW
  \right) \\
 \end{array}
}}
\]
Furthermore:
\[
{{
 \begin{array}{l}
  \left[ \begin{array}{c}
  \changein {\bf W}_{:i_1}^{[0,1]} \\
  \changein b_{i_1}^{[1]} \\
  \end{array} \right]^\tsp
  \left[ \begin{array}{c}
  \changein \prever{\bf x}^{[0,1]} \\
  0 \\
  \end{array} \right]
  = 
  \sum_l \alpha_{\{l\}i_1}^{[1]}
  \left[ \begin{array}{c}
  \prever{\bf x}^{[0,1]}_{\origin \{l\}} \\
  \shada \\
  \end{array} \right]^\tsp
  \left[ \begin{array}{c}
  \changein \prever{\bf x}^{[0,1]} \\
  0 \\
  \end{array} \right] \\

  \;\;\;\;=
  \mathop{\sum}\limits_{k \geq 1}
  \frac{1}{k!}
  \mathop{\sum}\limits_{l_1,l_2,\ldots,l_k}
  \changein R{}_{i'_1}{}^{i_1^1,i_1^2,\ldots,i_1^k}
  \left(
  \left\{ \featarg{{\bf x}}{k} \right\},
  \left\{ \featarg{{\bf x}_{\{l_1\}}}{1}, \featarg{{\bf x}_{\{l_2\}}}{1}, \ldots, \featarg{{\bf x}_{\{l_k\}}}{1} \right\}
  \right)
  \postver{{\alpha}}_{\{l_1\}i_1^1}^{[1]}
  \postver{{\alpha}}_{\{l_2\}i_1^2}^{[1]}
  \ldots
  \postver{{\alpha}}_{\{l_k\}i_1^k}^{[1]}
 \end{array}
}}
\]
where:
\[
{{
 \begin{array}{l}
  \changein R{}_{i'_1,i''_1,\ldots,i''''_1}{}^{i_1^{\backprime},i_1^{\backprime\backprime},\ldots,i_1^{\backprime\backprime\backprime\backprime}}
  \left(
  \left\{ \featarg{{\bf x}'}{k'}, \featarg{{\bf x}''}{k''}, \ldots, \featarg{{\bf x}''''}{k''''} \right\},
  \left\{ \featarg{{\bf x}^{\backprime}}{k^{\backprime}}, \featarg{{\bf x}^{\backprime\backprime}}{k^{\backprime\backprime}}, \ldots, \featarg{{\bf x}^{\backprime\backprime\backprime\backprime}}{k^{\backprime\backprime\backprime\backprime}} \right\},
  \right)
  \ldots \\
 \;\;\;\;\begin{array}{rl}
  &\!\!\!\!=
  \left(
  {W}_+^{[0,1]}{}^{i'_0}{}_{i'_1}
  {W}_+^{[0,1]}{}^{i''_0}{}_{i''_1}
  \ldots
  {W}_+^{[0,1]}{}^{i''''_0}{}_{i''''_1}
  -
  {W}^{[0,1]}{}^{i'_0}{}_{i'_1}
  {W}^{[0,1]}{}^{i''_0}{}_{i''_1}
  \ldots
  {W}^{[0,1]}{}^{i''''_0}{}_{i''''_1}
  \right)
  \ldots \\ &\ldots
  {W}^{[0,1]}{}^{i_0^{\backprime}}{}_{i_1^{\backprime}}
  {W}^{[0,1]}{}^{i_0^{\backprime\backprime}}{}_{i_1^{\backprime\backprime}}
  \ldots
  {W}^{[0,1]}{}^{i_0^{\backprime\backprime\backprime\backprime}}{}_{i_1^{\backprime\backprime\backprime\backprime}}
  \ldots \\ &\ldots
  \prever{K}^{[0,1]}{}_{i'_0,i''_0,\ldots,i''''_0}{}^{i_0^{\backprime},i_0^{\backprime\backprime},\ldots,i_0^{\backprime\backprime\backprime\backprime}}
  \left(
  \left\{ \featarg{{\bf x}'}{k'}, \featarg{{\bf x}''}{k''}, \ldots, \featarg{{\bf x}''''}{k''''} \right\},
  \left\{ \featarg{{\bf x}^{\backprime}}{k^{\backprime}}, \featarg{{\bf x}^{\backprime\backprime}}{k^{\backprime\backprime}}, \ldots, \featarg{{\bf x}^{\backprime\backprime\backprime\backprime}}{k^{\backprime\backprime\backprime\backprime}} \right\};
  \allofW_+, \allofW
  \right) \\
  &\!\!\!\!=
  \left<
  \changein \Postver{\bf G}_K^{[1]}{}_{i'_1,i''_1,\ldots,i''''_1}
  \left(
  \left\{ \featarg{{\bf x}'}{k'}, \featarg{{\bf x}''}{k''}, \ldots, \featarg{{\bf x}''''}{k''''} \right\}
  \right),
  \Postver{\bf G}_R^{[1]}{}_{i_1^{\backprime},i_1^{\backprime\backprime},\ldots,i_1^{\backprime\backprime\backprime\backprime}}
  \left(
  \left\{ \featarg{{\bf x}^{\backprime}}{k^{\backprime}}, \featarg{{\bf x}^{\backprime\backprime}}{k^{\backprime\backprime}}, \ldots, \featarg{{\bf x}^{\backprime\backprime\backprime\backprime}}{k^{\backprime\backprime\backprime\backprime}} \right\}
  \right)
  \right>
 \end{array} \\
 \end{array}
}}
\]
given:
\[
{{
 \begin{array}{l}
  \changein \Postver{\bf G}_K^{[1]}{}_{i'_1,i''_1,\ldots,i''''_1}
  \left(
  \left\{ \featarg{{\bf x}'}{k'}, \featarg{{\bf x}''}{k''}, \ldots, \featarg{{\bf x}''''}{k''''} \right\}
  \right)
  = \ldots \\ \;\;\;\; \ldots =
  \left(
  {W}_+^{[0,1]}{}^{i'_0}{}_{i'_1}
  {W}_+^{[0,1]}{}^{i''_0}{}_{i''_1}
  \ldots
  {W}_+^{[0,1]}{}^{i''''_0}{}_{i''''_1}
  -
  {W}^{[0,1]}{}^{i'_0}{}_{i'_1}
  {W}^{[0,1]}{}^{i''_0}{}_{i''_1}
  \ldots
  {W}^{[0,1]}{}^{i''''_0}{}_{i''''_1}
  \right)
  \ldots \\ \;\;\;\;\;\;\;\; \ldots
  \Prever{\bf F}^{[0,1]}{}_{i'_0,i''_0,\ldots,i''''_0}
  \left(
  \left\{ \featarg{{\bf x}'}{k'}, \featarg{{\bf x}''}{k''}, \ldots, \featarg{{\bf x}''''}{k''''} \right\}; \allofW_+
  \right) \\
 \end{array}
}}
\]
and so we find that, overall:
\[
{{
 \begin{array}{rl}
  \changein \postver{x}^{[1]}_{i_1}
  &\!\!\!\!=
  \left[ \begin{array}{c}
  \changein {\bf W}_{:i_1^0}^{[0,1]} \\
  \changein b_{i_1^0}^{[1]} \\
  \end{array} \right]^\tsp
  \left[ \begin{array}{c}
  \changein \prever{\bf x}^{[0,1]} \\
  0 \\
  \end{array} \right]
  +
  \left[ \begin{array}{c}
  \changein {\bf W}_{:i_1}^{[0,1]} \\
  \changein b_{i_1}^{[1]} \\
  \end{array} \right]^\tsp
  \left[ \begin{array}{c}
  \prever{\bf x}_{\origin}^{[0,1]} \\
  \shada \\
  \end{array} \right]
  +
  \left[ \begin{array}{c}
  {\bf W}_{:i_1}^{[0,1]} \\
  b_{i_1}^{[1]} \\
  \end{array} \right]^\tsp
  \left[ \begin{array}{c}
  \changein \prever{\bf x}^{[0,1]} \\
  0 \\
  \end{array} \right] \\

  &\!\!\!\!=
  \mathop{\sum}\limits_{k \geq 1}
  \frac{1}{k!}
  \mathop{\sum}\limits_{l_1,l_2,\ldots,l_k}
  \postver{K}^{[1]}{}_{i_1}{}^{i_1^1,i_1^2,\ldots,i_1^k}
  \left(
  \left\{ \featarg{{\bf x}}{k} \right\},
  \left\{ \featarg{{\bf x}_{\{l_1\}}}{1}, \featarg{{\bf x}_{\{l_2\}}}{1}, \ldots, \featarg{{\bf x}_{\{l_k\}}}{1} \right\};
  \allofW_+, \allofW
  \right)
  \postver{{\alpha}}_{\{l_1\}i_1^1}^{[1]}
  \postver{{\alpha}}_{\{l_2\}i_1^2}^{[1]}
  \ldots
  \postver{{\alpha}}_{\{l_k\}i_1^k}^{[1]}
 \end{array}
}}
\]
where:
\[
{{
 \begin{array}{rl}
  \postver{K}^{[1]}{}_{i_1}{}^{i'_1}
  \left(
  \left\{ \featarg{{\bf x}'}{1} \right\},
  \left\{ \featarg{{\bf x}^\backprime}{1} \right\}; 
  \allofW', \allofW^\backprime
  \right)
  &\!\!\!\!=
  \delta{}_{i_1}{}^{i'_1}
  \Sigma^{[1]}
  \left( {\bf x}', {\bf x}^\backprime \right)
  +
  {W}^{\prime[0,1]}{}^{i'_0}{}_{i'_1}
  {W}^{\backprime[0,1]}{}^{i_0^{\backprime}}{}_{i_1^{\backprime}}
  \prever{K}^{[0,1]}{}_{i'_0}{}^{i_0^{\backprime}}
  \left(
  \left\{ \featarg{{\bf x}'}{1} \right\},
  \left\{ \featarg{{\bf x}^{\backprime}}{1} \right\};
  \allofW', \allofW^\backprime
  \right) \\
  &\!\!\!\!=
  \left<
  \Postver{\bf F}^{[1]}{}_{i'_1}
  \left(
  \left\{ \featarg{{\bf x}'}{1} \right\};
  \allofW'
  \right),
  \Postver{\bf F}^{[1]}{}_{i_1^{\backprime}}
  \left(
  \left\{ \featarg{{\bf x}^{\backprime}}{1}  \right\};
  \allofW^\backprime
  \right)
  \right> \\
 \end{array}
}}
\]
and otherwise:
\[
{{
 \begin{array}{l}
  \postver{K}^{[1]}{}_{i'_1,i''_1,\ldots,i''''_1}{}^{i_1^{\backprime},i_1^{\backprime\backprime},\ldots,i_1^{\backprime\backprime\backprime\backprime}, i_1^{\shortmid}}
  \left(
  \left\{ \featarg{{\bf x}'}{k'}, \featarg{{\bf x}''}{k''}, \ldots, \featarg{{\bf x}''''}{k''''} \right\},
  \left\{ \featarg{{\bf x}^{\backprime}}{k^{\backprime}}, \featarg{{\bf x}^{\backprime\backprime}}{k^{\backprime\backprime}}, \ldots, \featarg{{\bf x}^{\backprime\backprime\backprime\backprime}}{k^{\backprime\backprime\backprime\backprime}}, \featarg{{\bf x}^{\shortmid}}{k^{\shortmid}} \right\};
  \allofW', \allofW^\backprime
  \right)
  \ldots \\ \hfill \;\;\;\;\;\;
 \begin{array}{rl}
  &\!\!\!\!=
  {W}^{\prime[0,1]}{}^{i'_0}{}_{i'_1}
  {W}^{\prime[0,1]}{}^{i''_0}{}_{i''_1}
  \ldots
  {W}^{\prime[0,1]}{}^{i''''_0}{}_{i''''_1}
  {W}^{\backprime[0,1]}{}^{i_0^{\backprime}}{}_{i_1^{\backprime}}
  {W}^{\backprime[0,1]}{}^{i_0^{\backprime\backprime}}{}_{i_1^{\backprime\backprime}}
  \ldots
  {W}^{\backprime[0,1]}{}^{i_0^{\backprime\backprime\backprime\backprime}}{}_{i_1^{\backprime\backprime\backprime\backprime}}
  \ldots \\ &\ldots
  \prever{K}^{[0,1]}{}_{i'_0,i''_0,\ldots,i''''_0}{}^{i_0^{\backprime},i_0^{\backprime\backprime},\ldots,i_0^{\backprime\backprime\backprime\backprime}}
  \left(
  \left\{ \featarg{{\bf x}'}{k'}, \featarg{{\bf x}''}{k''}, \ldots, \featarg{{\bf x}''''}{k''''} \right\},
  \left\{ \featarg{{\bf x}^{\backprime}}{k^{\backprime}}, \featarg{{\bf x}^{\backprime\backprime}}{k^{\backprime\backprime}}, \ldots, \featarg{{\bf x}^{\backprime\backprime\backprime\backprime}}{k^{\backprime\backprime\backprime\backprime}} \right\};
  \allofW', \allofW^\backprime
  \right) \\
  &\!\!\!\!=
  \left<
  \Postver{\bf F}^{[1]}{}_{i'_1,i''_1,\ldots,i''''_1}
  \left(
  \left\{ \featarg{{\bf x}'}{k'}, \featarg{{\bf x}''}{k''}, \ldots, \featarg{{\bf x}''''}{k''''} \right\};
  \allofW'
  \right),
  \Postver{\bf F}^{[1]}{}_{i_1^{\backprime},i_1^{\backprime\backprime},\ldots,i_1^{\backprime\backprime\backprime\backprime}}
  \left(
  \left\{ \featarg{{\bf x}^{\backprime}}{k^{\backprime}}, \featarg{{\bf x}^{\backprime\backprime}}{k^{\backprime\backprime}}, \ldots, \featarg{{\bf x}^{\backprime\backprime\backprime\backprime}}{k^{\backprime\backprime\backprime\backprime}} \right\};
  \allofW^\backprime
  \right)
  \right>
 \end{array} \\
 \end{array}
}}
\]
where:
\[
 \begin{array}{rl}
  \Sigma^{[1]} \left( {\bf x}', {\bf x}^\backprime \right)
  &\!\!\!\!=
  \shada^2 + \prever{\bf x}^{\prime[0,1]\tsp} \prever{\bf x}^{\backprime[0,1]}
  =
  \Prever{\bg{\varphi}}^{[1]} \left( {\bf x}' \right)^\tsp
  \Prever{\bg{\varphi}}^{[1]} \left( {\bf x}^\backprime \right) \\
 \end{array}
\]
\[
 \begin{array}{rl}
  \Prever{\bg{\varphi}}^{[1]} \left( {\bf x}' \right)
  &\!\!\!\!=
  \left[ \begin{array}{c} \shada \\ \prever{\bf x}^{\prime[0,1]} \\ \end{array} \right] \\
 \end{array}
\]
and:
\[
{{
 \begin{array}{l}
  \Postver{\bf F}^{[1]}{}_{i'_1}
  \left(
  \left\{ \featarg{{\bf x}'}{1} \right\}; \allofW'
  \right)
  = 
  \left[ \begin{array}{c}
  \left[ \delta_{i'_1,i''_1} \right]_{i''_1} \otimes \Prever{\bg{\varphi}}^{[1]} \left( {\bf x}' \right) \\

  {W}^{\prime[0,1]}{}^{i'_0}{}_{i'_1}
  \Prever{\bf F}^{[0,1]}{}_{i'_0}
  \left(
  \left\{ \featarg{{\bf x}'}{1} \right\}; \allofW'
  \right) \\
  \end{array} \right]
 \end{array}
}}
\]
\[
{{
 \begin{array}{l}
  \Postver{\bf F}^{[1]}{}_{i'_1,i''_1,\ldots,i''''_1}
  \left(
  \left\{ \featarg{{\bf x}'}{k'}, \featarg{{\bf x}''}{k''}, \ldots, \featarg{{\bf x}''''}{k''''} \right\}; \allofW'
  \right)
  = \ldots \\ \;\;\;\; \ldots =
  {W}^{\prime[0,1]}{}^{i'_0}{}_{i'_1}
  {W}^{\prime[0,1]}{}^{i''_0}{}_{i''_1}
  \ldots
  {W}^{\prime[0,1]}{}^{i''''_0}{}_{i''''_1}
  \Prever{\bf F}^{[0,1]}{}_{i'_0,i''_0,\ldots,i''''_0}
  \left(
  \left\{ \featarg{{\bf x}'}{k'}, \featarg{{\bf x}''}{k''}, \ldots, \featarg{{\bf x}''''}{k''''} \right\}; \allofW'
  \right) \\
 \end{array}
}}
\]

For the next layer:\newline
\[
{\!\!\!\!\!\!\!\!\!\!\!\!\!\!\!\!\!\!\!\!\!\!\!\!\!\!\!\!\!\!\!\!\!\!\!\!\!\!\!\!\!\!\!\!\!\!\!\!\!\!\!\!\!\!\!\!\!\!\!\!{
 \begin{array}{l}
  \changein \prever{x}^{[1,2]}_{i_1}
  =
  \frac{1}{1!}
  {{\tau}}^{[1,2](1)} \left( \postver{x}_{\origin i_1}^{[1]} \right)
  \mathop{\sum}\limits_{k \geq 1}
  \frac{1}{k!}
  \mathop{\sum}\limits_{\!\!\begin{array}{c} {\scriptstyle l_{1,1},l_{2,1},\ldots,l_{k,1}} \\ \end{array}\!\!}
  \Postver{K}^{[1]}{}_{i_1}{}^{i_1^{1,1},i_1^{2,1},\ldots,i_1^{k,1}}
  \left(
  \left\{ \featarg{{\bf x}}{k} \right\},
  \left\{
  \featarg{{\bf x}_{\{l_{1,1}\}}}{1},
  \ldots,
  \featarg{{\bf x}_{\{l_{k,1}\}}}{1} \right\};
  \allofW_+, \allofW
  \right)
  \postver{{\alpha}}_{\{l_{1,1}\}i_1^{1,1}}^{[1]}
  \ldots
  \postver{{\alpha}}_{\{l_{k,1}\}i_1^{k,1}}^{[1]}

  + \ldots \\ \ldots +
  \frac{1}{2!}
  {{\tau}}^{[1,2](2)} \left( \postver{x}_{\origin i_1}^{[1]} \right)
  \mathop{\sum}\limits_{k \geq 1}
  \frac{1}{k!}
  \mathop{\sum}\limits_{\!\!\begin{array}{c} {\scriptstyle l_{1,1},l_{2,1},\ldots,l_{k,1}} \\ {\scriptstyle l_{1,2},l_{2,2},\ldots,l_{k,2}} \\ \end{array}\!\!}
  \Postver{K}^{[1]}{}_{i_1}{}^{i_1^{1,1},i_1^{2,1},\ldots,i_1^{k,1}}
  \left(
  \left\{ \featarg{{\bf x}}{k} \right\},
  \left\{
  \featarg{{\bf x}_{\{l_{1,1}\}}}{1},
  \ldots,
  \featarg{{\bf x}_{\{l_{k,1}\}}}{1} \right\};
  \allofW_+, \allofW
  \right)
  \postver{{\alpha}}_{\{l_{1,1}\}i_1^{1,1}}^{[1]}
  \ldots
  \postver{{\alpha}}_{\{l_{k,1}\}i_1^{k,1}}^{[1]}
  \ldots \\ \hfill \ldots
  \Postver{K}^{[1]}{}_{i_1}{}^{i_1^{1,2},i_1^{2,2},\ldots,i_1^{k,2}}
  \left(
  \left\{ \featarg{{\bf x}}{k} \right\},
  \left\{
  \featarg{{\bf x}_{\{l_{1,2}\}}}{1},
  \ldots,
  \featarg{{\bf x}_{\{l_{k,2}\}}}{1} \right\};
  \allofW_+, \allofW
  \right)
  \postver{{\alpha}}_{\{l_{1,2}\}i_1^{1,2}}^{[1]}
  \ldots
  \postver{{\alpha}}_{\{l_{k,2}\}i_1^{k,2}}^{[1]}

  + \ldots \\ \ldots +
  \frac{1}{3!}
  {{\tau}}^{[1,2](3)} \left( \postver{x}_{\origin i_1}^{[1]} \right)
  \mathop{\sum}\limits_{k \geq 1}
  \frac{1}{k!}
  \mathop{\sum}\limits_{\!\!\begin{array}{c} {\scriptstyle l_{1,1},l_{2,1},\ldots,l_{k,1}} \\ {\scriptstyle l_{1,2},l_{2,2},\ldots,l_{k,2}} \\ {\scriptstyle l_{1,3},l_{2,3},\ldots,l_{k,3}} \\ \end{array}\!\!}
  \Postver{K}^{[1]}{}_{i_1}{}^{i_1^{1,1},i_1^{2,1},\ldots,i_1^{k,1}}
  \left(
  \left\{ \featarg{{\bf x}}{k} \right\},
  \left\{
  \featarg{{\bf x}_{\{l_{1,1}\}}}{1},
  \ldots,
  \featarg{{\bf x}_{\{l_{k,1}\}}}{1} \right\};
  \allofW_+, \allofW
  \right)
  \postver{{\alpha}}_{\{l_{1,1}\}i_1^{1,1}}^{[1]}
  \ldots
  \postver{{\alpha}}_{\{l_{k,1}\}i_1^{k,1}}^{[1]}
  \ldots \\ \hfill \ldots
  \Postver{K}^{[1]}{}_{i_1}{}^{i_1^{1,2},i_1^{2,2},\ldots,i_1^{k,2}}
  \left(
  \left\{ \featarg{{\bf x}}{k} \right\},
  \left\{
  \featarg{{\bf x}_{\{l_{1,2}\}}}{1},
  \ldots,
  \featarg{{\bf x}_{\{l_{k,2}\}}}{1} \right\};
  \allofW_+, \allofW
  \right)
  \postver{{\alpha}}_{\{l_{1,2}\}i_1^{1,2}}^{[1]}
  \ldots
  \postver{{\alpha}}_{\{l_{k,2}\}i_1^{k,2}}^{[1]}
  \ldots \\ \hfill \ldots
  \Postver{K}^{[1]}{}_{i_1}{}^{i_1^{1,3},i_1^{2,3},\ldots,i_1^{k,3}}
  \left(
  \left\{ \featarg{{\bf x}}{k} \right\},
  \left\{
  \featarg{{\bf x}_{\{l_{1,3}\}}}{1},
  \ldots,
  \featarg{{\bf x}_{\{l_{k,3}\}}}{1} \right\}
  \right)
  \postver{{\alpha}}_{\{l_{1,3}\}i_1^{1,3}}^{[1]}
  \ldots
  \postver{{\alpha}}_{\{l_{k,3}\}i_1^{k,3}}^{[1]}

  + \ldots \\ \ldots + \ldots \\
 \end{array}
}}
\]
Recall:
\[
 \begin{array}{rl}
  \postver{\bg{\alpha}}_{\{l\}}^{[1]}
  &\!\!\!\!=
  {{\tau}}^{[1,2](1)} \left( \postver{\bf x}_{\{l\}}^{[1]} \right)
  \odot
  \prever{\bg{\alpha}}_{\{l\}}^{[1,2]} \\
 \end{array}
\]
and so:
\[
{\!\!\!\!\!\!\!\!\!\!\!\!\!\!\!\!\!\!\!\!\!\!\!\!\!\!\!\!\!\!\!\!\!\!\!\!\!\!\!\!\!\!\!\!\!\!{
 \begin{array}{rl}
  \changein \prever{x}_{i_1}^{[1,2]}
  &\!\!\!\!=
  \mathop{\sum}\limits_{{\bf k} \in \mathbb{Z}_+^2}
  \frac{1}{{\bf k}!}
  \mathop{\sum}\limits_{l_{\mathbb{N}_{k_0},\mathbb{N}_{k_1}}}
  \prever{K}^{[1,2]}{}_{i_1}{}^{i_1^{1,1},i_1^{2,1},\ldots,i_1^{k_0,1},i_1^{1,2},i_1^{2,2},\ldots,i_1^{k_0,2},\ldots,i_1^{1,k_1},i_1^{2,k_1},\ldots,i_1^{k_0,k_1}}
  \Bigg(
  \left\{
  \featarg{\left\{ \featarg{{\bf x}}{k_0} \right\}}{k_1}
  \right\},
  \ldots \\ & \ldots
  \left\{
  \featarg{\left\{ \featarg{{\bf x}_{\{l_{1,1  }\}}}{1}, \featarg{{\bf x}_{\{l_{2,1  }\}}}{1}, \ldots \featarg{{\bf x}_{\{l_{k_0,1  }\}}}{1} \right\}}{1},
  \featarg{\left\{ \featarg{{\bf x}_{\{l_{1,2  }\}}}{1}, \featarg{{\bf x}_{\{l_{2,2  }\}}}{1}, \ldots \featarg{{\bf x}_{\{l_{k_0,2  }\}}}{1} \right\}}{1},
  \ldots,
  \featarg{\left\{ \featarg{{\bf x}_{\{l_{1,k_1}\}}}{1}, \featarg{{\bf x}_{\{l_{2,k_1}\}}}{1}, \ldots \featarg{{\bf x}_{\{l_{k_0,k_1}\}}}{1} \right\}}{1}
  \right\};
  \allofW_+, \allofW
  \Bigg)
  \ldots \\ & \ldots 
  \prever{{\alpha}}_{\{l_{1,1  }\}i_1^{1,1  }}^{[1,2]} \prever{{\alpha}}_{\{l_{2,1  }\}i_1^{2,1  }}^{[1,2]} \ldots \prever{{\alpha}}_{\{l_{k_0,1  }\}i_1^{k_0,1  }}^{[1,2]}
  \prever{{\alpha}}_{\{l_{1,2  }\}i_1^{1,2  }}^{[1,2]} \prever{{\alpha}}_{\{l_{2,2  }\}i_1^{2,2  }}^{[1,2]} \ldots \prever{{\alpha}}_{\{l_{k_0,2  }\}i_1^{k_0,2  }}^{[1,2]}
  \ldots
  \prever{{\alpha}}_{\{l_{1,k_1}\}i_1^{1,k_1}}^{[1,2]} \prever{{\alpha}}_{\{l_{2,k_1}\}i_1^{2,k_1}}^{[1,2]} \ldots \prever{{\alpha}}_{\{l_{k_0,k_1}\}i_1^{k_0,k_1}}^{[1,2]} \\
 \end{array}
}}
\]
where, defining:
\[
 \begin{array}{rl}
  {{\tau}}^{[1,2](k)} \left( \left\{ \featarg{a}{\ldots}, \featarg{a'}{\ldots}, \ldots \right\} \right)
  =
  {{\tau}}^{[1,2](k)} \left( a \right)
  {{\tau}}^{[1,2](k)} \left( a' \right)
  \ldots
 \end{array}
\]
we have:
\[
{\!\!\!\!\!\!\!\!\!\!\!\!\!\!\!\!\!\!\!\!\!\!\!\!\!\!\!\!\!\!\!\!\!\!\!\!\!\!\!\!\!\!\!\!\!\!\!\!\!\!\!\!\!\!\!\!\!\!\!\!\!\!\!\!{
 \begin{array}{l}
  \prever{K}^{[1,2]}{}_{i_1^1,i_1^2,\ldots}{}^{\underline{i}_1^1,\underline{i}_1^1,\ldots}
  \left(
  \left\{
  \featarg{\mathbb{X}'}{k'},
  \featarg{\mathbb{X}''}{k''},
  \ldots
  \featarg{\mathbb{X}''''}{k''''}
  \right\},
  \left\{
  \featarg{\mathbb{X}^{\backprime}}{k^{\backprime}},
  \featarg{\mathbb{X}^{\backprime\backprime}}{k^{\backprime\backprime}},
  \ldots
  \featarg{\mathbb{X}^{\backprime\backprime\backprime\backprime}}{k^{\backprime\backprime\backprime\backprime}};
  \allofW', \allofW^\backprime
  \right\}
  \right)
  = \ldots \\ \;\;\;\;\;\;\;\; \ldots 
  \left<
  \Prever{\bf F}^{[1,2]}{}_{:i_1^1,i_1^2,\ldots}
  \left(
  \left\{
  \featarg{\mathbb{X}'}{k'},
  \featarg{\mathbb{X}''}{k''},
  \ldots
  \featarg{\mathbb{X}''''}{k''''}
  \right\};
  \allofW'
  \right),
  \Prever{\bf F}^{[1,2]}{}^{:\underline{i}_1^1,\underline{i}_1^2,\ldots}
  \left(
  \left\{
  \featarg{\mathbb{X}^{\backprime}}{k^{\backprime}},
  \featarg{\mathbb{X}^{\backprime\backprime}}{k^{\backprime\backprime}},
  \ldots
  \featarg{\mathbb{X}^{\backprime\backprime\backprime\backprime}}{k^{\backprime\backprime\backprime\backprime}}
  \right\};
  \allofW^\backprime
  \right)
  \right> \\
 \end{array}
}}
\]
and:
\[
{{
 \begin{array}{l}
  \Prever{\bf F}^{[1,2]}{}_{:i_1^1,i_1^2,\ldots}
  \left(
  \left\{
  \featarg{\mathbb{X}'}{k'},
  \featarg{\mathbb{X}''}{k''},
  \ldots
  \featarg{\mathbb{X}''''}{k''''}
  \right\};
  \allofW'
  \right)
  = \ldots \\ \;\;\;\;\ldots =
  {{\tau}}^{[1,2](k') } \left( \mathbb{X}' \right)
  \Postver{\bf F}^{[1]}{}_{:i_1^1,i_1^2,\ldots,i_1^{|\mathbb{X}'|}} \left( \mathbb{X}'; \allofW' \right)^{\otimes k'}
  \otimes
  {{\tau}}^{[1,2](k'') } \left( \mathbb{X}'' \right)
  \Postver{\bf F}^{[1]}{}_{:i_1^{1+|\mathbb{X}'|},i_1^{2+|\mathbb{X}'|},\ldots,i_1^{|\mathbb{X}''|+|\mathbb{X}'|}} \left( \mathbb{X}''; \allofW' \right)^{\otimes k''}
  \otimes \ldots \\
 \end{array}
}}
\]
noting that with this definition:
\[
{\!\!\!\!\!\!\!\!\!\!\!\!\!\!\!\!\!\!\!\!\!\!\!\!\!\!\!\!\!\!\!\!\!\!\!\!\!\!\!\!\!\!\!\!\!\!\!\!{
 \begin{array}{l}
  \prever{K}^{[1,2]}{}_{i_1}{}^{i_1^{1,1},i_1^{2,1},\ldots,i_1^{k_0,1},i_1^{1,2},i_1^{2,2},\ldots,i_1^{k_0,2},\ldots,i_1^{1,k_1},i_1^{2,k_1},\ldots,i_1^{k_0,k_1}}
  \Bigg(
  \left\{
  \featarg{\left\{ \featarg{{\bf x}}{k_0} \right\}}{k_1}
  \right\},
  \ldots \\ \ldots
  \left\{
  \featarg{\left\{ \featarg{{\bf x}_{\{l_{1,1  }\}}}{1}, \featarg{{\bf x}_{\{l_{2,1  }\}}}{1}, \ldots \featarg{{\bf x}_{\{l_{k_0,1  }\}}}{1} \right\}}{1},
  \featarg{\left\{ \featarg{{\bf x}_{\{l_{1,2  }\}}}{1}, \featarg{{\bf x}_{\{l_{2,2  }\}}}{1}, \ldots \featarg{{\bf x}_{\{l_{k_0,2  }\}}}{1} \right\}}{1},
  \ldots,
  \featarg{\left\{ \featarg{{\bf x}_{\{l_{1,k_1}\}}}{1}, \featarg{{\bf x}_{\{l_{2,k_1}\}}}{1}, \ldots \featarg{{\bf x}_{\{l_{k_0,k_1}\}}}{1} \right\}}{1}
  \right\};
  \allofW_+, \allofW
  \Bigg)
  =
  \ldots \\ \;\;\;\;\;\; \ldots
  {{\tau}}^{[1,2](k_1)} \left( \postver{x}_{i_1} \right)
  \ldots \\ \;\;\;\;\;\; \ldots
  \postver{K}^{[1]}{}_{i_1}{}^{i_1^{1,1}, i_1^{2,1}, \ldots, i_1^{k_0,1}}
  \left(
  \left\{ \featarg{{\bf x}}{k_0} \right\},
  \left\{ \featarg{{\bf x}_{\{l_{1,1}\}}}{1}, \featarg{{\bf x}_{\{l_{2,1}\}}}{1}, \ldots, \featarg{{\bf x}_{\{l_{k_0,1}\}}}{1} \right\};
  \allofW_+, \allofW
  \right)
  \ldots \\ \;\;\;\;\;\; \ldots
  \postver{K}^{[1]}{}_{i_1}{}^{i_1^{1,2}, i_1^{2,2}, \ldots, i_1^{k_0,2}}
  \left(
  \left\{ \featarg{{\bf x}}{k_0} \right\},
  \left\{ \featarg{{\bf x}_{\{l_{1,2}\}}}{1}, \featarg{{\bf x}_{\{l_{2,2}\}}}{1}, \ldots, \featarg{{\bf x}_{\{l_{k_0,2}\}}}{1} \right\};
  \allofW_+, \allofW
  \right)
  \ldots \\ \;\;\;\;\;\; \ldots
  \ldots
  \ldots \\ \;\;\;\;\;\; \ldots
  \postver{K}^{[1]}{}_{i_1}{}^{i_1^{1,k_1}, i_1^{2,k_1}, \ldots, i_1^{k_0,k_1}}
  \left(
  \left\{ \featarg{{\bf x}}{k_0} \right\},
  \left\{ \featarg{{\bf x}_{\{l_{1,k_1}\}}}{1}, \featarg{{\bf x}_{\{l_{2,k_1}\}}}{1}, \ldots, \featarg{{\bf x}_{\{l_{k_0,k_1}\}}}{1} \right\};
  \allofW_+, \allofW
  \right)
  \ldots \\ \;\;\;\;\;\; \ldots
  {{\tau}}^{[0,1](1)} \left( \postver{x}_{\{l_{1,1}\}}^{[0]}{}_{{i}_0^{1,1}} \right)
  {{\tau}}^{[0,1](1)} \left( \postver{x}_{\{l_{2,1}\}}^{[0]}{}_{{i}_0^{2,1}} \right)
  \ldots
  {{\tau}}^{[0,1](1)} \left( \postver{x}_{\{l_{k_0,1}\}}^{[0]}{}_{{i}_0^{k_0,1}} \right)
  \ldots \\ \;\;\;\;\;\; \ldots
  {{\tau}}^{[0,1](1)} \left( \postver{x}_{\{l_{1,2}\}}^{[0]}{}_{{i}_0^{1,2}} \right)
  {{\tau}}^{[0,1](1)} \left( \postver{x}_{\{l_{2,2}\}}^{[0]}{}_{{i}_0^{2,2}} \right)
  \ldots
  {{\tau}}^{[0,1](1)} \left( \postver{x}_{\{l_{k_0,2}\}}^{[0]}{}_{{i}_0^{k_0,2}} \right)
  \ldots \\ \;\;\;\;\;\; \ldots
  \ldots
  \ldots \\ \;\;\;\;\;\; \ldots
  {{\tau}}^{[0,1](1)} \left( \postver{x}_{\{l_{1,k_1}\}}^{[0]}{}_{{i}_0^{1,k_1}} \right)
  {{\tau}}^{[0,1](1)} \left( \postver{x}_{\{l_{2,k_1}\}}^{[0]}{}_{{i}_0^{2,k_1}} \right)
  \ldots
  {{\tau}}^{[0,1](1)} \left( \postver{x}_{\{l_{k_0,k_1}\}}^{[0]}{}_{{i}_0^{k_0,k_1}} \right)
 \end{array}
}}
\]
and continuing to the output of this layer:
\[
{\!\!\!\!\!\!\!\!\!\!\!\!\!\!\!\!\!\!\!\!\!\!\!\!\!\!\!\!\!\!\!\!\!\!\!\!\!\!\!\!\!\!\!\!\!\!{
 \begin{array}{rl}
  \changein \postver{x}_{i_2}^{[2]}
  &\!\!\!\!=
  \mathop{\sum}\limits_{{\bf k} \in \mathbb{Z}_+^2}
  \frac{1}{{\bf k}!}
  \mathop{\sum}\limits_{l_{\mathbb{N}_{k_0},\mathbb{N}_{k_1}}}
  \postver{K}^{[2]}{}_{i_2}{}^{i_2^{1,1},i_2^{2,1},\ldots,i_2^{k_0,1},i_2^{1,2},i_2^{2,2},\ldots,i_2^{k_0,2},\ldots,i_2^{1,k_1},i_2^{2,k_1},\ldots,i_2^{k_0,k_1}}
  \Bigg(
  \left\{
  \featarg{\left\{ \featarg{{\bf x}}{k_0} \right\}}{k_1}
  \right\},
  \ldots \\ & \!\!\!\!\!\!\!\!\!\!\!\!\ldots
  \left\{
  \featarg{\left\{ \featarg{{\bf x}_{\{l_{1,1  }\}}}{1}, \featarg{{\bf x}_{\{l_{2,1  }\}}}{1}, \ldots \featarg{{\bf x}_{\{l_{k_0,1  }\}}}{1} \right\}}{1},
  \featarg{\left\{ \featarg{{\bf x}_{\{l_{1,2  }\}}}{1}, \featarg{{\bf x}_{\{l_{2,2  }\}}}{1}, \ldots \featarg{{\bf x}_{\{l_{k_0,2  }\}}}{1} \right\}}{1},
  \ldots,
  \featarg{\left\{ \featarg{{\bf x}_{\{l_{1,k_1}\}}}{1}, \featarg{{\bf x}_{\{l_{2,k_1}\}}}{1}, \ldots \featarg{{\bf x}_{\{l_{k_0,k_1}\}}}{1} \right\}}{1}
  \right\};
  \allofW_+, \allofW
  \Bigg)
  \ldots \\ & \ldots 
  \postver{{\alpha}}_{\{l_{1,1  }\}i_2^{1,1  }}^{[1,2]} \postver{{\alpha}}_{\{l_{2,1  }\}i_2^{2,1  }}^{[1,2]} \ldots \postver{{\alpha}}_{\{l_{k_0,1  }\}i_2^{k_0,1  }}^{[1,2]}
  \postver{{\alpha}}_{\{l_{1,2  }\}i_2^{1,2  }}^{[1,2]} \postver{{\alpha}}_{\{l_{2,2  }\}i_2^{2,2  }}^{[1,2]} \ldots \postver{{\alpha}}_{\{l_{k_0,2  }\}i_2^{k_0,2  }}^{[1,2]}
  \ldots
  \postver{{\alpha}}_{\{l_{1,k_1}\}i_2^{1,k_1}}^{[1,2]} \postver{{\alpha}}_{\{l_{2,k_1}\}i_2^{2,k_1}}^{[1,2]} \ldots \postver{{\alpha}}_{\{l_{k_0,k_1}\}i_2^{k_0,k_1}}^{[1,2]} \\
 \end{array}
}}
\]
where:
\[
{\!\!\!\!\!\!\!\!\!\!\!\!\!\!\!\!\!\!\!\!\!\!\!\!\!\!\!\!\!\!\!\!\!\!\!\!\!\!\!\!\!\!\!\!\!\!\!\!\!\!\!\!\!\!\!\!\!\!\!\!{
 \begin{array}{rl}
  \postver{K}^{[2]}{}_{i_2}{}^{i'_2}
  \left(
  \left\{ \featarg{\left\{ \featarg{{\bf x}'}{1} \right\}}{1} \right\},
  \left\{ \featarg{\left\{ \featarg{{\bf x}^\backprime}{1} \right\}}{1} \right\}; 
  \allofW', \allofW^\backprime
  \right)
  &\!\!\!\!=
  \delta{}_{i_2}{}^{i'_2}
  \Sigma^{[2]}
  \left( {\bf x}', {\bf x}^\backprime \right)
  +
  {W}^{\prime[1,2]}{}^{i'_1}{}_{i'_2}
  {W}^{\backprime[1,2]}{}^{i_1^{\backprime}}{}_{i_2^{\backprime}}
  \prever{K}^{[1,2]}{}_{i'_1}{}^{i_1^{\backprime}}
  \left(
  \left\{ \featarg{\left\{ \featarg{{\bf x}'}{1} \right\}}{1} \right\},
  \left\{ \featarg{\left\{ \featarg{{\bf x}^{\backprime}}{1} \right\}}{1} \right\};
  \allofW', \allofW^\backprime
  \right) \\
  &\!\!\!\!=
  \left<
  \Postver{\bf F}^{[2]}{}_{i'_2}
  \left(
  \left\{ \featarg{\left\{ \featarg{{\bf x}'}{1} \right\}}{1} \right\};
  \allofW'
  \right),
  \Postver{\bf F}^{[2]}{}_{i_2^{\backprime}}
  \left(
  \left\{ \featarg{\left\{ \featarg{{\bf x}^{\backprime}}{1}  \right\}}{1}  \right\};
  \allofW^\backprime
  \right)
  \right> \\
 \end{array}
}}
\]
and otherwise:
\[
{\!\!\!\!\!\!\!\!\!\!\!\!\!\!\!\!\!\!\!\!\!\!\!\!\!\!\!\!\!\!\!\!\!\!\!\!\!\!\!\!\!\!\!\!\!\!\!\!\!\!\!\!\!\!\!\!\!\!\!\!\!\!\!\!{
 \begin{array}{l}
  \postver{K}^{[2]}{}_{i_2^{\prime 1},i_2^{\prime\prime 2},\ldots}{}^{i_2^{\backprime 1},i_2^{\backprime\backprime 1},\ldots}
  \left(
  \left\{
  \featarg{\mathbb{X}'}{k'},
  \featarg{\mathbb{X}''}{k''},
  \ldots
  \featarg{\mathbb{X}''''}{k''''}
  \right\},
  \left\{
  \featarg{\mathbb{X}^{\backprime}}{k^{\backprime}},
  \featarg{\mathbb{X}^{\backprime\backprime}}{k^{\backprime\backprime}},
  \ldots
  \featarg{\mathbb{X}^{\backprime\backprime\backprime\backprime}}{k^{\backprime\backprime\backprime\backprime}}
  \right\};
  \allofW', \allofW^\backprime
  \right)
  \ldots \\ \;\;\;\;\;\;\;\; =
  {W}^{\prime[1,2]}{}^{i_1^{\prime 1}}{}_{i_2^{\prime 1}}
  {W}^{\prime[1,2]}{}^{i_1^{\prime\prime 2}}{}_{i_2^{\prime\prime 2}}
  \ldots
  {W}^{\backprime[1,2]}{}_{i_1^{\backprime 1}}{}^{i_2^{\backprime 1}}
  {W}^{\backprime[1,2]}{}_{i_1^{\backprime\backprime 2}}{}^{i_2^{\backprime\backprime 2}}
  \ldots \\ \;\;\;\;\;\;\;\;\;\;\;\;\;\;\;\;\;\;\;\; \ldots
  \prever{K}^{[1,2]}{}_{i_1^{\prime 1},i_1^{\prime\prime 2},\ldots}{}^{i_1^{\backprime 1},i_1^{\backprime\backprime 1},\ldots}
  \left(
  \left\{
  \featarg{\mathbb{X}'}{k'},
  \featarg{\mathbb{X}''}{k''},
  \ldots
  \featarg{\mathbb{X}''''}{k''''}
  \right\},
  \left\{
  \featarg{\mathbb{X}^{\backprime}}{k^{\backprime}},
  \featarg{\mathbb{X}^{\backprime\backprime}}{k^{\backprime\backprime}},
  \ldots
  \featarg{\mathbb{X}^{\backprime\backprime\backprime\backprime}}{k^{\backprime\backprime\backprime\backprime}}
  \right\};
  \allofW', \allofW^\backprime
  \right)
  \\ \;\;\;\;\;\;\;\; =
  \left<
  \Postver{\bf F}^{[2]}{}_{:i_2^{\prime 1},i_2^{\prime\prime 2},\ldots}
  \left(
  \left\{
  \featarg{\mathbb{X}'}{k'},
  \featarg{\mathbb{X}''}{k''},
  \ldots
  \featarg{\mathbb{X}''''}{k''''}
  \right\};
  \allofW'
  \right),
  \Postver{\bf F}^{[2]}{}^{:i_2^{\backprime 1},i_2^{\backprime\backprime 2},\ldots}
  \left(
  \left\{
  \featarg{\mathbb{X}^{\backprime}}{k^{\backprime}},
  \featarg{\mathbb{X}^{\backprime\backprime}}{k^{\backprime\backprime}},
  \ldots
  \featarg{\mathbb{X}^{\backprime\backprime\backprime\backprime}}{k^{\backprime\backprime\backprime\backprime}}
  \right\};
  \allofW^\backprime
  \right)
  \right> \\
 \end{array}
}}
\]
where:
\[
{{
 \begin{array}{l}
  \Postver{\bf F}^{[2]}{}_{i'_2}
  \left(
  \left\{ \featarg{\left\{ \featarg{{\bf x}'}{1} \right\}}{1} \right\}; \allofW'
  \right)
  = 
  \left[ \begin{array}{c}
  \left[ \delta_{i'_2,i''_2} \right]_{i''_2} \otimes \Prever{\bg{\varphi}}^{[2]} \left( {\bf x}' \right) \\

  {W}^{\prime[1,2]}{}^{i'_1}{}_{i'_2}
  \Prever{\bf F}^{[1,2]}{}_{i'_1}
  \left(
  \left\{ \featarg{\featarg{{\bf x}'}{1}}{1} \right\}; \allofW'
  \right) \\
  \end{array} \right]
 \end{array}
}}
\]
and otherwise:
\[
{\!\!\!\!\!\!\!\!\!\!\!\!\!\!\!\!\!\!\!\!\!\!\!\!\!\!\!\!\!\!\!\!{
 \begin{array}{rl}
  \Postver{\bf F}^{[2]}{}_{:i_2^1,i_2^2,\ldots}
  \left(
  \left\{
  \featarg{\mathbb{X}'}{k'},
  \featarg{\mathbb{X}''}{k''},
  \ldots
  \featarg{\mathbb{X}''''}{k''''}
  \right\};
  \allofW'
  \right)
  &\!\!\!\!=
  {W}^{\prime[1,2]}{}^{i_1^1}{}_{i_2^1}
  {W}^{\prime[1,2]}{}^{i_1^2}{}_{i_2^2}
  \ldots
  \Prever{\bf F}^{[1,2]}{}_{:i_1^1,i_1^2,\ldots}
  \left(
  \left\{
  \featarg{\mathbb{X}'}{k'},
  \featarg{\mathbb{X}''}{k''},
  \ldots
  \featarg{\mathbb{X}''''}{k''''}
  \right\};
  \allofW'
  \right) \\
 \end{array}
}}
\]

The process may then be repeated up to the output layer to obtain the 
result given in the body of the paper.  Note that we have reformulated 
the final version in terms of Kronecker products rather than indexed 
notation for the sake of clarity.  Note that in the body we extract the 
$\eta$ out of the representation.

\end{document}